\def\year{2022}\relax
\documentclass[letterpaper]{article} 
\usepackage{aaai22}  
\usepackage{times}  
\usepackage{helvet}  
\usepackage{courier}  
\usepackage[hyphens]{url}  
\usepackage{graphicx} 
\usepackage{natbib}  
\usepackage{caption} 
\DeclareCaptionStyle{ruled}{labelfont=normalfont,labelsep=colon,strut=off} 
\usepackage{stfloats}
\usepackage{algorithmic}
\usepackage[linesnumbered,ruled,noend]{algorithm2e}

\usepackage{amsmath,amssymb,amsfonts}
\usepackage{bm}
\usepackage{textcomp}
\usepackage{amsthm}
\usepackage{subfigure}
\usepackage[usenames,dvipsnames]{xcolor}
\usepackage{multirow}
\usepackage{siunitx}
\usepackage{cite}
\usepackage{bm}
\usepackage{paralist}

\usepackage[utf8]{inputenc}
\usepackage{kotex}
\usepackage{cancel}

\title{Graph Neural Controlled Differential Equations for Traffic Forecasting}
\author {
    Jeongwhan Choi, Hwangyong Choi, Jeehyun Hwang, Noseong Park
}
\affiliations{
    Yonsei University, Seoul, South Korea \\
    \{jeongwhan.choi, hwangyong753, hwanggh96, noseong\}@yonsei.ac.kr
}

\begin{document}

\maketitle

\begin{abstract}
Traffic forecasting is one of the most popular spatio-temporal tasks in the field of machine learning. A prevalent approach in the field is to combine graph convolutional networks and recurrent neural networks for the spatio-temporal processing. There has been fierce competition and many novel methods have been proposed. In this paper, we present the method of spatio-temporal graph neural controlled differential equation (STG-NCDE). Neural controlled differential equations (NCDEs) are a breakthrough concept for processing sequential data. We extend the concept and design two NCDEs: one for the temporal processing and the other for the spatial processing. After that, we combine them into a single framework. We conduct experiments with 6 benchmark datasets and 20 baselines. STG-NCDE shows the best accuracy in all cases, outperforming all those 20 baselines by non-trivial margins.
\end{abstract}

\section{Introduction}
The spatio-temporal graph data frequently happens in real-world applications, ranging from traffic to climate forecasting~\cite{zaytar2016sequence, shi2015convolutional,shi2017deep, liu2016application, racah2016extremeweather, kurth2018exascale, cheng2018ensemble, cheng2018neural, hossain2015forecasting, ren2021deep, tekin2021spatio,li2018dcrnn_traffic,bing2018stgcn,wu2019graphwavenet,guo2019astgcn,bai2019STG2Seq,song2020stsgcn,huang2020lsgcn,NEURIPS2020_ce1aad92,li2021stfgnn,chen2021ZGCNET,fang2021STODE}. For instance, the traffic forecasting task launched by California Performance of Transportation (PeMS) is one of the most popular problems in the area of spatio-temporal processing~\cite{chen2001freeway,bing2018stgcn,guo2019astgcn}.

Given a time-series of graphs $\{\mathcal{G}_{t_i} \stackrel{\text{def}}{=} (\mathcal{V},\mathcal{E},\bm{F}_{i}, t_i)\}_{i=0}^{N}$, where $\mathcal{V}$ is a fixed set of nodes, $\mathcal{E}$ is a fixed set of edges, $t_i$ is a time-point when $\mathcal{G}_{t_i}$ is observed, and $\bm{F}_{i} \in \mathbb{R}^{|\mathcal{V}| \times D}$ is a feature matrix at time $t_i$ which contains $D$-dimensional input features of the nodes, the spatio-temporal forecasting is to predict $\hat{\bm{Y}} \in \mathbb{R}^{|\mathcal{V}| \times S  \times M}$, e.g., predicting the traffic volume for each location of a road network for the next $S$ time-points (or horizons) given past $N+1$ historical traffic patterns, where $|\mathcal{V}|$ is the number of locations to predict and $M=1$ because the volume is a scalar, i.e., the number of vehicles. We note that $\mathcal{V}$ and $\mathcal{E}$ do not change over time --- in other words, the graph topology is fixed --- whereas the node input features can change over time. We use upper boldface to denote matrices and lower boldface for vectors.

\begin{figure}[t]
    \centering
    \includegraphics[width=0.8\columnwidth]{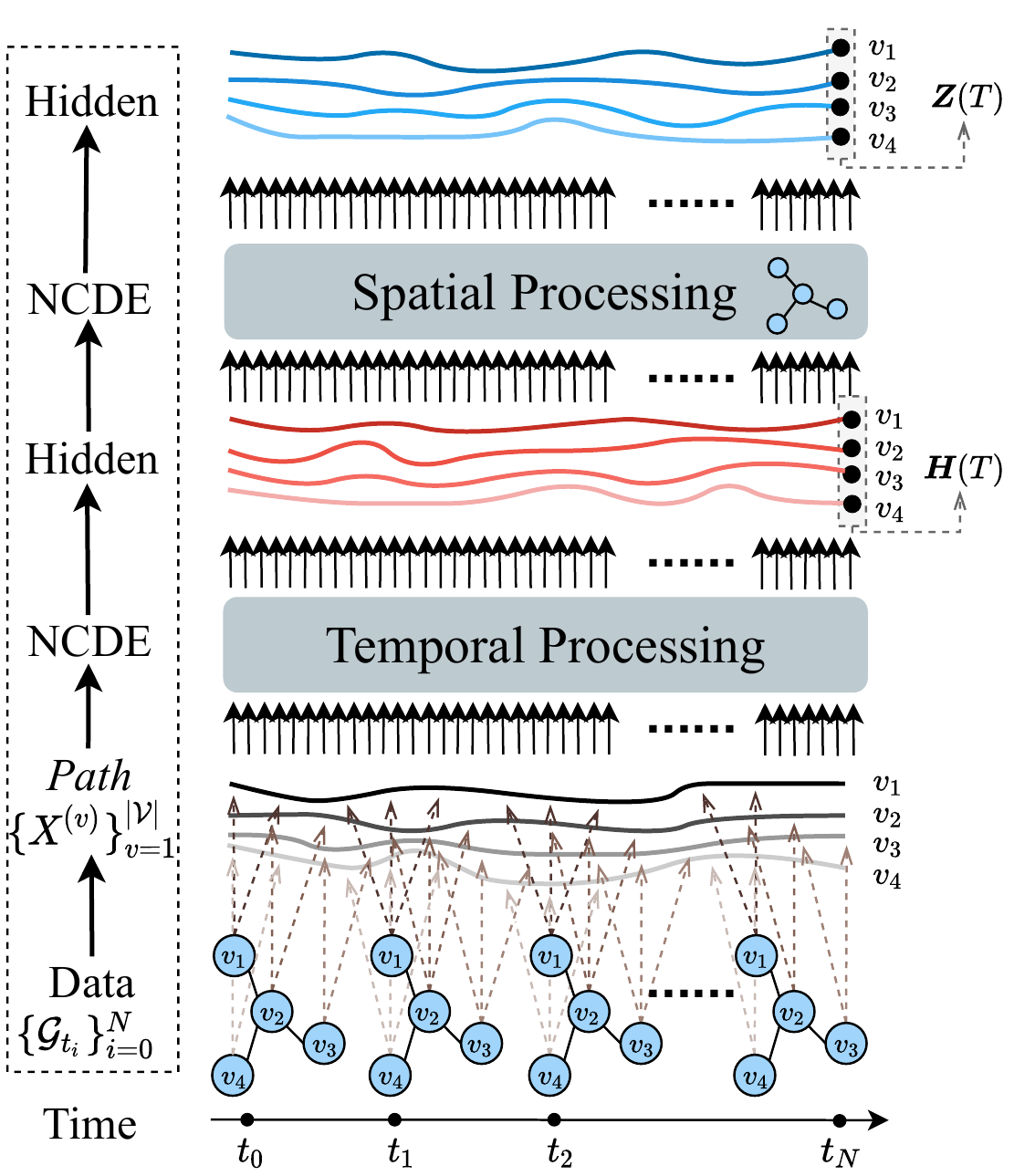}
    \caption{The overall workflow in our proposed STG-NCDE}
    \label{fig:stgncde}
\end{figure}

\begin{figure}[t]
    \centering
    \includegraphics[width=0.8\columnwidth]{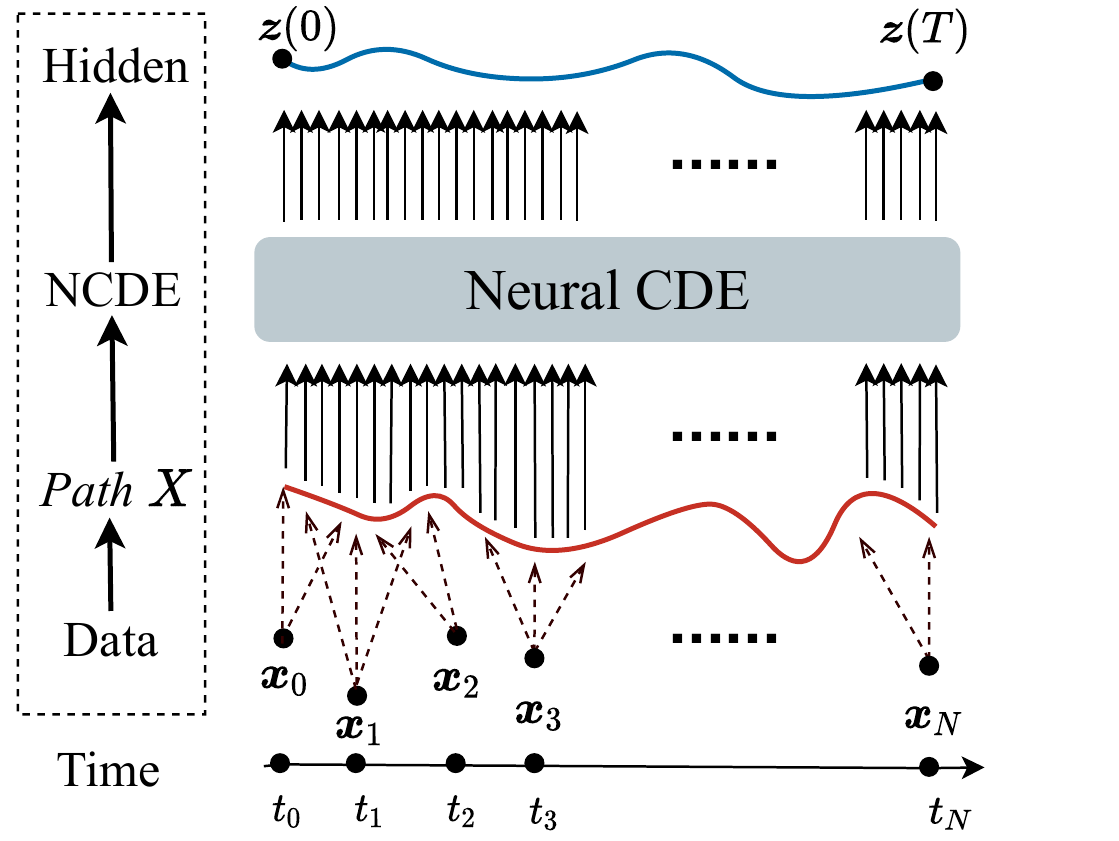}
    \caption{The overall workflow of the original NCDE for processing time-series. The path $X$ is created from $\{(t_i, \bm{x}_{i})\}_{i=0}^N$ by an interpolation algorithm and therefore, this technology is robust to irregular time-series data.}
    \label{fig:ncde}
\end{figure}

For this task, a diverse set of techniques have been proposed. In this paper, however, we design a method based on neural controlled differential equations (NCDEs) for the first time. NCDEs, which are considered as a \emph{continuous} analogue to recurrent neural networks (RNNs), can be written as follows:
\begin{align}\label{eq:ncde}
\bm{z}(T) &= \bm{z}(0) + \int_{0}^{T} f(\bm{z}(t);\bm{\theta}_f) dX(t)\\&= \bm{z}(0) + \int_{0}^{T} f(\bm{z}(t);\bm{\theta}_f) \frac{dX(t)}{dt} dt,\label{eq:ncde2}
\end{align}where $X$ is a continuous path taking values in a Banach space. The entire trajectory of $\bm{z}(t)$ is controlled over time by the path $X$ (cf. Fig.~\ref{fig:ncde}). Leaning the CDE function $f$ for a downstream task is a key point in NCDEs.

The theory of the controlled differential equation (CDE) had been developed to extend the stochastic differential equation and the It\^{o} calculus far beyond the semimartingale setting of $X$ --- in other words, Eq.~\eqref{eq:ncde} reduces to the stochastic differential equation if and only if $X$ meets the semimartingale requirement. For instance, a prevalent example of the path $X$ is a Wiener process in the case of the stochastic differential equation. In CDEs, however, the path $X$ does not need to be such semimartingale or martingale processes. NODEs are a technology to parameterize such CDEs and learn from data. In addition, Eq.~\eqref{eq:ncde2} continuously reads the values $\frac{dX(t)}{dt}$ and integrates them over time. In this regard, NODEs are equivalent to continuous RNNs and show the state-of-the-art accuracy in many time-series tasks and data.

However, it has not been studied yet how to combine the NCDE technology (i.e., temporal processing) and the graph convolutional processing technology (i.e., spatial processing). We integrate them into a single framework to solve the spatio-temporal forecasting problem.

In the original setting of NCDEs, there exists a single time-series, denoted $\{(t_i, \bm{x}_{i})\}_{i=0}^N$, where $\bm{x}_{i} \in \mathbb{R}^D$ is a $D$-dimensional vector and $t_i$ is a time-point when $\bm{x}_{i}$ is observed. In our setting, however, there exist $|\mathcal{V}|$ different time-series patterns to consider, each of which is somehow correlated to neighboring time-series patterns. Figs.~\ref{fig:stgncde} and.~\ref{fig:ncde} show the difference between them.

The pre-processing step in our method is to create a continuous path $X^{(v)}$ for each node $v \in \mathcal{V}$. For this, we use the same technique as that in the original NCDE design. Given a discrete time-series $\{\bm{x}_{i}\}_{i=0}^N$, the original NCDE runs an interpolation algorithm to build its continuous path. We apply the same method for each node separately, and a set of paths, denoted $\{X^{(v)}\}_{v=1}^{|\mathcal{V}|}$, will be created.

The main step is to jointly apply a spatial and a temporal processing method to $\{X^{(v)}\}_{v=1}^{|\mathcal{V}|}$, considering its graph connectivity. In our case, we design an NCDE model equipped with a graph processing technique for both the spatial and the temporal processing. We then derive the last hidden vector $\bm{z}^{(v)}(T)$ for each node $v$ and there is the last output layer to predict $\hat{\bm{y}}^{(v)} \in \mathbb{R}^{S \times M}$, which collectively constitutes $\hat{\bm{Y}} \in \mathbb{R}^{|\mathcal{V}| \times S \times M}$.

We conduct experiments with 6 benchmark datasets collected by California Performance of Transportation (PeMS), which are the most widely used datasets in this topic, to compare with 20 baseline methods. Our proposed method clearly outperforms all those methods in terms of three standard evaluation metrics. Our contributions can be summarized as follows:
\begin{compactenum}
    \item We design two NCDEs for learning the temporal and spatial dependencies of traffic conditions and combine them into a single framework.
    \item NCDEs are robust to irregular time-series by the design. Owing to this characteristic, our method is also robust to the irregularity of the temporal sequence, i.e., some observations can be missing. 
    \item Our large-scale experiments with 6 datasets and 20 baselines clearly show the efficacy of the proposed method. We, for the first time, perform irregular traffic forecasting to reflect real-world environments where sensing values can be missing (see Table~\ref{tbl:missing_pemsd4} and Table~\ref{tbl:missing_pemsd8}).
\end{compactenum}

\section{Related Work and Preliminaries}
In this section, we summarize our literature review related to our NCDE-based spatio-temporal forecasting.
\subsection{Neural ordinary differential equations (NODEs)}
Prior to NCDEs, neural ordinary differential equations (NODEs) introduced how to \emph{continuously} model residual neural networks (ResNets) with differential equations. NODE can be written as follows:
\begin{align}\label{eq:node}
\bm{z}(T) = \bm{z}(0) + \int_{0}^{T} f(\bm{z}(t), t;\bm{\theta}_f) dt;
\end{align}where the neural network parameterized by $\bm{\theta}_f$ approximates $\frac{\bm{z}(t)}{dt}$, and we rely on various ODE solvers to solve the integral problem, ranging from the explicit Euler method to the 4th order Runge--Kutta (RK4) method and the Dormand--Prince (DOPRI) method ~\cite{DORMAND198019}.

In particular, Eq.~\eqref{eq:node} reduces to the residual connection when being solved by the explicit Euler method. In this regard, NODEs generalizes ResNets in a continuous manner. STGODE utilizes this NODE technology to solve the spatio-temporal forecasting problem ~\cite{fang2021STODE}.

\subsection{Neural controlled differential equations (NCDEs)}
Whereas NODEs generalize ResNets, NCDEs in Eq.~\eqref{eq:ncde} generalize RNNs in a continuous manner. Controlled differential equations (CDEs) are a more advanced concept than ordinary differential equations (ODEs). The integral problem in Eq.~\eqref{eq:node} is a Rienmann integral problem whereas it is a Riemann--Stieltjes integral problem in Eq.~\eqref{eq:ncde2}. The original CDE formulation in Eq.~\eqref{eq:ncde} reduces Eq.~\eqref{eq:ncde2} where $\frac{\bm{z}(t)}{dt}$ is approximated by $f(\bm{z}(t);\bm{\theta}_f) \frac{dX(t)}{dt}$.

Once $\frac{\bm{z}(t)}{dt}$ is somehow successfully formulated in a close math form, we can utilize those existing ODE solvers to solve Eq.~\eqref{eq:ncde2}. Therefore, many techniques developed for solving NODEs can be applied to NCDEs as well.

\subsection{Traffic forecasting}
The problem of traffic forecasting is an emerging research topic in the field of spatio-temporal machine learning. When being solved in high accuracy, it has non-trivial impacts to our daily life. We introduce several milestone papers in this field. DCRNN~\cite{li2018dcrnn_traffic} combines graph convolution with recurrent neural networks in an encoder-decoder manner. STGCN~\cite{bing2018stgcn} combines graph convolution with gated temporal convolution. GraphWaveNet~\cite{wu2019graphwavenet} combines adaptive graph convolution with dilated casual convolution to capture spatial-temporal dependencies. ASTGCN~\cite{guo2019astgcn} utilizes both the spatial-temporal attention mechanism and the spatial-temporal convolution. STG2Seq~\cite{bai2019STG2Seq} uses a multiple gated graph convolutional module and a seq2seq architecture with an attention mechanisms to make multi-step prediction. STSGCN~\cite{song2020stsgcn} utilizes multiple localized spatial-temporal subgraph modules to synchronously capture the localized spatial-temporal correlations directly. LSGCN~\cite{huang2020lsgcn} integrates a novel attention mechanism and graph convolution into a spatial gated block. AGCRN~\cite{NEURIPS2020_ce1aad92} utilizes node adaptive parameter learning to capture node-specific spatial and temporal correlations in time-series data automatically without a pre-defined graph. STFGNN~\cite{li2021stfgnn} captures hidden spatial-dependenies by a novel fusion operation of various spatial and temporal graphs, treated for different time periods in parallel. Z-GCNETs~\cite{chen2021ZGCNET} integrates the new time-aware zigzag topological layer into time-conditioned GCNs. STGODE~\cite{fang2021STODE} captures spatial-temporal dynamics through a tensor-based ODE.

\section{Proposed Method}
The spatio-temporal processing of a time-series of graphs $\{\mathcal{G}_{t_i} \stackrel{\text{def}}{=} (\mathcal{V},\mathcal{E},\bm{F}_{i})\}_{i=0}^{N}$ is obviously more difficult than the spatial processing only (i.e., GCNs) or the temporal processing only (i.e., RNNs). As such, there have been proposed many neural networks combining GCNs and RNNs. In this paper, we design a novel spatio-temporal model based on the NCDE and the adaptive topology generation technologies. We describe our proposed method in this section. We first review its overall design and then introduce details.

\subsection{Overall design}

Our method includes one pre-processing and one main processing steps as follows:
\begin{compactenum}
    \item Its pre-processing step is to create a continuous path $X^{(v)}$ for each node $v$, where $1 \leq v \leq |\mathcal{V}|$, from $\{\bm{F}_{i}^{(v)}\}_{i=0}^{N}$. $\bm{F}_{i}^{(v)} \in \mathbb{R}^D$ means the $v$-th row of $\bm{F}_{i}$, and $\{\bm{F}_{i}^{(v)}\}$ stands for the time-series of the input features of $v$. We use the natural cubic spline method for interpolating the discrete time-series $\{\bm{F}_{i}^{(v)}\}$ and building a continuous path. Among many, the natural cubic spline has a couple of suitable characteristics to be used in our method: i) it creates a continuous path and ii) the created path is twice differentiable. In particular, the second characteristic is important when it comes to calculating the gradients of the proposed model.
    \item The above pre-processing step happens before training our model. Then, our main step, which combines a GCN and an NCDE technologies, calculates the last hidden vector for each node $v$, denoted $\bm{z}^{(v)}(T)$.
    \item After that, we have an output layer to predict $\hat{\bm{y}}^{(v)} \in \mathbb{R}^{S \times M}$ for each node $v$. After collecting those predictions for all nodes in $\mathcal{V}$, we have the prediction matrix $\hat{\bm{Y}} \in \mathbb{R}^{|\mathcal{V}| \times S \times M}$.
\end{compactenum}

\subsection{Graph neural controlled differential equations}
Our proposed spatio-temporal graph neural controlled differential equation (STG-NCDE) consists of two NCDEs: one for processing the temporal information and the other for processing the spatial information.

\paragraph{Temporal processing} The first NCDE for the temporal processing can be written as follows:
\begin{align}
\bm{h}^{(v)}(T) &= \bm{h}^{(v)}(0) + \int_{0}^{T} f(\bm{h}^{(v)}(t);\bm{\theta}_f) \frac{dX^{(v)}(t)}{dt} dt, \label{eq:type1}
\end{align} where $\bm{h}^{(v)}(t)$ is a hidden trajectory (over time $t \in [0,T]$) of the temporal information of node $v$. After stacking $\bm{h}^{(v)}(t)$ for all $v$, we can define a matrix $\bm{H}(t) \in \mathbb{R}^{|\mathcal{V}| \times \dim(\bm{h}^{(v)})}$. Therefore, the trajectory created by $\bm{H}(t)$ over time $t$ contains the hidden information of the temporal processing results. Eq.~\eqref{eq:type1} can be equivalently rewritten as follows using the matrix notation:
\begin{align}
\bm{H}(T) &= \bm{H}(0) + \int_{0}^{T} f(\bm{H}(t);\bm{\theta}_f) \frac{d\bm{X}(t)}{dt} dt, \label{eq:type1-2}
\end{align}where $\bm{X}(t)$ is a matrix whose $v$-th row is $X^{(v)}$. The CDE function $f$ separately processes each row in $\bm{H}(t)$. The key in this design is how to define the CDE function $f$ parameterized by $\bm{\theta}_f$. We will describe shortly how to define it. One good thing is that $f$ does not need to be a RNN. By designing it with fully-connected layers only, for instance, Eq.~\eqref{eq:type1-2} converts it to a \emph{continuous} RNN. 

\paragraph{Spatial processing} After that, the second NCDE starts for its spatial processing as follows:
\begin{align}\begin{split}
\bm{Z}(T) = \bm{Z}(0) + \int_{0}^{T} g(\bm{Z}(t);\bm{\theta}_g) \frac{d\bm{H}(t)}{dt} dt,\label{eq:type2}
\end{split}\end{align}where the hidden trajectory $\bm{Z}(t)$ is controlled by $\bm{H}(t)$ which is created by the temporal processing.

After combining Eqs.~\eqref{eq:type1-2} and~\eqref{eq:type2}, we have the following single equation which incorporates both the temporal and the spatial processing:
\begin{align}\begin{split}
\bm{Z}(T) = \bm{Z}(0) + \int_{0}^{T} g(\bm{Z}(t);\bm{\theta}_g)f(\bm{H}(t);\bm{\theta}_f) \frac{d\bm{X}(t)}{dt} dt , \label{eq:type2-2}
\end{split}\end{align}where $\bm{Z}(t) \in \mathbb{R}^{|\mathcal{V}| \times \dim(\bm{z}^{(v)})}$ is a matrix created after stacking the hidden trajectory $\bm{z}^{(v)}$ for all $v$. In this NCDE, a hidden trajectory $\bm{z}^{(v)}$ is created after considering the trajectories of its neighbors --- for ease of writing, we use the matrix notation in Eqs.~\eqref{eq:type2} and~\eqref{eq:type2-2}. The key part is how to design the CDE function $g$ parameterized by $\bm{\theta}_g$ for the spatial processing.

\paragraph{CDE functions} We now describe the two CDE functions $f$ and $g$. The definition of $f:\mathbb{R}^{|\mathcal{V}| \times \dim(\bm{h}^{(v)})} \rightarrow \mathbb{R}^{|\mathcal{V}| \times \dim(\bm{h}^{(v)})}$ is as as follows:
\begin{align}\begin{split}
f(\bm{H}(t);\bm{\theta}_f) &= \psi(\texttt{FC}_{|\mathcal{V}| \times \dim(\bm{h}^{(v)}) \rightarrow |\mathcal{V}| \times \dim(\bm{h}^{(v)})}(\bm{A}_{K})),\\
&\vdots\\
\bm{A}_1 &= \sigma(\texttt{FC}_{|\mathcal{V}| \times \dim(\bm{h}^{(v)}) \rightarrow |\mathcal{V}| \times \dim(\bm{h}^{(v)})}(\bm{A}_0)),\\
\bm{A}_0 &= \sigma(\texttt{FC}_{|\mathcal{V}| \times \dim(\bm{h}^{(v)}) \rightarrow |\mathcal{V}| \times \dim(\bm{h}^{(v)})}(\bm{H}(t))),\label{eq:fun_f}
\end{split}\end{align}
where $\sigma$ is a rectified linear unit, $\psi$ is a hyperbolic tangent, and $\mathtt{FC}_{input\_size \rightarrow output\_size}$ means a fully-connected layer whose input size is $input\_size$ and output size is also $output\_size$. $\bm{\theta}_f$ refers to the parameters of the fully-connected layers. This function $f$ independently processes each row of $\bm{H}(t)$ with the $K$ fully connected-layers.

For the spatial processing, we need to define one more CDE function $g$. The definition of $g:\mathbb{R}^{|\mathcal{V}| \times \dim(\bm{z}^{(v)})} \rightarrow \mathbb{R}^{|\mathcal{V}| \times \dim(\bm{z}^{(v)})}$ is as follows:
\begin{align}
g(\bm{Z}(t);\bm{\theta}_g) &= \psi(\texttt{FC}_{|\mathcal{V}| \times \dim(\bm{z}^{(v)}) \rightarrow |\mathcal{V}| \times \dim(\bm{z}^{(v)})}(\bm{B}_1)),\label{eq:fun_g1}\\
\bm{B}_1 &= (\bm{I} + \phi(\sigma(\bm{E}\cdot\bm{E}^{\intercal})))\bm{B}_0\bm{W}_{spatial},\label{eq:fun_g2}\\
\bm{B}_0 &= \sigma(\texttt{FC}_{|\mathcal{V}| \times \dim(\bm{z}^{(v)}) \rightarrow |\mathcal{V}| \times \dim(\bm{z}^{(v)})}(\bm{Z}(t))),\label{eq:fun_g3}
\end{align}
where $\bm{I}$ is the $|\mathcal{V}| \times |\mathcal{V}|$ identity matrix, $\phi$ is a softmax activation, $\bm{E} \in \mathbb{R}^{|\mathcal{V}| \times C} $ is a trainable node-embedding matrix, $\bm{E}^{\intercal}$ is its transpose, and $\bm{W}_{spatial}$ is a trainable weight transformation matrix. Conceptually, $\phi(\sigma(\bm{E}\cdot\bm{E}^{\intercal}))$ corresponds to the normalized adjacency matrix $\bm{D}^{-\frac{1}{2}}\bm{A}\bm{D}^{-\frac{1}{2}}$, where $\bm{A} = \sigma(\bm{E}\cdot\bm{E}^{\intercal})$ and the softmax activation plays a role of normalizing the adaptive adjacency matrix ~\cite{wu2019graphwavenet,NEURIPS2020_ce1aad92}. We also note that Eq.~\eqref{eq:fun_g2} is identical to the first order Chebyshev polynomial expansion of the graph convolution operation~\cite{kipf2017semi} with the normalized adaptive adjacency matrix. Eqs.~\eqref{eq:fun_g1} and~\eqref{eq:fun_g3} do not mix the rows of their input matrices $\bm{Z}(t)$ and $\bm{B}_1$. It is Eq.~\eqref{eq:fun_g2} where the rows of $\bm{B}_0$ are mixed for the spatial processing.

\paragraph{Initial value generation} The initial value of the temporal processing, i.e., $\bm{H}(0)$, is created from $\bm{F}_{t_0}$ as follows: $\bm{H}(0) = FC_{D \rightarrow \dim(\bm{h}^{(v)})}(\bm{F}_{t_0})$. We also use the following similar strategy to generate $\bm{Z}(0)$: $\bm{Z}(0) = FC_{\dim(\bm{h}^{(v)}) \rightarrow \dim(\bm{z}^{(v)})}(\bm{H}(0))$. After generating these initial values for the two NCDEs, we can calculate $\bm{Z}(T)$ after solving the Riemann--Stieltjes integral problem in Eq.~\eqref{eq:type2-2}.

\subsection{How to train}
To implement Eq.~\eqref{eq:type2-2} --- we do not separately implement Eqs.~\eqref{eq:type1-2} and~\eqref{eq:type2} --- we define the following augmented ODE:
\begin{align}
\frac{d}{dt}{\begin{bmatrix}
  \bm{Z}(t) \\
  \bm{H}(t) \\
  \end{bmatrix}\!} = {\begin{bmatrix}
  g(\bm{Z}(t);\bm{\theta}_g)f(\bm{H}(t);\bm{\theta}_f) \frac{d\bm{X}(t)}{dt} \\
  f(\bm{H}(t);\bm{\theta}_f) \frac{d\bm{X}(t)}{dt}\\
  \end{bmatrix}\!},
\end{align}where the initial values $\bm{Z}(0)$ and $\bm{H}(0)$ are generated in the aforementioned ways. We then train the parameters of the initial value generation layer, the CDE functions, including the node-embedding matrix $\bm{E}$, and the output layer. From $\bm{z}^{(v)}(T)$, i.e., the $v$-th row of $\bm{Z}(T)$, the following output layer produces $\hat{\bm{y}}^{(v)}$.
\begin{align}
    \hat{\bm{y}}^{(v)} = \bm{z}^{(v)}(T)\bm{W}_{output} + \bm{b}_{output}, \label{eq:output}
\end{align}where $\bm{W}_{output} \in \mathbb{R}^{\dim(\bm{z}^{(v)}(T)) \rightarrow S \times M}$ and $\bm{b}_{output} \in \mathbb{R}^{S \times M}$ are a trainable weight and a bias of the output layer. We use the following $L^1$ loss as the training objective, which is defined as:
\begin{align}
\mathcal{L} = \frac{\sum_{\tau \in \mathcal{T}}\sum_{v \in \mathcal{V}} \|\bm{y}^{(\tau,v)} - \hat{\bm{y}}^{(\tau,v)}\|_1}{|\mathcal{V}| \times |\mathcal{T}|}, \label{eq:loss}
\end{align}where $\mathcal{T}$ is a training set, $\tau$ is a training sample, and $\bm{y}^{(\tau,v)}$ is the ground-truth of node $v$ in $\tau$. We also use the standard $L^2$ regularization of the parameters, i.e., weight decay.

The well-posedness\footnote{A well-posed problem means i) its solution uniquely exists, and ii) its solution continuously changes as input data changes.} of NCDEs was already proved in \cite[Theorem 1.3]{lyons2007differential} under the mild condition of the Lipschitz continuity. We show that our NCDE layers are also well-posed problems. Almost all activations, such as ReLU, Leaky ReLU, SoftPlus, Tanh, Sigmoid, ArcTan, and Softsign, have a Lipschitz constant of 1. Other common neural network layers, such as dropout, batch normalization and other pooling methods, have explicit Lipschitz constant values. Therefore, the Lipschitz continuity of $f$ and $g$ can be fulfilled in our case. Therefore, it is a well-posed training problem. Therefore, our training algorithm solves a well-posed problem so its training process is stable in practice.

\section{Experiments}
We describe our experimental environments and results. We conduct experiments with time-series forecasting. Our software and hardware environments are as follows: \textsc{Ubuntu} 18.04 LTS, \textsc{Python} 3.9.5, \textsc{Numpy} 1.20.3, \textsc{Scipy} 1.7, \textsc{Matplotlib} 3.3.1, \textsc{torchdiffeq} 0.2.2, \textsc{PyTorch} 1.9.0, \textsc{CUDA} 11.4, and \textsc{NVIDIA} Driver 470.42, i9 CPU, and \textsc{NVIDIA RTX A6000}. We use 6 datasets and 20 baseline models, which is one of the largest scale experiments in the field of traffic forecasting. For additional figures, tables, and best hyperparameter settings are in Appendix.
\subsection{Datasets}
In the experiment, we use six real-world traffic datasets, namely PeMSD7(M), PeMSD7(L), PeMS03, PeMS04, PeMS07, and PeMS08, which were collected by California Performance of Transportation (PeMS)~\cite{chen2001freeway} in real-time every 30 second and widely used in the previous studies~\cite{bing2018stgcn,guo2019astgcn,fang2021STODE,chen2021ZGCNET,song2020stsgcn}. More details of the datasets are in Table~\ref{tab:dataset}. We note that they contain different types of values: i) the number of vehicles, or ii) velocity.

\begin{table}[t]
\setlength{\tabcolsep}{2pt}
    \centering
    \small
    \begin{tabular}{ccccc}
    \hline
        Dataset     & $|\mathcal{V}|$  & Time Steps& Time Range & Type \\ \hline
        PeMSD3      & 358       & 26,208    & 09/2018 - 11/2018 & Volume \\
        PeMSD4      & 307       & 16,992    & 01/2018 - 02/2018 & Volume \\
        PeMSD7      & 883       & 28,224    & 05/2017 - 08/2017 & Volume \\ 
        PeMSD8      & 170       & 17,856    & 07/2016 - 08/2016 & Volume \\ 
        PeMSD7(M)   & 228       & 12,672    & 05/2012 - 06/2012 & Velocity\\
        PeMSD7(L)   & 1,026     & 12,672    & 05/2012 - 06/2012 & Velocity\\
    \hline
    \end{tabular}
    \caption{The summary of the datasets used in our work. We predict either traffic volume (i.e., \# of vehicles) or velocity.}
    \label{tab:dataset}
\end{table}

\subsection{Experimental Settings}
The datasets are already split with a ratio of 6:2:2 into training, validating, and testing sets. In these datasets, the interval between two consecutive time-points is 5 minutes. All existing papers, including our paper, use the forecasting settings of $S=12$ and $M=1$ after reading past 12 graph snapshots, i.e., $N=11$ --- note that the graph snapshot index $i$ starts from $0$. In short, we conduct a 12-sequence-to-12-sequence forecasting, which is the standard benchmark setting in this domain.

We use the mean absolute error (MAE), the mean absolute percentage error (MAPE), and the root mean squared error (RMSE) to measure the performance of different models.
\subsubsection{Baselines}
We compare our proposed STG-NCDE with the following baseline models in conjunction with the previous models we introduced in the related work section --- in total, we use 20 baseline models:
\begin{compactenum}
    \item HA~\cite{hamilton2020time} uses the average value of the last 12 times slices to predict the next value.
    \item ARIMA is a statistical model of time series analysis.
    \item VAR~\cite{hamilton2020time} is a time series model that captures spatial correlations among all traffic series.
    \item TCN~\cite{BaiTCN2018} consists of a stack of causal convolutional layers with exponentially enlarged dilation factors.
    \item FC-LSTM~\cite{sutskever2014sequence} is LSTM with fully connected hidden unit.
    \item GRU-ED~\cite{cho2014grued} is an GRU-based baseline and utilize the encoder-decoder framework for multi-step time series prediction.
    \item DSANet~\cite{Huang2019DSANet} is a correlated time series prediction model using CNN networks and self-attention mechanism for spatial correlations.
\end{compactenum}

\subsubsection{Hyperparameters}
For our method, we test with the following hyperparameter configurations: we train for 200 epochs using the Adam optimizer, with a batch size of 64 on all datasets. The two dimensionalities of $\dim(\bm{h}^{(v)})$ and $\dim(\bm{z}^{(v)})$ are \{32, 64, 128, 256\}, the node embedding size $C$ is from 1 to 10, and the number of $K$ in Eq.~\eqref{eq:fun_f} is in \{1, 2, 3\}.  The learning rate in all methods is in \{\num{1e-2}, \num{5e-3}, \num{1e-3}, \num{5e-4}, \num{1e-4}\} and the weight decay coefficient is in \{\num{1e-4}, \num{1e-3}, \num{1e-2}\}. An early stop strategy with a patience of 15 iterations on the validation dataset is used. The best hyperparameters are in Appendix for reproducibility. For baselines, we run their codes with a hyperparameter search process based on their recommended configurations if their accuracy is not known for a dataset. If known, we use their officially reported accuracy.

\begin{table}[t]
\small
    \setlength{\tabcolsep}{2pt}
    \centering
    \begin{tabular}{c ccc}
    \hline
        Model                &   MAE                 &   RMSE                &  MAPE\\\hline
        STGCN                & 14.88 (117.0\%)       & 24.22 (113.6\%)       & 12.30 (121.8\%)\\
        DCRNN                & 14.90 (117.1\%)       & 24.04 (112.7\%)       & 12.75 (126.1\%)\\
        GraphWaveNet         & 15.94 (125.3\%)       & 26.22 (122.9\%)       & 12.96 (128.2\%) \\
        ASTGCN(r)            & 14.86 (116.9\%)       & 23.95 (112.3\%)       & 12.25 (121.3\%) \\
        STSGCN               & 14.45 (113.5\%)       & 23.58 (110.5\%)       & 11.42 (113.0\%) \\
        AGCRN                & 13.32 (104.7\%)       & 22.29 (104.5\%)       & 10.37 (102.7\%) \\
        STFGNN               & 13.92 (109.5\%)       & 22.57 (105.8\%)       & 11.30 (111.9\%) \\
        STGODE               & 13.56 (106.6\%)       & 22.37 (104.8\%)       & 10.77 (106.6\%) \\
        Z-GCNETs             & 13.22 (104.0\%)       & 21.92 (102.7\%)       & 10.44 (103.4\%) \\\hline
        \textbf{STG-NCDE}    & 12.72 (100.0\%)       & 21.33 (100.0\%)       & 10.10 (100.0\%)\\
    \hline
    \end{tabular}
    \caption{The average error of some selected highly performing models across all the six datasets. Inside the parentheses, we show their performance relative to our method.}
    \label{tab:average}
\end{table}

\begin{table*}[t]
    \centering
    \setlength{\tabcolsep}{4pt}
    \small
    \begin{tabular}{c ccc c ccc c ccc c ccc}
        \hline
        \multirow{2}{*}{Model}  & \multicolumn{3}{c}{PeMSD3}    && \multicolumn{3}{c}{PeMSD4}      && \multicolumn{3}{c}{PeMSD7}      && \multicolumn{3}{c}{PeMSD8}\\\cline{2-4} \cline{6-8} \cline{10-12} \cline{14-16}
                                & MAE & RMSE & MAPE             && MAE & RMSE & MAPE               && MAE & RMSE & MAPE               && MAE & RMSE & MAPE               \\ \hline
        HA                      & 31.58 & 52.39 & 33.78\%       && 38.03 & 59.24 & 27.88\%         && 45.12 & 65.64 & 24.51\%         && 34.86 & 59.24 & 27.88\%         \\       
        ARIMA                   & 35.41 & 47.59 & 33.78\%       && 33.73 & 48.80 & 24.18\%         && 38.17 & 59.27 & 19.46\%         && 31.09 & 44.32 & 22.73\%\\        
        VAR                     & 23.65 & 38.26 & 24.51\%       && 24.54 & 38.61 & 17.24\%         && 50.22 & 75.63 & 32.22\%         && 19.19 & 29.81 & 13.10\%         \\                     
        FC-LSTM                 & 21.33 & 35.11 & 23.33\%       && 26.77 & 40.65 & 18.23\%         && 29.98 & 45.94 & 13.20\%         && 23.09 & 35.17 & 14.99\%         \\ 
        TCN                     & 19.32 & 33.55 & 19.93\%       && 23.22 & 37.26 & 15.59\%         && 32.72 & 42.23 & 14.26\%         && 22.72 & 35.79 & 14.03\%         \\  
        TCN(w/o causal)         & 18.87 & 32.24 & 18.63\%       && 22.81 & 36.87 & 14.31\%         && 30.53 & 41.02 & 13.88\%         && 21.42 & 34.03 & 13.09\%         \\  
        GRU-ED                  & 19.12 & 32.85 & 19.31\%       && 23.68 & 39.27 & 16.44\%         && 27.66 & 43.49 & 12.20\%         && 22.00 & 36.22 & 13.33\%         \\       
        DSANet                  & 21.29 & 34.55 & 23.21\%       && 22.79 & 35.77 & 16.03\%         && 31.36 & 49.11 & 14.43\%         && 17.14 & 26.96 & 11.32\%         \\       
        STGCN                   & 17.55 & 30.42 & 17.34\%       && 21.16 & 34.89 & 13.83\%         && 25.33 & 39.34 & 11.21\%         && 17.50 & 27.09 & 11.29\%         \\ 
        DCRNN                   & 17.99 & 30.31 & 18.34\%       && 21.22 & 33.44 & 14.17\%         && 25.22 & 38.61 & 11.82\%         && 16.82 & 26.36 & 10.92\%         \\ 
        GraphWaveNet            & 19.12 & 32.77 & 18.89\%       && 24.89 & 39.66 & 17.29\%         && 26.39 & 41.50 & 11.97\%         && 18.28 & 30.05 & 12.15\%         \\ 
        ASTGCN(r)               & 17.34 & 29.56 & 17.21\%       && 22.93 & 35.22 & 16.56\%         && 24.01 & 37.87 & 10.73\%         && 18.25 & 28.06 & 11.64\%         \\ 
        MSTGCN                  & 19.54 & 31.93 & 23.86\%       && 23.96 & 37.21 & 14.33\%         && 29.00 & 43.73 & 14.30\%         && 19.00 & 29.15 & 12.38\%         \\       
        STG2Seq                 & 19.03 & 29.83 & 21.55\%       && 25.20 & 38.48 & 18.77\%         && 32.77 & 47.16 & 20.16\%         && 20.17 & 30.71 & 17.32\%         \\ 
        LSGCN                   & 17.94 & 29.85 & 16.98\%       && 21.53 & 33.86 & 13.18\%         && 27.31 & 41.46 & 11.98\%         && 17.73 & 26.76 & 11.20\%         \\       
        STSGCN                  & 17.48 & 29.21 & 16.78\%       && 21.19 & 33.65 & 13.90\%         && 24.26 & 39.03 & 10.21\%         && 17.13 & 26.80 & 10.96\%         \\        
        AGCRN                   & \underline{15.98} & 28.25 & \underline{15.23}\%       && 19.83 & 32.26 & 12.97\%   && 22.37 & 36.55 &  \underline{9.12}\%    && 15.95 & 25.22 & 10.09\%         \\
        STFGNN                  & 16.77 & 28.34 & 16.30\%       && 20.48 & 32.51 & 16.77\%         && 23.46 & 36.60 &  9.21\%         && 16.94 & 26.25 & 10.60\%         \\        
        STGODE                   & 16.50 & \underline{27.84} & 16.69\%       && 20.84 & 32.82 & 13.77\%         && 22.59 & 37.54 & 10.14\% && 16.81 & 25.97 & 10.62\%         \\
        Z-GCNETs                & 16.64 & 28.15 & 16.39\%       && \underline{19.50} & \underline{31.61} & \underline{12.78}\%        && \underline{21.77} & \underline{35.17} & 9.25\%   && \underline{15.76} & \underline{25.11} & \underline{10.01}\% \\
        \hline
        \textbf{STG-NCDE}       & \textbf{15.57} & \textbf{27.09}    &   \textbf{15.06}\% && \textbf{19.21}    & \textbf{31.09}    &  \textbf{12.76}\%   && \textbf{20.53} & \textbf{33.84} & \textbf{8.80}\%    && \textbf{15.45} & \textbf{24.81} &  \textbf{9.92}\% \\
        \textbf{Only temporal}  & 20.44 & 32.82 & 20.03\%       && 26.31 & 40.97 & 17.95\%          && 28.77 & 44.39 & 12.60\%         && 20.83 & 32.55 & 13.01\% \\
        \textbf{Only spatial}   & 15.92 & 27.17 & 15.14\%       && 19.86 & 31.92 & 13.35\%          && 21.72 & 34.73 &  9.24\%         && 17.58 & 27,76 & 11.27\% \\
        \hline
    \end{tabular}
    \caption{Forecasting error on PeMSD3, PeMSD4, PeMSD7 and PeMSD8}
    \label{tab:main_exp}
\end{table*}

\subsection{Experimental Results}
Tables~\ref{tab:main_exp} and~\ref{tab:main_exp_2} present the detailed prediction performance. Overall, our proposed method, STG-NCDE, clearly marks the best average accuracy as summarized in Table~\ref{tab:average}. For each notable model, we list its average MAE/RMSE/MAPE from the six datasets. Inside the parentheses, we also show the relative accuracy in comparison with our method. For instance, STGCN shows an MAE that is 17.0\% worse than that of our method. All existing methods show worse errors in all metrics than our method (by large margins for many baselines).

We now describe experimental results in each dataset. STG-NCDE shows the best accuracy in all cases, followed by Z-GCNETs, AGCRN, STGODE and so on. There are no existing methods that are as stable as STG-NCDE. For instance, STGODE shows reasonably low errors in many cases, e.g., an RMSE of 27.84 in PeMSD3 by STGODE, which is the second best result vs. 27.09 by STG-NCDE. However, it is outperformed by AGCRN and Z-GCNETs for PeMSD7. Only our method, STG-NCDE, shows reliable predictions in all cases.

\begin{table}[!ht]
    \centering
    \setlength{\tabcolsep}{2pt}
    \small
    \begin{tabular}{ccccccc c ccc}
        \hline
        \multirow{2}{*}{Model} & \multicolumn{3}{c}{PeMSD7(M)}   && \multicolumn{3}{c}{PeMSD7(L)}     \\\cline{2-4} \cline{6-8}
                                & MAE  & RMSE  & MAPE           && MAE  & RMSE  & MAPE               \\ \hline
        HA                      & 4.59 &  8.63 & 14.35\%        && 4.84 &  9.03 & 14.90\%         \\       
        ARIMA                   & 7.27 & 13.20 & 15.38\%        && 7.51 & 12.39 & 15.83\%         \\        
        VAR                     & 4.25 &  7.61 & 10.28\%        && 4.45 &  8.09 & 11.62\%         \\                     
        FC-LSTM                 & 4.16 &  7.51 & 10.10\%        && 4.66 &  8.20 & 11.69\%         \\               
        TCN                     & 4.36 &  7.20 &  9.71\%        && 4.05 &  7.29 & 10.43\%         \\               
        TCN(w/o causal)         & 4.43 &  7.53 &  9.44\%        && 4.58 &  7.77 & 11.53\%         \\               
        GRU-ED                  & 4.78 &  9.05 & 12.66\%        && 3.98 &  7.71 & 10.22\%         \\       
        DSANet                  & 3.52 &  6.98 &  8.78\%        && 3.66 &  7.20 &  9.02\%         \\       
        STGCN                   & 3.86 &  6.79 & 10.06\%        && 3.89 &  6.83 & 10.09\%         \\       
        DCRNN                   & 3.83 &  7.18 &  9.81\%        && 4.33 &  8.33 & 11.41\%         \\       
        GraphWaveNet            & 3.19 &  6.24 &  8.02\%        && 3.75 &  7.09 &  9.41\%         \\       
        ASTGCN(r)               & 3.14 &  6.18 &  8.12\%        && 3.51 &  6.81 &  9.24\%         \\       
        MSTGCN                  & 3.54 &  6.14 &  9.00\%        && 3.58 &  6.43 &  9.01\%         \\       
        STG2Seq                 & 3.48 &  6.51 &  8.95\%        && 3.78 &  7.12 &  9.50\%         \\       
        LSGCN                   & 3.05 &  5.98 &  7.62\%        && 3.49 &  6.55 &  8.77\%          \\       
        STSGCN                  & 3.01 &  5.93 &  7.55\%        && 3.61 &  6.88 &  9.13\%          \\        
        AGCRN                   & 2.79 & \underline{5.54} & 7.02\%   && 2.99 & 5.92 & 7.59\%        \\
        STFGNN                  & 2.90 & 5.79 & 7.23\%          && 2.99 & 5.91 & 7.69\%          \\        
        STGODE                  & 2.97 & 5.66 & 7.36\%         && 3.22 & 5.98 & 7.94\%        \\
        Z-GCNETs                & \underline{2.75} & 5.62 & \underline{6.89}\%  && \underline{2.91} & \underline{5.83} & \underline{7.33}                    \%  \\
        \hline
        \textbf{STG-NCDE}           & \textbf{2.68}    & \textbf{5.39}    &  \textbf{6.76}\% && \textbf{2.87} & \textbf{5.76} &  \textbf{7.31}\%\\
        \textbf{Only temporal}  & 3.34 & 6.68 & 8.41\%          && 3.54 & 7.03 & 8.89\%  \\
        \textbf{Only spatial}   & 2.77 & 5.40 & 7.00\%          && 2.99 & 5.85 & 7.60\%  \\
        \hline
    \end{tabular}
    \caption{Forecasting error on PeMSD7(M) and PeMSD7(L)}
    \label{tab:main_exp_2}
\end{table}

\begin{figure}[t]
    \centering
    \subfigure[Node 111 in PeMSD4]{\includegraphics[width=0.48\columnwidth]{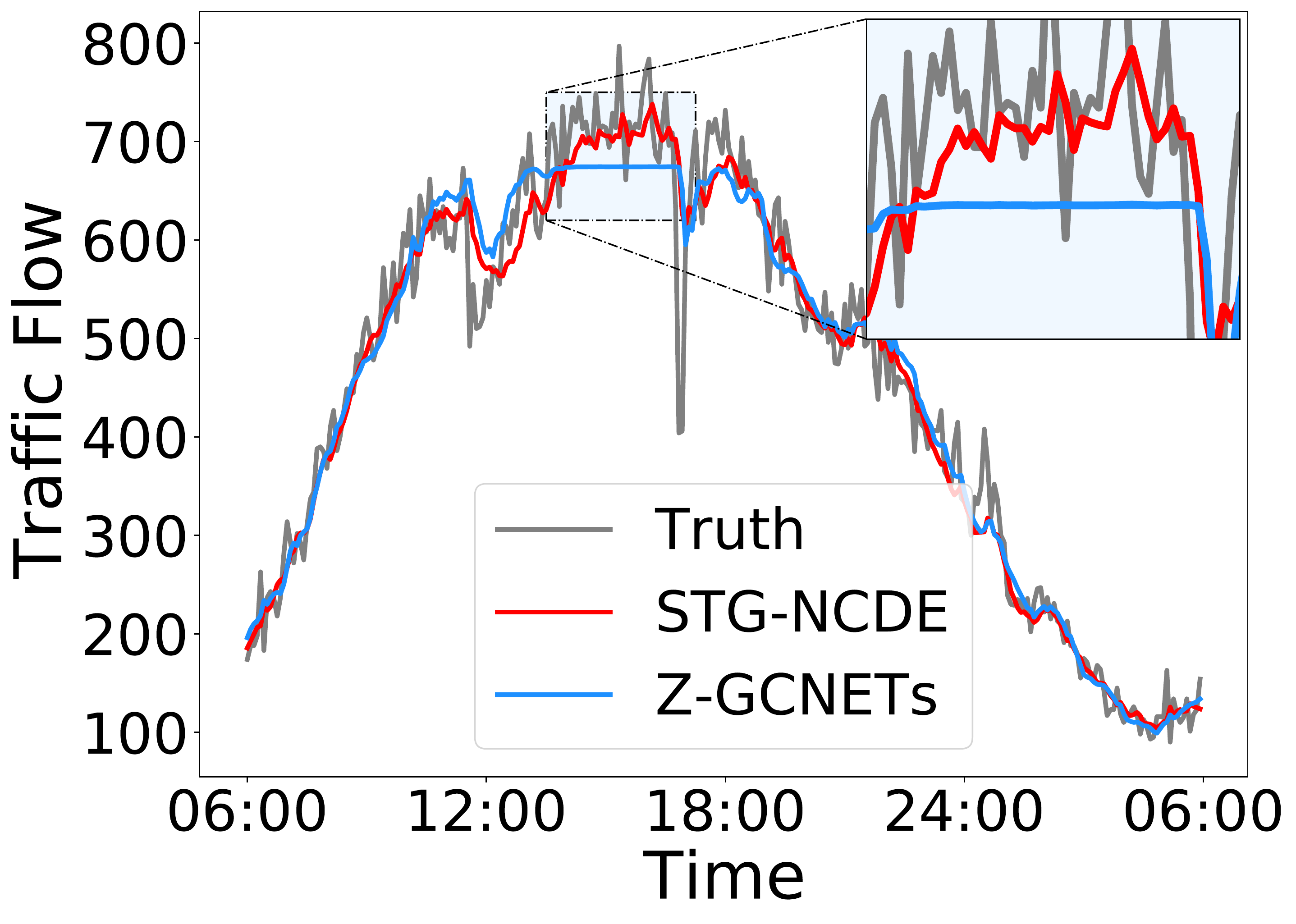}}
    \subfigure[Node 261 in PeMSD4]{\includegraphics[width=0.48\columnwidth]{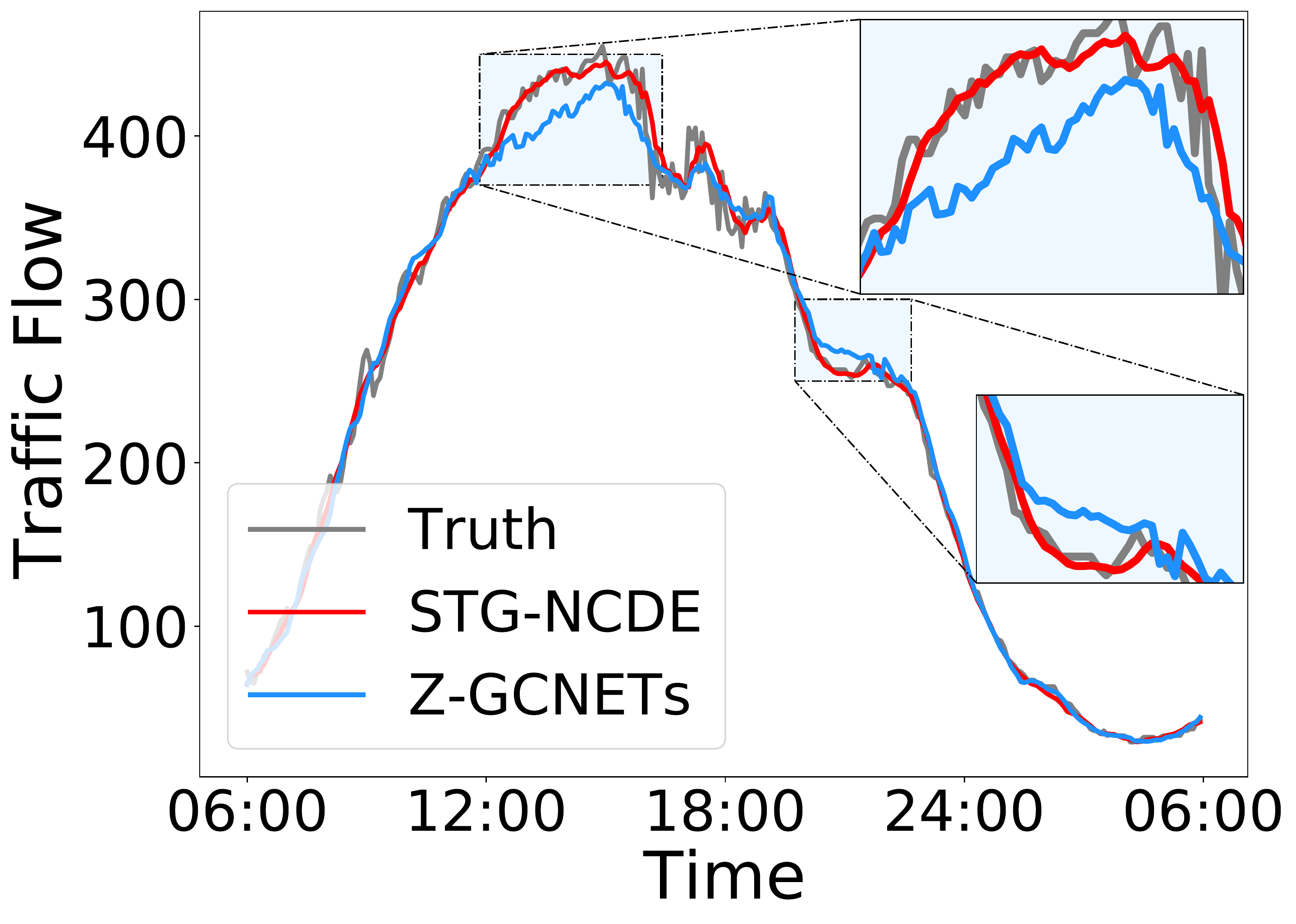}}
    \subfigure[Node 9 in PeMSD8]{\includegraphics[width=0.48\columnwidth]{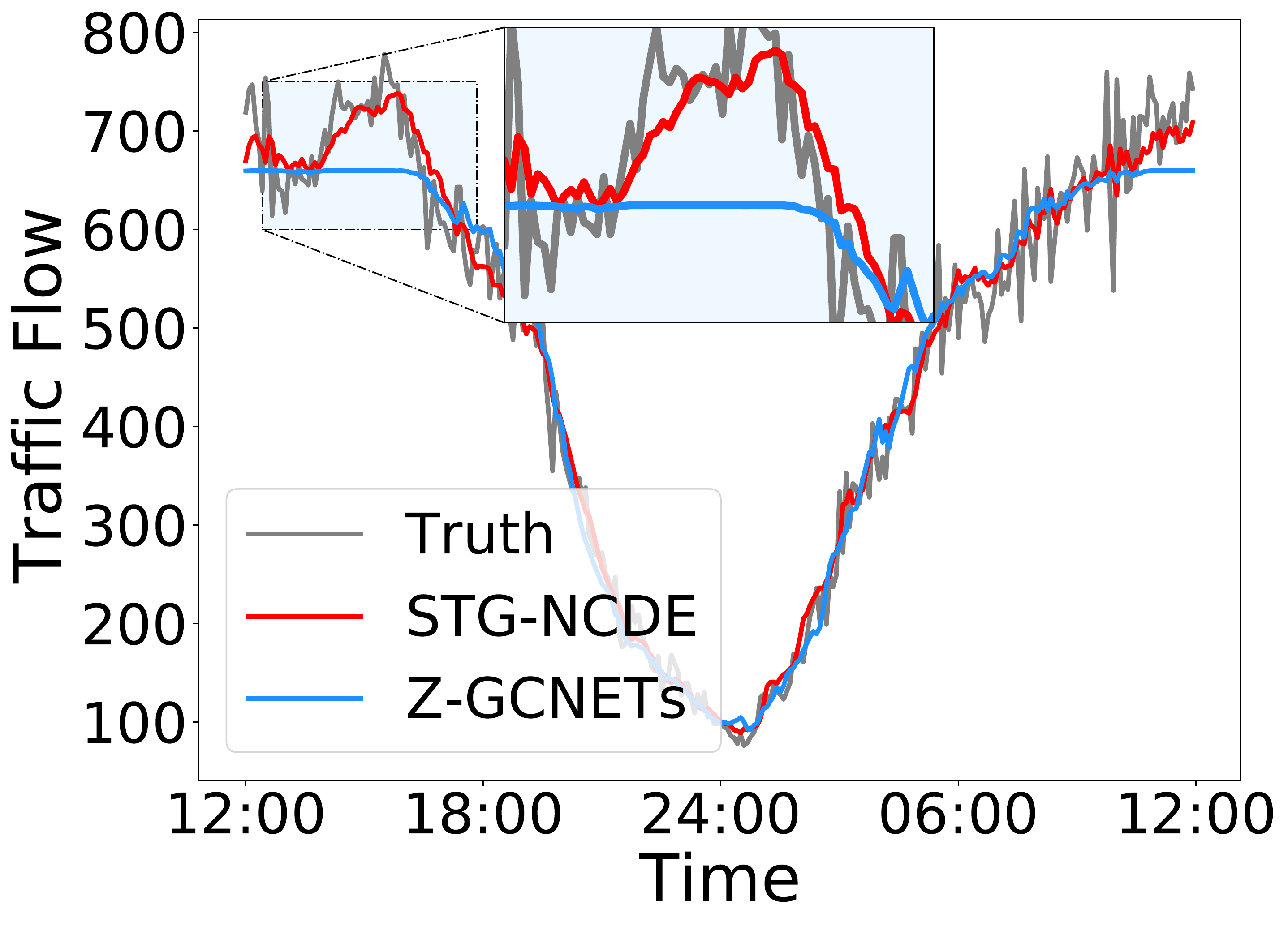}}
    \subfigure[Node 112 in PeMSD8]{\includegraphics[width=0.48\columnwidth]{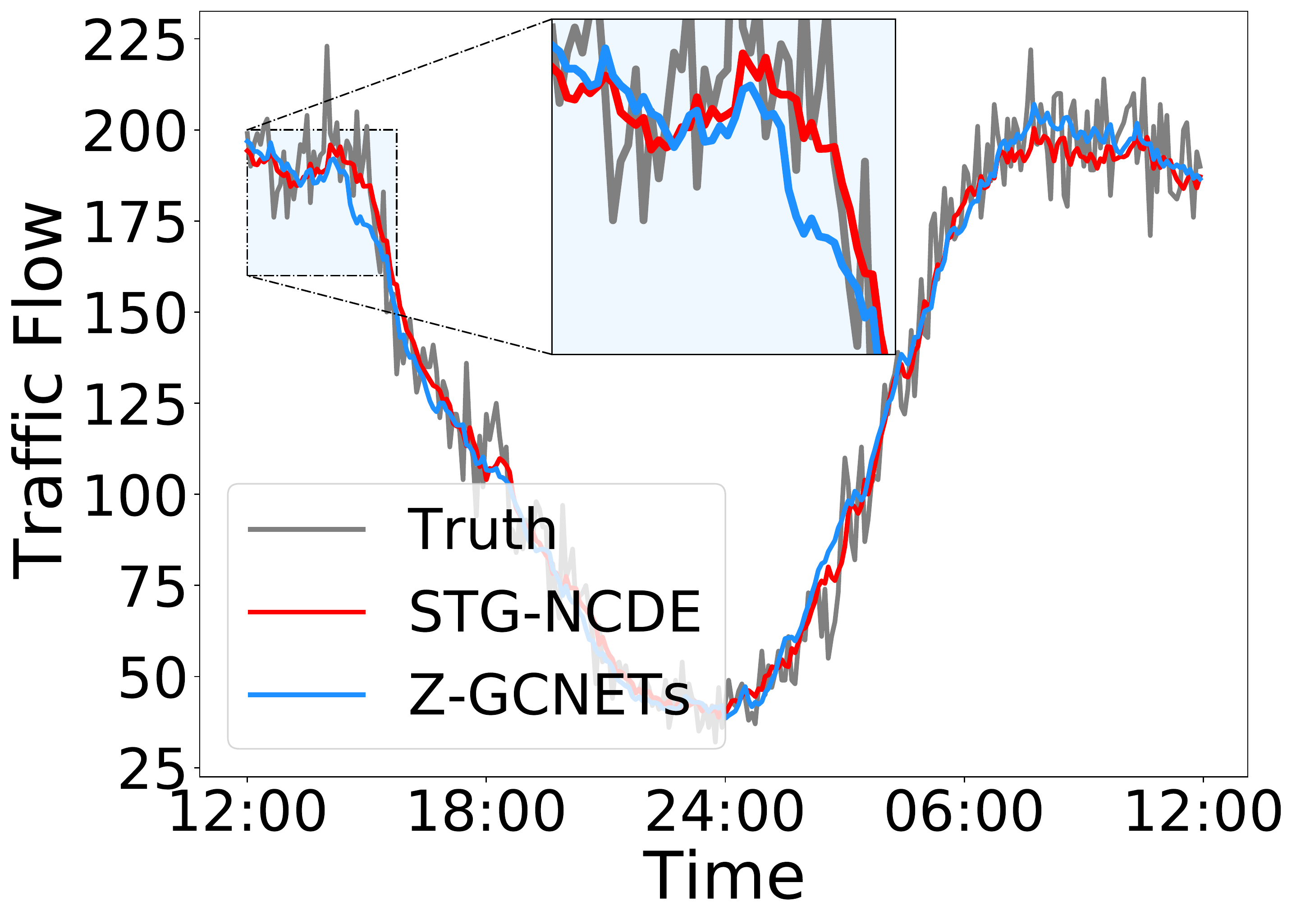}}
    \caption{Traffic forecasting visualization. More visualizations in other datasets are in Appendix.}   \label{fig:pred_vis}

    \label{fig:pred_vis_pemsd8}
\end{figure}

We also visualize the ground-truth and the predicted curves by our method and Z-GCNETs in Fig.~\ref{fig:pred_vis}. Node 111 and 261 (resp. Node 9 and 112) are two of the highest traffic areas in PeMSD4 (resp. PeMSD8). Since Z-GCNETs shows reasonable performance, its predicted curve is similar to that of our method in many time-points. As highlighted with boxes, however, our method shows much more accurate predictions for challenging cases. In particular, our method significantly outperforms Z-GCNETs for the highlighted time-points for Node 111 in PeMSD4 and Node 9 in PeMSD8, for which Z-GCNETs shows nonsensical predictions, i.e., the prediction curves are straight. 

\subsection{Ablation, Sensitivity, and Additional Studies}\label{sec:abl}
\paragraph{Ablation study} As ablation study models, we define the following two models: i) the first ablation model has only the temporal processing part, i.e., Eq.~\eqref{eq:type1-2}, and ii) the second ablation model has only the spatial processing part which can be written as follows:
\begin{align}\begin{split}
\bm{Z}(T) = \bm{Z}(0) + \int_{0}^{T} g(\bm{Z}(t);\bm{\theta}_g) \frac{d\bm{X}(t)}{dt} dt,\label{eq:abl}
\end{split}\end{align}where the trajectory $\bm{Z}(t)$ over time is controlled by $\bm{X}(t)$. We accordingly change the model architecture for this ablation study model. The first (resp. second) model is denoted as ``Only temporal'' (resp. ``Only spatial'') in the tables.

In all cases, the ablation study model only with the spatial processing significantly outperforms that only with the temporal processing, e.g., an RMSE of 15.92 in PeMSD3 by the spatial processing vs. 20.44 by the temporal processing. However, STG-NCDE, which utilizes both the temporal and the spatial processing, outperforms them. This shows that we need both of them to achieve the best model accuracy.

\begin{figure}[!t]
    \centering
    \subfigure[Training curve in PeMSD7]{\includegraphics[width=0.48\columnwidth]{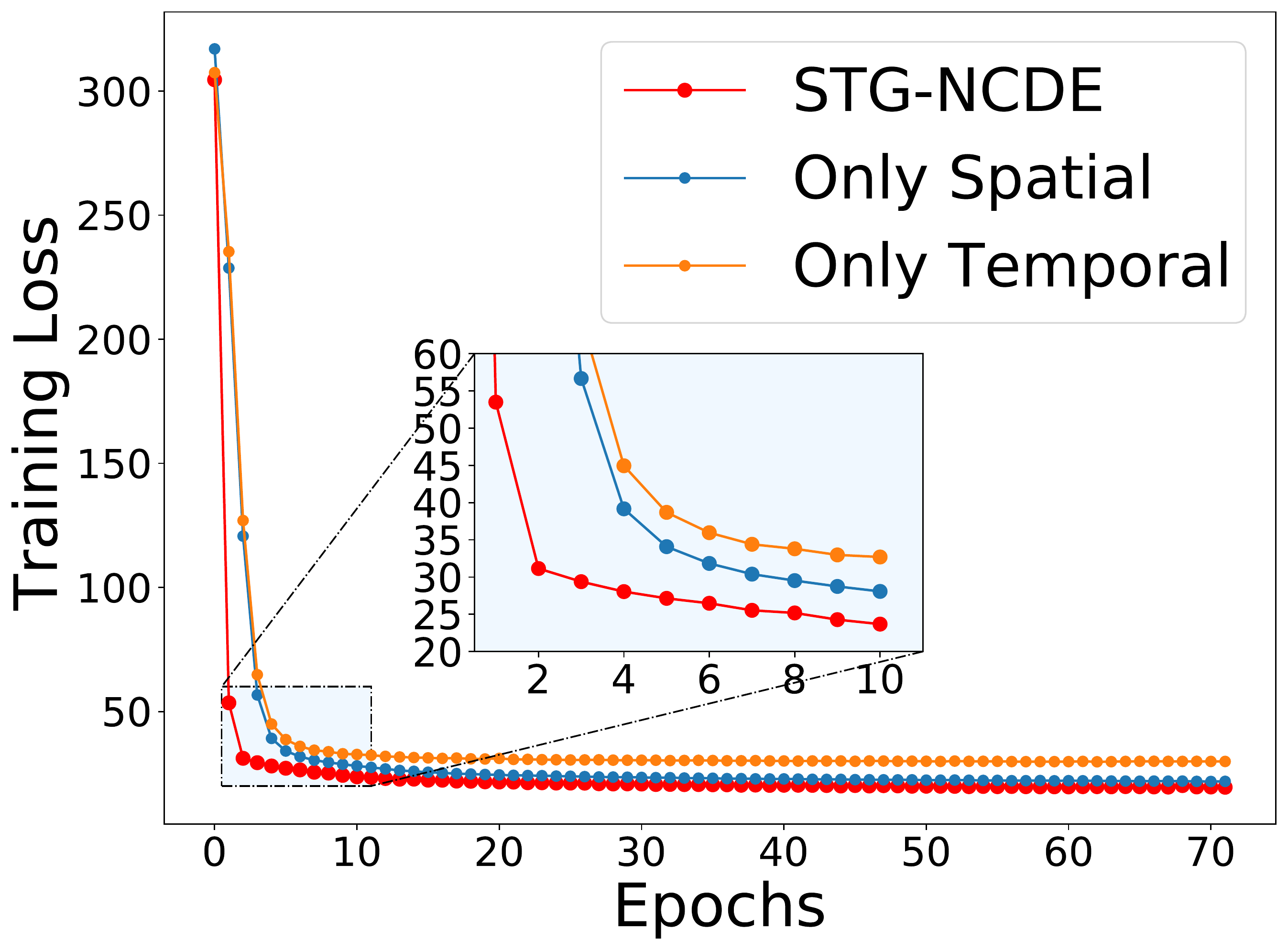}}
    \subfigure[Sensitivity to $C$ in PeMSD7]{\includegraphics[width=0.48\columnwidth]{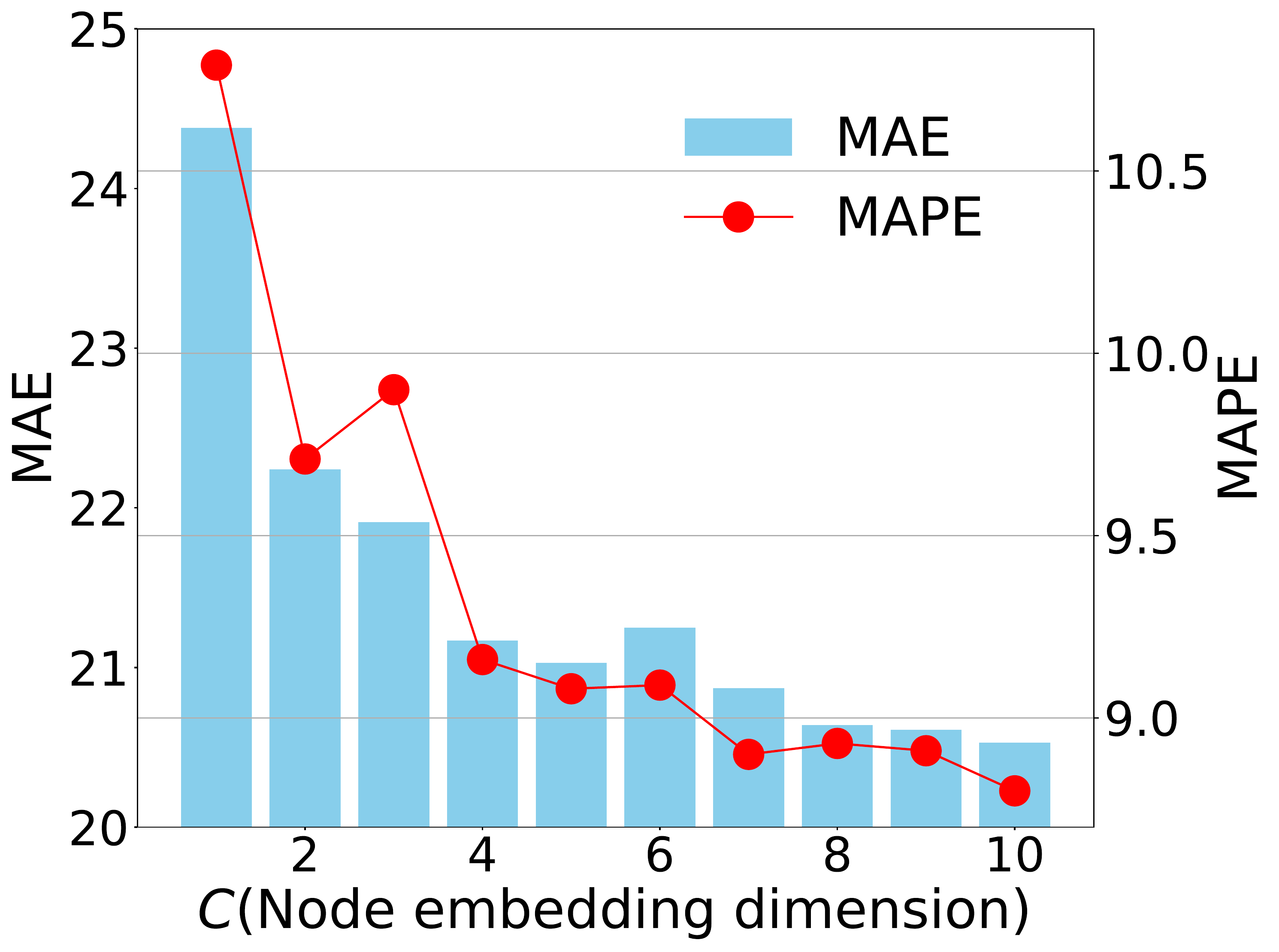}}
    \caption{Training curve and sensitivity analysis. More results in other datasets are in Appendix.}
    \label{fig:loss_pemsd7}
\end{figure}

In Fig.~\ref{fig:loss_pemsd7} (a), we also compare their training curves in PeMSD7. STG-NODE's loss curve is stabilized after the second epoch whereas the other two ablation models require longer time until their loss curves are stabilized.

\begin{figure}[!t]
    \centering
    \subfigure[MAE on PeMSD7]{\includegraphics[width=0.48\columnwidth]{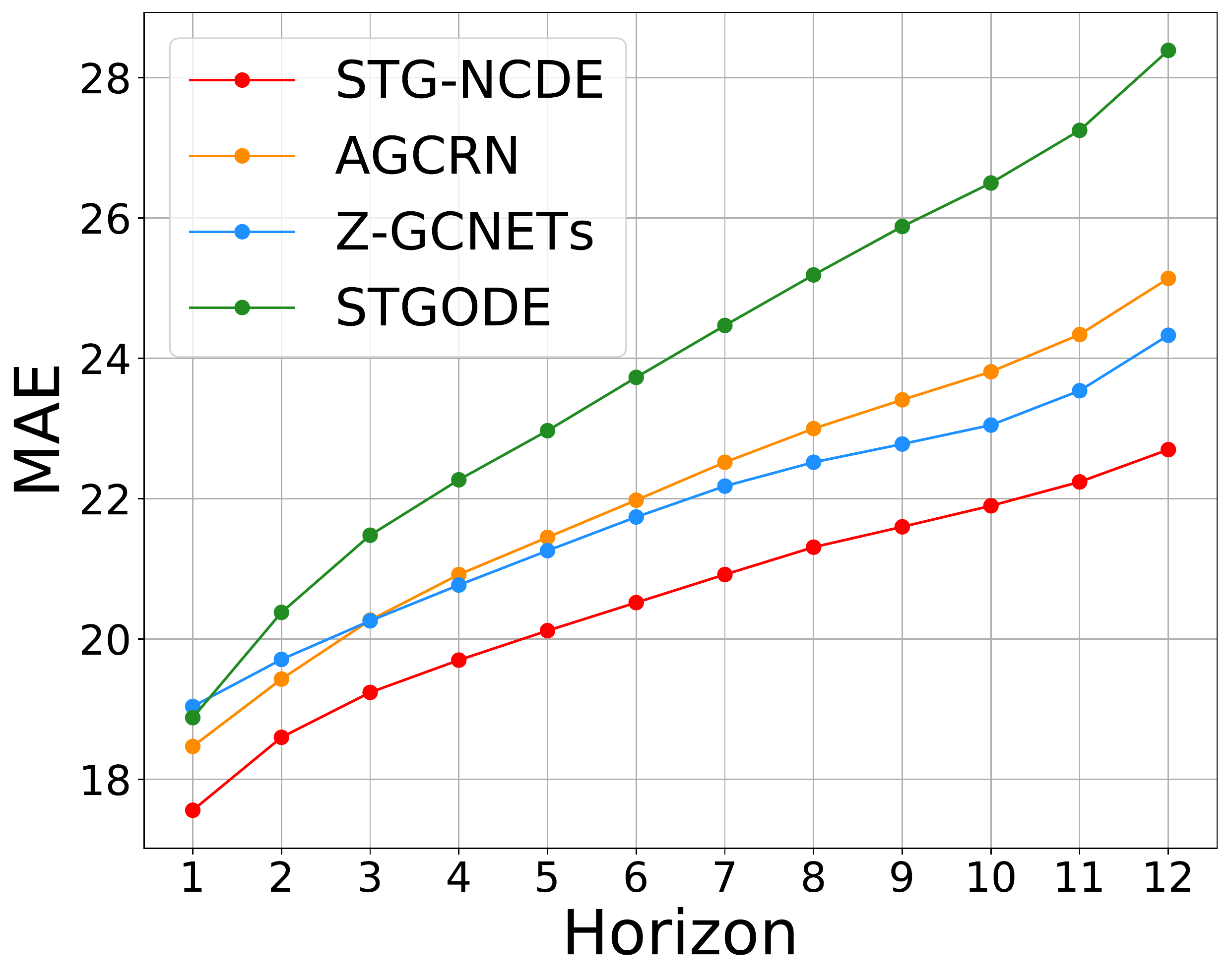}}
    \subfigure[MAPE on PeMSD7]{\includegraphics[width=0.48\columnwidth]{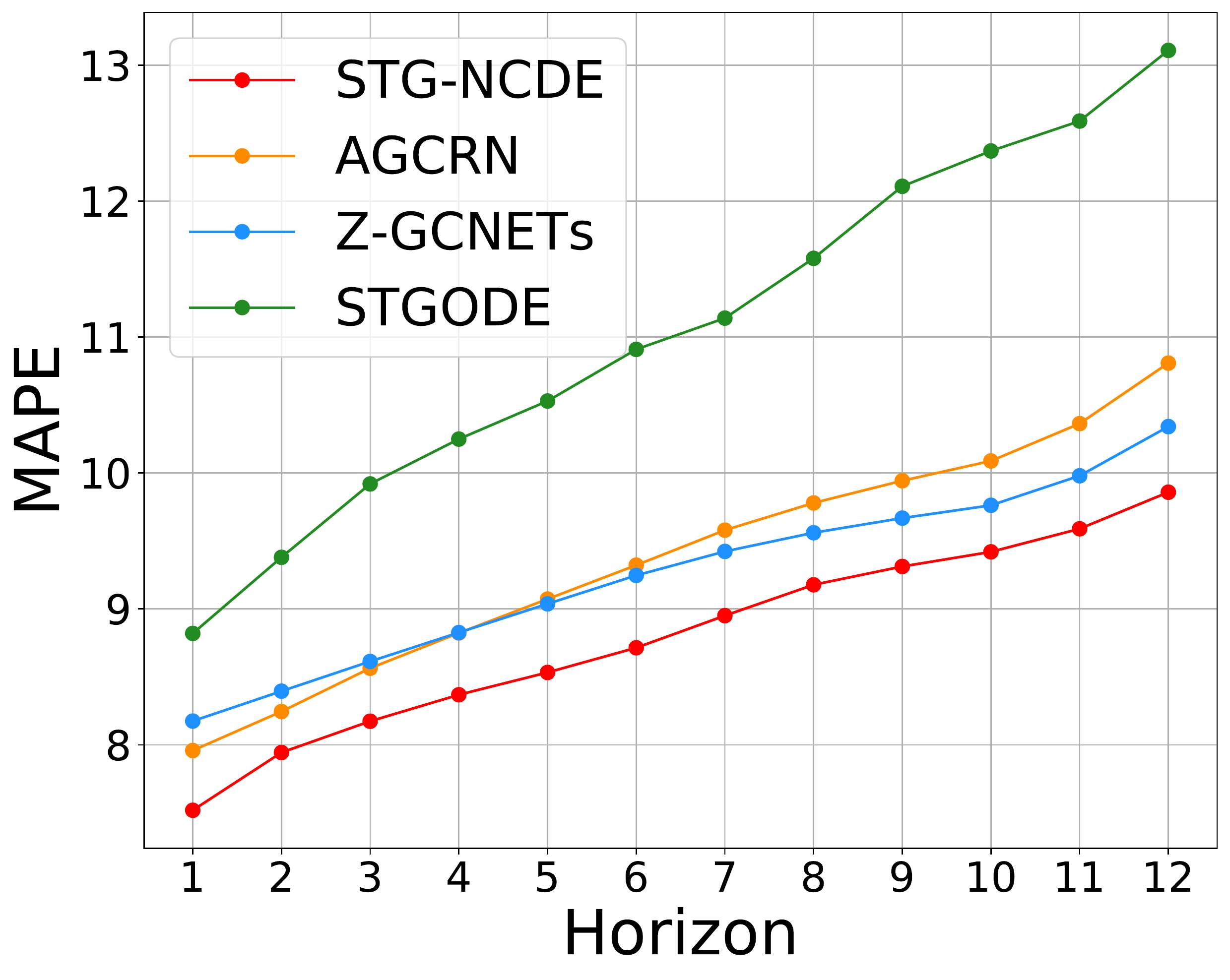}}
    \subfigure[MAE on PeMSD8]{\includegraphics[width=0.48\columnwidth]{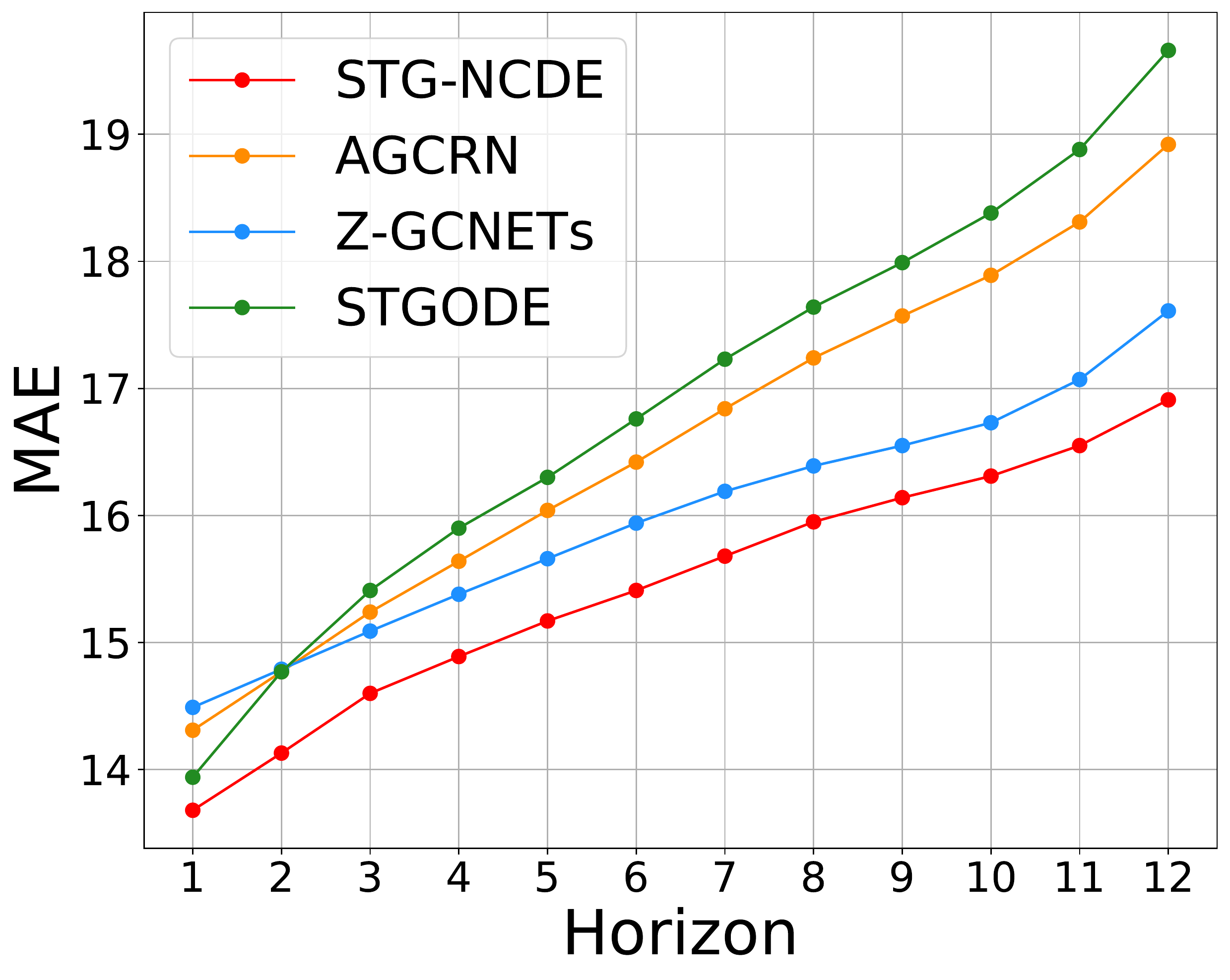}}
    \subfigure[MAPE on PeMSD8]{\includegraphics[width=0.48\columnwidth]{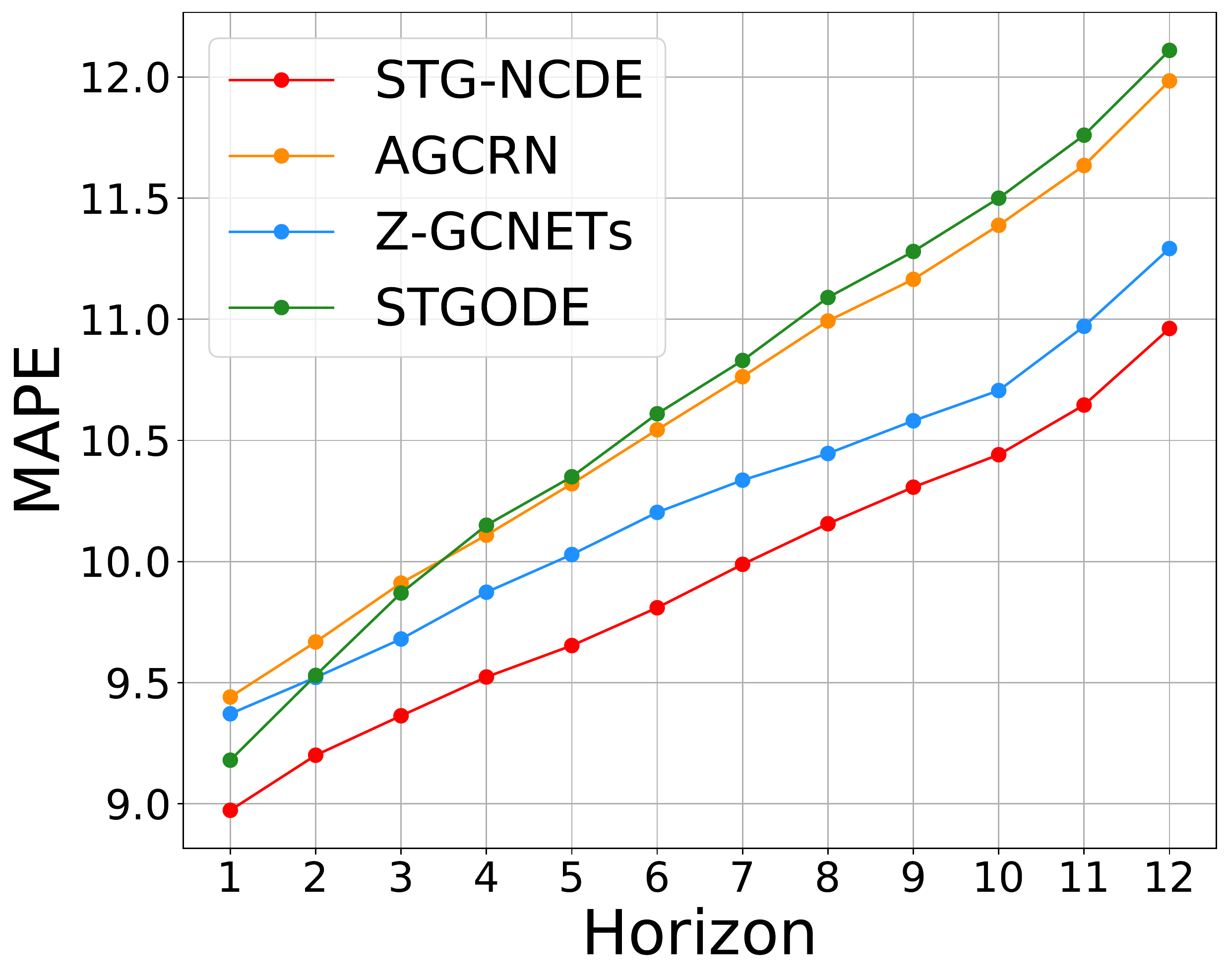}}
    \caption{Prediction error at each horizon. More results in other datasets are in Appendix.}
    \label{fig:horizon}
\end{figure}
\paragraph{Sensitivity to $C$} Fig.~\ref{fig:loss_pemsd7} (b) shows the MAE and MAPE by varying the node embedding size $C$. The two error metrics are stabilized after $C=7$. With $C=10$, we can achieve the best accuracy.

\paragraph{Error for each horizon} In our notation, $S$ denotes the length of forecasting, i.e., the number of forecasting horizons. Since the benchmark dataset has a setting of $S=12$, we show the model error for each forecasting horizon in Fig.~\ref{fig:horizon}. It is obvious that the error levels show a high correlation to $S$. For all horizons, STG-NCDE shows smaller errors than other baselines.

\paragraph{Irregular traffic forecasting} In reality, traffic sensors can be damaged and we cannot collect data in some areas for a certain amount of time. In order to reflect this situation, we randomly drop 10\% to 50\% of sensing values for each node independently. Since NCDEs are able to consider irregular time-series by the design, STG-NCDE is also able to do it without any changes on its model design, which is one of the  most distinguishable points in comparison with existing baselines. Tables~\ref{tbl:missing_pemsd4} and~\ref{tbl:missing_pemsd8} summarize its results. In comparison with the results in Table~\ref{tab:main_exp}, our model's performance is not significantly degraded. We note that other baselines listed in Table~\ref{tab:main_exp} cannot do irregular forecasting and we compare STG-NCDE with its ablation models in Tables~\ref{tbl:missing_pemsd4} and~\ref{tbl:missing_pemsd8}.

\begin{table}[t]
\centering
\setlength{\tabcolsep}{4pt}
\caption{Forecasting error on irregular PeMSD4}\label{tbl:missing_pemsd4}
\begin{tabular}{cc ccc}
\hline
Model                  & Missing rate         &  MAE   &     RMSE    &   MAPE  \\ \hline
\textbf{STG-NCDE}      & \multirow{3}{*}{10\%}& 19.36  & 31.28  & 12.79\% \\
\textbf{Only Temporal} &                      & 26.26  & 40.89  & 17.66\% \\
\textbf{Only Spatial } &                      & 19.73  & 31.67  & 13.20\% \\\hline
\textbf{STG-NCDE}      & \multirow{3}{*}{30\%}& 19.40  & 31.30  & 13.04\%\\
\textbf{Only Temporal} &                      & 26.86  & 41.73  & 18.35\% \\
\textbf{Only Spatial } &                      & 19.83  & 31.95  & 13.29\% \\\hline
\textbf{STG-NCDE}      & \multirow{3}{*}{50\%}& 19.98  & 32.09  & 13.48\% \\
\textbf{Only Temporal} &                      & 28.15  & 43.54  & 19.14\% \\
\textbf{Only Spatial } &                      & 20.14  & 32.30  & 13.30\% \\\hline
\end{tabular}
\end{table}

\begin{table}[t]
\centering
\setlength{\tabcolsep}{4pt}
\caption{Forecasting error on irregular PeMSD8. More results in other datasets are in Appendix.}\label{tbl:missing_pemsd8}
\begin{tabular}{cc ccc}
\hline
Model                  & Missing rate         &  MAE   &     RMSE    &   MAPE  \\ \hline
\textbf{STG-NCDE}      & \multirow{3}{*}{10\%}& 15.68  & 24.96  & 10.05\% \\
\textbf{Only Temporal} &                      & 21.18  & 33.02  & 13.26\% \\
\textbf{Only Spatial } &                      & 16.85  & 26.63  & 11.12\% \\\hline
\textbf{STG-NCDE}      & \multirow{3}{*}{30\%}& 16.21  & 25.64  & 10.43\% \\
\textbf{Only Temporal} &                      & 21.46  & 33.37  & 13.57\% \\
\textbf{Only Spatial } &                      & 18.46  & 29.03  & 12.16\% \\\hline
\textbf{STG-NCDE}      & \multirow{3}{*}{50\%}& 16.68  & 26.17  & 10.67\% \\
\textbf{Only Temporal} &                      & 22.68  & 35.14  & 14.11\% \\
\textbf{Only Spatial } &                      & 17.98  & 28.12  & 11.87\% \\\hline
\end{tabular}
\end{table}

\section{Conclusions}
We presented a spatio-temporal NCDE model to perform traffic forecasting. Our model has two NCDEs: one for temporal processing and the other for spatial processing. In particular, our NCDE for spatial processing can be considered as an NCDE-based interpretation of graph convolutional networks. In our experiments with 6 datasets and 20 baselines, our method clearly shows the best overall accuracy. In addition, our model can perform irregular traffic forecasting where some input observations can be missing, which is a practical problem setting but not actively considered by existing methods. We believe that the combination of NCDEs and GCNs is a promising research direction for spatio-temporal processing.

\section*{Acknowledgement}
Noseong Park is the corresponding author. This work was supported by the Yonsei University Research Fund of 2021, and the Institute of Information \& Communications Technology Planning \& Evaluation (IITP) grant funded by the Korean government (MSIT) (No. 2020-0-01361, Artificial Intelligence Graduate School Program (Yonsei University).

\bibliography{reference}

\begin{thebibliography}{31}
\providecommand{\natexlab}[1]{#1}

\bibitem[{Bai et~al.(2019)Bai, Yao, Kanhere, Wang, and Sheng}]{bai2019STG2Seq}
Bai, L.; Yao, L.; Kanhere, S.~S.; Wang, X.; and Sheng, Q.~Z. 2019.
\newblock STG2Seq: Spatial-Temporal Graph to Sequence Model for Multi-step
  Passenger Demand Forecasting.
\newblock In \emph{IJCAI}.

\bibitem[{Bai et~al.(2020)Bai, Yao, Li, Wang, and Wang}]{NEURIPS2020_ce1aad92}
Bai, L.; Yao, L.; Li, C.; Wang, X.; and Wang, C. 2020.
\newblock Adaptive Graph Convolutional Recurrent Network for Traffic
  Forecasting.
\newblock In \emph{NeurIPS}, volume~33, 17804--17815.

\bibitem[{Bai, Kolter, and Koltun(2018)}]{BaiTCN2018}
Bai, S.; Kolter, J.~Z.; and Koltun, V. 2018.
\newblock An Empirical Evaluation of Generic Convolutional and Recurrent
  Networks for Sequence Modeling.
\newblock \emph{arXiv:1803.01271}.

\bibitem[{Chen et~al.(2001)Chen, Petty, Skabardonis, Varaiya, and
  Jia}]{chen2001freeway}
Chen, C.; Petty, K.; Skabardonis, A.; Varaiya, P.; and Jia, Z. 2001.
\newblock Freeway performance measurement system: mining loop detector data.
\newblock \emph{Transportation Research Record}, 1748(1): 96--102.

\bibitem[{Chen, Segovia-Dominguez, and Gel(2021)}]{chen2021ZGCNET}
Chen, Y.; Segovia-Dominguez, I.; and Gel, Y.~R. 2021.
\newblock Z-GCNETs: Time Zigzags at Graph Convolutional Networks for Time
  Series Forecasting.
\newblock In \emph{ICML}.

\bibitem[{Cheng et~al.(2018{\natexlab{a}})Cheng, Zang, Ding, Sun, Wang, Wei,
  and Sun}]{cheng2018ensemble}
Cheng, L.; Zang, H.; Ding, T.; Sun, R.; Wang, M.; Wei, Z.; and Sun, G.
  2018{\natexlab{a}}.
\newblock Ensemble recurrent neural network based probabilistic wind speed
  forecasting approach.
\newblock \emph{Energies}, 11(8).

\bibitem[{Cheng et~al.(2018{\natexlab{b}})Cheng, Shen, Zhu, and
  Huang}]{cheng2018neural}
Cheng, W.; Shen, Y.; Zhu, Y.; and Huang, L. 2018{\natexlab{b}}.
\newblock A neural attention model for urban air quality inference: Learning
  the weights of monitoring stations.
\newblock In \emph{AAAI}.

\bibitem[{Cho et~al.(2014)Cho, {van Merrienboer}, Gulcehre, Bougares, Schwenk,
  and Bengio}]{cho2014grued}
Cho, K.; {van Merrienboer}, B.; Gulcehre, C.; Bougares, F.; Schwenk, H.; and
  Bengio, Y. 2014.
\newblock Learning phrase representations using RNN encoder-decoder for
  statistical machine translation.
\newblock In \emph{EMNLP}.

\bibitem[{Dormand and Prince(1980)}]{DORMAND198019}
Dormand, J.; and Prince, P. 1980.
\newblock A family of embedded Runge-Kutta formulae.
\newblock \emph{Journal of Computational and Applied Mathematics}, 6(1): 19 --
  26.

\bibitem[{Fang et~al.(2021)Fang, Long, Song, and Xie}]{fang2021STODE}
Fang, Z.; Long, Q.; Song, G.; and Xie, K. 2021.
\newblock Spatial-Temporal Graph ODE Networks for Traffic Flow Forecasting.
\newblock In \emph{KDD}.

\bibitem[{Guo et~al.(2019)Guo, Lin, Feng, Song, and Wan}]{guo2019astgcn}
Guo, S.; Lin, Y.; Feng, N.; Song, C.; and Wan, H. 2019.
\newblock Attention Based Spatial-Temporal Graph Convolutional Networks for
  Traffic Flow Forecasting.
\newblock In \emph{AAAI}.

\bibitem[{Hamilton(2020)}]{hamilton2020time}
Hamilton, J.~D. 2020.
\newblock \emph{Time series analysis}.
\newblock Princeton university press.

\bibitem[{Hossain et~al.(2015)Hossain, Rekabdar, Louis, and
  Dascalu}]{hossain2015forecasting}
Hossain, M.; Rekabdar, B.; Louis, S.~J.; and Dascalu, S. 2015.
\newblock Forecasting the weather of Nevada: A deep learning approach.
\newblock In \emph{IJCNN}.

\bibitem[{Huang et~al.(2020)Huang, Huang, Liu, Dai, and Kong}]{huang2020lsgcn}
Huang, R.; Huang, C.; Liu, Y.; Dai, G.; and Kong, W. 2020.
\newblock LSGCN: Long Short-Term Traffic Prediction with Graph Convolutional
  Networks.
\newblock In \emph{IJCAI}, 2355--2361.

\bibitem[{Huang et~al.(2019)Huang, Wang, Wu, and Tang}]{Huang2019DSANet}
Huang, S.; Wang, D.; Wu, X.; and Tang, A. 2019.
\newblock DSANet: Dual Self-Attention Network for Multivariate Time Series
  Forecasting.
\newblock In \emph{CIKM}.

\bibitem[{Kipf and Welling(2017)}]{kipf2017semi}
Kipf, T.~N.; and Welling, M. 2017.
\newblock Semi-Supervised Classification with Graph Convolutional Networks.
\newblock In \emph{ICLR}.

\bibitem[{Kurth et~al.(2018)Kurth, Treichler, Romero, Mudigonda, Luehr,
  Phillips, Mahesh, Matheson, Deslippe, Fatica et~al.}]{kurth2018exascale}
Kurth, T.; Treichler, S.; Romero, J.; Mudigonda, M.; Luehr, N.; Phillips, E.;
  Mahesh, A.; Matheson, M.; Deslippe, J.; Fatica, M.; et~al. 2018.
\newblock Exascale deep learning for climate analytics.
\newblock In \emph{International Conference for High Performance Computing,
  Networking, Storage and Analysis}. IEEE.

\bibitem[{Li and Zhu(2021)}]{li2021stfgnn}
Li, M.; and Zhu, Z. 2021.
\newblock Spatial-Temporal Fusion Graph Neural Networks for Traffic Flow
  Forecasting.
\newblock In \emph{AAAI}.

\bibitem[{Li et~al.(2018)Li, Yu, Shahabi, and Liu}]{li2018dcrnn_traffic}
Li, Y.; Yu, R.; Shahabi, C.; and Liu, Y. 2018.
\newblock Diffusion Convolutional Recurrent Neural Network: Data-Driven Traffic
  Forecasting.
\newblock In \emph{ICLR}.

\bibitem[{Liu et~al.(2016)Liu, Racah, Correa, Khosrowshahi, Lavers, Kunkel,
  Wehner, Collins et~al.}]{liu2016application}
Liu, Y.; Racah, E.; Correa, J.; Khosrowshahi, A.; Lavers, D.; Kunkel, K.;
  Wehner, M.; Collins, W.; et~al. 2016.
\newblock Application of deep convolutional neural networks for detecting
  extreme weather in climate datasets.
\newblock \emph{arXiv preprint}.

\bibitem[{Lyons, Caruana, and L{\'e}vy(2007)}]{lyons2007differential}
Lyons, T.~J.; Caruana, M.; and L{\'e}vy, T. 2007.
\newblock \emph{Differential equations driven by rough paths}.
\newblock Springer.

\bibitem[{Racah et~al.(2016)Racah, Beckham, Maharaj, Kahou, Pal
  et~al.}]{racah2016extremeweather}
Racah, E.; Beckham, C.; Maharaj, T.; Kahou, S.~E.; Pal, C.; et~al. 2016.
\newblock ExtremeWeather: A large-scale climate dataset for semi-supervised
  detection, localization, and understanding of extreme weather events.
\newblock \emph{arXiv preprint}.

\bibitem[{Ren et~al.(2021)Ren, Li, Ren, Song, Xu, Deng, and Wang}]{ren2021deep}
Ren, X.; Li, X.; Ren, K.; Song, J.; Xu, Z.; Deng, K.; and Wang, X. 2021.
\newblock Deep Learning-Based Weather Prediction: A Survey.
\newblock \emph{Big Data Research}, 23.

\bibitem[{Shi et~al.(2015)Shi, Chen, Wang, Yeung, Wong, and
  Woo}]{shi2015convolutional}
Shi, X.; Chen, Z.; Wang, H.; Yeung, D.-Y.; Wong, W.-K.; and Woo, W.-c. 2015.
\newblock Convolutional LSTM network: A machine learning approach for
  precipitation nowcasting.
\newblock In \emph{NeurIPS}.

\bibitem[{Shi et~al.(2017)Shi, Gao, Lausen, Wang, Yeung, Wong, and
  Woo}]{shi2017deep}
Shi, X.; Gao, Z.; Lausen, L.; Wang, H.; Yeung, D.-Y.; Wong, W.-k.; and Woo,
  W.-c. 2017.
\newblock Deep learning for precipitation nowcasting: A benchmark and a new
  model.
\newblock \emph{arXiv preprint}.

\bibitem[{Song et~al.(2020)Song, Lin, Guo, and Wan}]{song2020stsgcn}
Song, C.; Lin, Y.; Guo, S.; and Wan, H. 2020.
\newblock Spatial-Temporal Synchronous Graph Convolutional Networks: A New
  Framework for Spatial-Temporal Network Data Forecasting.
\newblock In \emph{AAAI}.

\bibitem[{Sutskever, Vinyals, and Le(2014)}]{sutskever2014sequence}
Sutskever, I.; Vinyals, O.; and Le, Q.~V. 2014.
\newblock Sequence to sequence learning with neural networks.
\newblock In \emph{NeurIPS}, 3104--3112.

\bibitem[{Tekin et~al.(2021)Tekin, Karaahmetoglu, Ilhan, Balaban, and
  Kozat}]{tekin2021spatio}
Tekin, S.~F.; Karaahmetoglu, O.; Ilhan, F.; Balaban, I.; and Kozat, S.~S. 2021.
\newblock Spatio-temporal Weather Forecasting and Attention Mechanism on
  Convolutional LSTMs.
\newblock \emph{arXiv preprint}.

\bibitem[{Wu et~al.(2019)Wu, Pan, Long, Jiang, and Zhang}]{wu2019graphwavenet}
Wu, Z.; Pan, S.; Long, G.; Jiang, J.; and Zhang, C. 2019.
\newblock Graph WaveNet for Deep Spatial-Temporal Graph Modeling.
\newblock In \emph{IJCAI}, 1907--1913.

\bibitem[{Yu, Yin, and Zhu(2018)}]{bing2018stgcn}
Yu, B.; Yin, H.; and Zhu, Z. 2018.
\newblock Spatio-Temporal Graph Convolutional Networks: A Deep Learning
  Framework for Traffic Forecasting.
\newblock In \emph{IJCAI}.

\bibitem[{Zaytar and El~Amrani(2016)}]{zaytar2016sequence}
Zaytar, M.~A.; and El~Amrani, C. 2016.
\newblock Sequence to sequence weather forecasting with long short-term memory
  recurrent neural networks.
\newblock \emph{International Journal of Computer Applications}.

\end{thebibliography}

\clearpage

\appendix
\section{Best Hyperparameters}
For reproducibility, we introduce the best hyperparameter configurations for each dataset as follows:
\begin{enumerate}
    \item In PeMSD3, we set the number of $K$ to 1 and the node embedding size $C$ to 2. The dimensionality of hidden vector is 64. The learning rate was set to \num{1e-3} and the weight decay was \num{1e-3}.
    \item In PeMSD4, we set the number of $K$ to 2 and the node embedding size $C$ to 8. The dimensionality of hidden vector is 64. The learning rate was set to \num{1e-3} and the weight decay was \num{1e-3}.
    \item In PeMSD7, we set the number of $K$ to 2 and the node embedding size $C$ to 10. The dimensionality of hidden vector is 64. The learning rate was set to \num{1e-3} and the weight decay was \num{1e-3}.
    \item In PeMSD8, we set the number of $K$ to 1 and the node embedding size $C$ to 2. The dimensionality of hidden vector is 32. The learning rate was set to \num{1e-3} and the weight decay was \num{1e-3}.
    \item In PeMSD7(M), we set the number of $K$ to 1 and the node embedding size $C$ to 10. The dimensionality of hidden vector is 32. The learning rate was set to \num{1e-3} and the weight decay was \num{1e-3}.
    \item In PeMSD7(L), we set the number of $K$ to 1 and the node embedding size $C$ to 10. The dimensionality of hidden vector is 32. The learning rate was set to \num{1e-3} and the weight decay was \num{1e-3}.
\end{enumerate}
\begin{table}[!ht]
\centering
\setlength{\tabcolsep}{4pt}
\caption{Forecasting error on irregular PeMSD3}\label{tbl:missing_pemsd3}
\begin{tabular}{cc ccc}
\hline
Model                  & Missing rate         &  MAE   &     RMSE    &   MAPE  \\ \hline
\textbf{STG-NCDE}      & \multirow{3}{*}{10\%}&  15.89 & 27.61  & 15.83\% \\ 
\textbf{Only Temporal} &                      & 57.15  & 85.11  & 54.38\% \\
\textbf{Only Spatial } &                      & 15.85  & 27.24  & 15.22\% \\\hline
\textbf{STG-NCDE}      & \multirow{3}{*}{30\%}& 16.08  & 27.78  & 16.05\% \\ 
\textbf{Only Temporal} &                      & 58.47  & 86.33  & 60.39\% \\ 
\textbf{Only Spatial } &                      & 16.18  & 27.61  & 15.18\% \\\hline 
\textbf{STG-NCDE}      & \multirow{3}{*}{50\%}& 16.50  & 28.52  & 16.03\% \\
\textbf{Only Temporal} &                      & 60.29  & 87.86 & 63.17\% \\
\textbf{Only Spatial } &                      & 16.75  & 28.64  & 16.00\% \\\hline
\end{tabular}
\vspace{1em}
\centering
\setlength{\tabcolsep}{4pt}
\caption{Forecasting error on irregular PeMSD7}\label{tbl:missing_pemsd7}
\begin{tabular}{cc ccc}
\hline
Model                  & Missing rate         &  MAE   &     RMSE    &   MAPE  \\ \hline
\textbf{STG-NCDE}      & \multirow{3}{*}{10\%}& 20.65  & 33.95  &  8.86\% \\
\textbf{Only Temporal} &                      & 29.49  & 45.19  & 12.96\% \\
\textbf{Only Spatial } &                      & 21.85  & 34.83  &  9.25\% \\\hline
\textbf{STG-NCDE}      & \multirow{3}{*}{30\%}& 20.76  & 34.20  &  8.91\% \\
\textbf{Only Temporal} &                      & 29.73  & 45.60  & 13.16\% \\
\textbf{Only Spatial } &                      & 21.82  & 34.93  &  9.21\% \\ \hline
\textbf{STG-NCDE}      & \multirow{3}{*}{50\%}& 21.51  & 34.91  &  9.25\% \\
\textbf{Only Temporal} &                      & 31.13  & 47.51  & 13.69\% \\
\textbf{Only Spatial } &                      & 22.64  & 35.94  &  9.60\% \\\hline
\end{tabular}
\vspace{1em}
\centering
\setlength{\tabcolsep}{4pt}
\caption{Forecasting error on irregular PeMSD7(M)}\label{tbl:missing_pemsd7m}
\begin{tabular}{cc ccc}
\hline
Model                  & Missing rate         &  MAE  &     RMSE    &   MAPE  \\ \hline
\textbf{STG-NCDE}      & \multirow{3}{*}{10\%}& 2.67  & 5.38  & 6.78\% \\
\textbf{Only Temporal} &                      & 3.33  & 6.67  & 8.34\% \\
\textbf{Only Spatial } &                      & 2.71  & 5.38  & 6.98\% \\\hline
\textbf{STG-NCDE}      & \multirow{3}{*}{30\%}& 2.72  & 5.45  & 6.81\% \\
\textbf{Only Temporal} &                      & 3.37  & 6.72  & 8.52\% \\
\textbf{Only Spatial } &                      & 2.73  & 5.41  & 6.89\% \\\hline
\textbf{STG-NCDE}      & \multirow{3}{*}{50\%}& 2.75  & 5.53  & 6.79\% \\
\textbf{Only Temporal} &                      & 3.49  & 6.95  & 8.76\% \\
\textbf{Only Spatial } &                      & 2.78  & 5.49  & 7.00\% \\\hline
\end{tabular}%
\vspace{1em}
\centering
\setlength{\tabcolsep}{4pt}
\caption{Forecasting error on irregular PeMSD7(L)}\label{tbl:missing_pemsd7l}
\begin{tabular}{cc ccc}
\hline
Model                  & Missing rate         &  MAE   &     RMSE    &   MAPE  \\ \hline
\textbf{STG-NCDE}      & \multirow{3}{*}{10\%}& 2.90  & 5.78  & 7.34\% \\
\textbf{Only Temporal} &                      & 3.52  & 7.04  & 8.84\% \\ 
\textbf{Only Spatial } &                      & 2.96  & 5.86  & 7.59\% \\\hline
\textbf{STG-NCDE}      & \multirow{3}{*}{30\%}& 2.89  & 5.80  & 7.25\% \\
\textbf{Only Temporal} &                      & 3.55  & 7.04  & 8.97\% \\
\textbf{Only Spatial } &                      & 2.97  & 5.85  & 7.45\% \\\hline
\textbf{STG-NCDE}      & \multirow{3}{*}{50\%}& 3.03  & 5.97  & 7.63\% \\
\textbf{Only Temporal} &                      & 3.68  & 7.27  & 9.28\% \\
\textbf{Only Spatial } &                      & 3.02  & 5.96  & 7.68\% \\\hline
\end{tabular}%
\end{table}

\section{Irregular Traffic Forecasting}
Tables~\ref{tbl:missing_pemsd3} to~\ref{tbl:missing_pemsd7l} show the irregular traffic forecasting results in the remaining datasets that are not reported in our main paper.

\section{Sensitivity Analysis}
\begin{figure}[t]
    \centering
    \subfigure[PeMSD3]{\includegraphics[width=0.48\columnwidth]{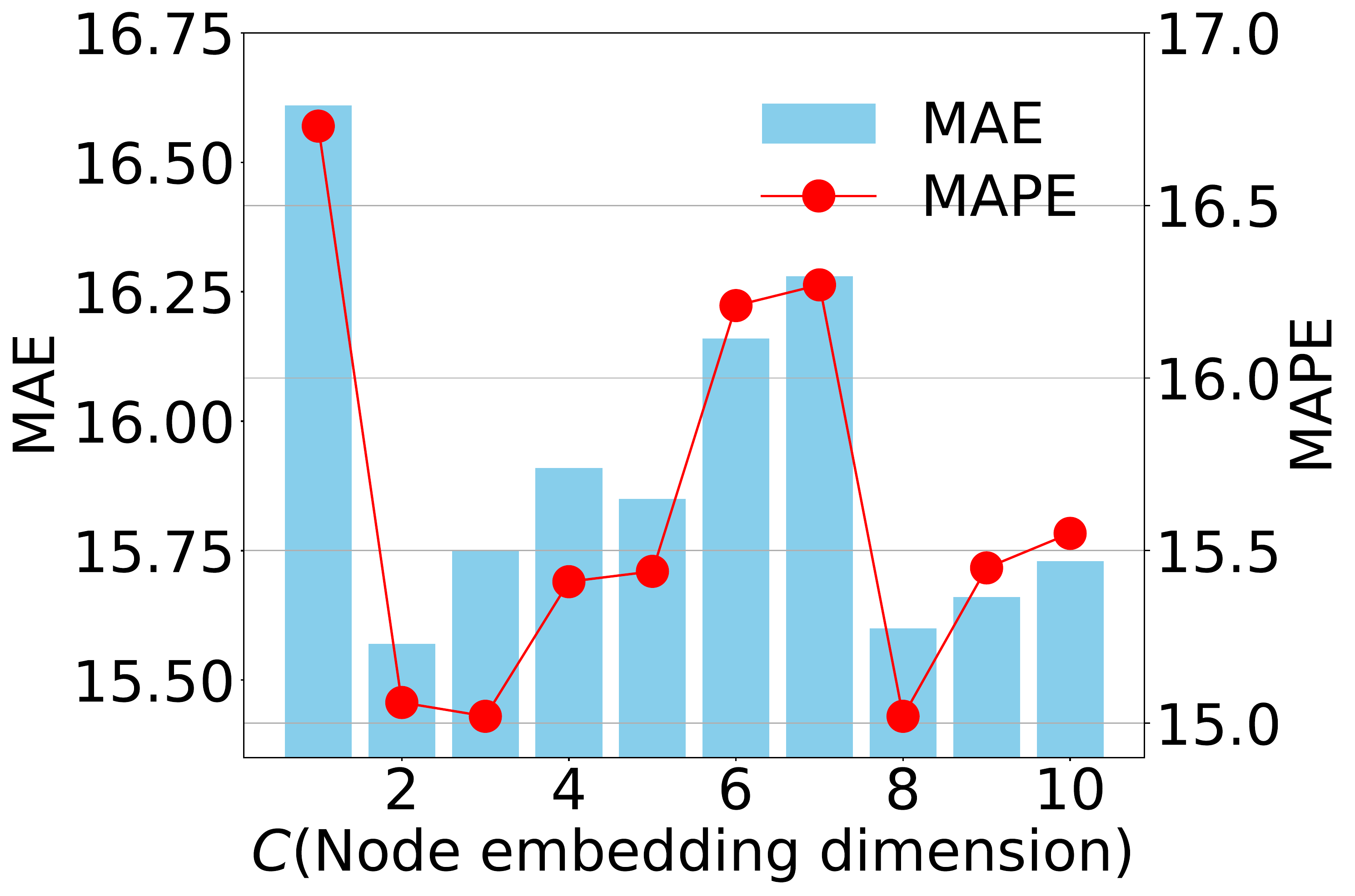}}
    \subfigure[PeMSD4]{\includegraphics[width=0.48\columnwidth]{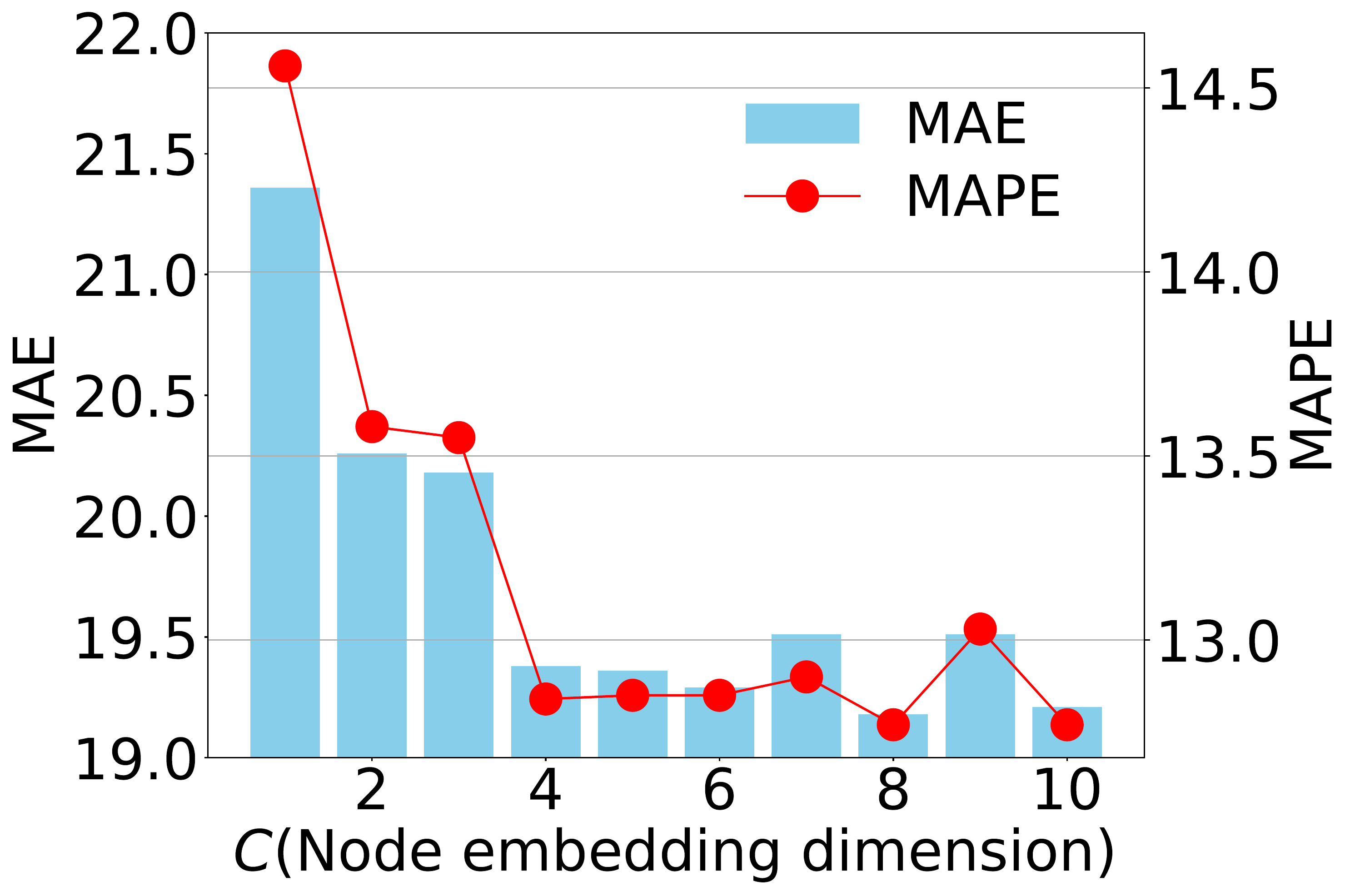}}
    \subfigure[PeMSD8]{\includegraphics[width=0.48\columnwidth]{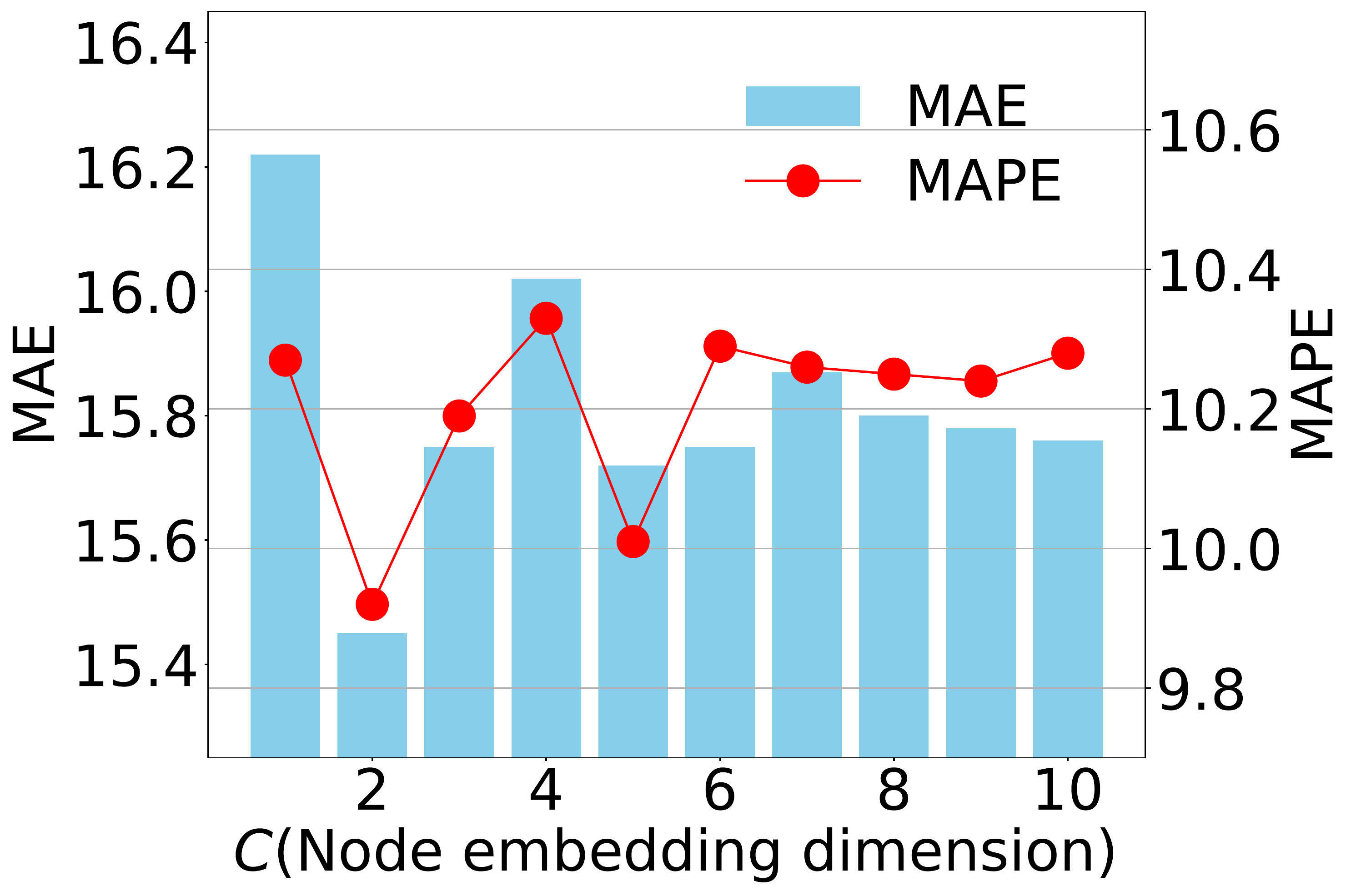}}
    \subfigure[PeMSD7(M)]{\includegraphics[width=0.48\columnwidth]{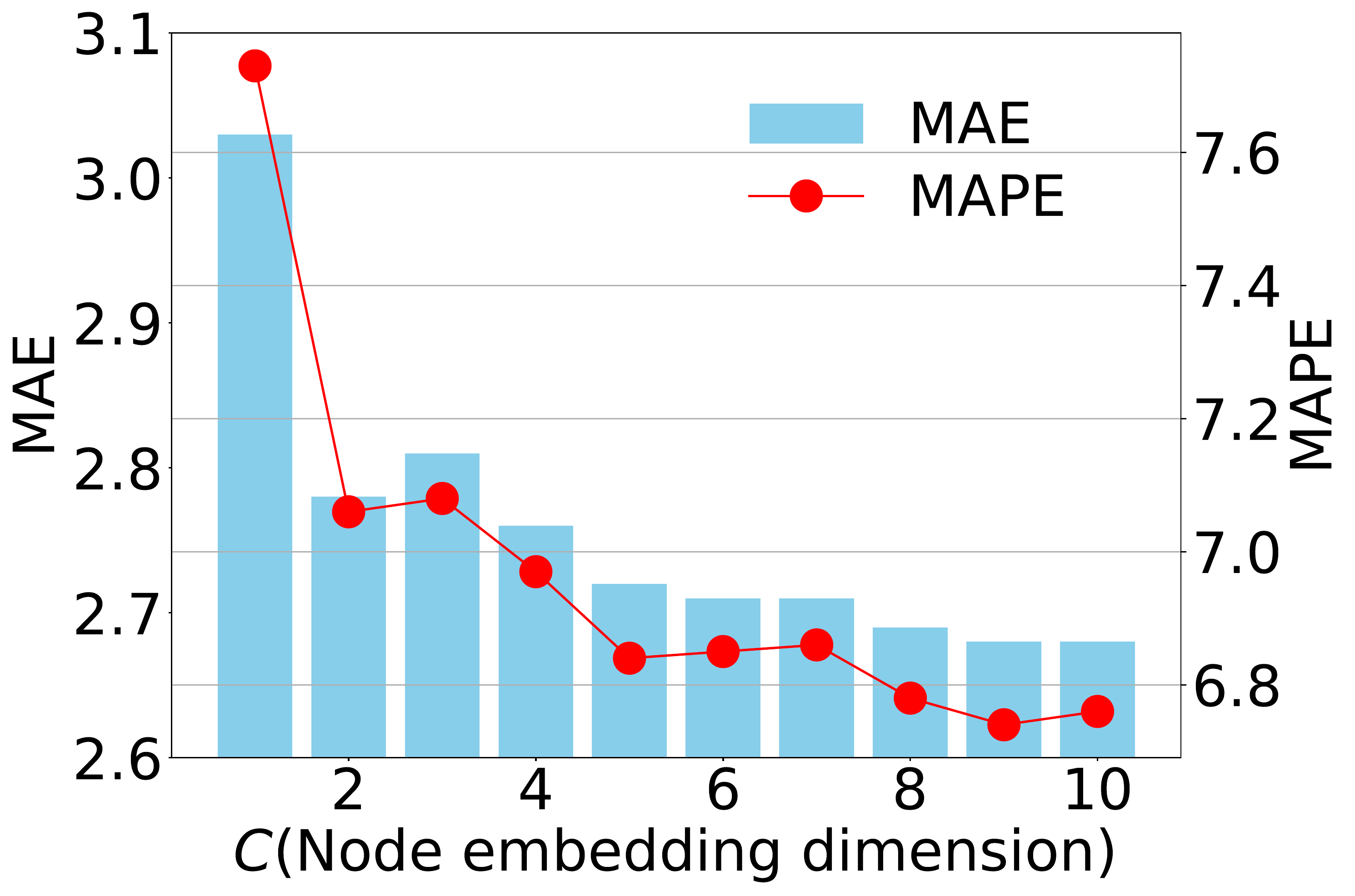}}
    \subfigure[PeMSD7(L)]{\includegraphics[width=0.48\columnwidth]{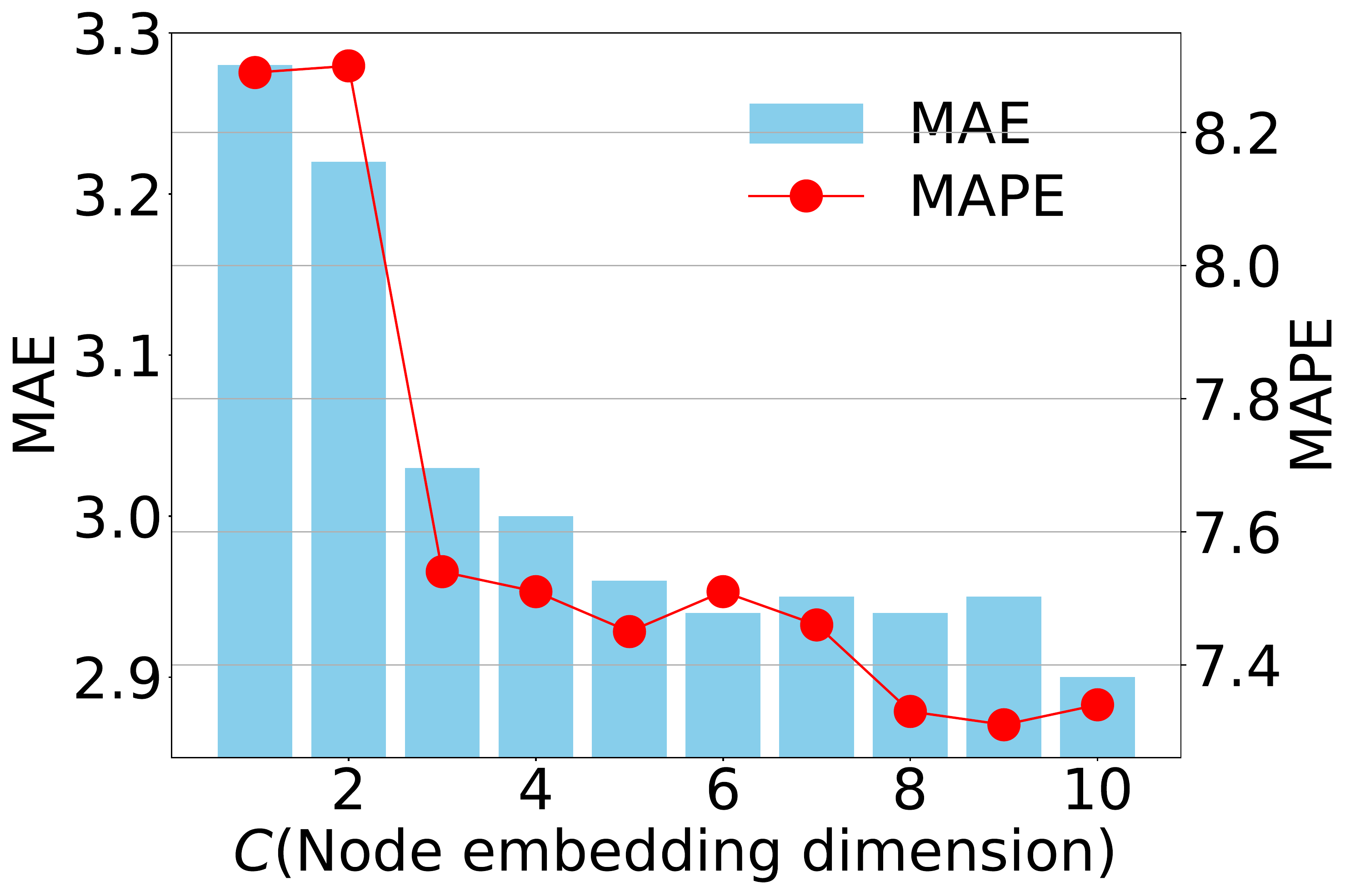}}
    \caption{Sensitivity to $C$}
    \label{fig:sensitivity_appendix}
\end{figure}

Fig.~\ref{fig:sensitivity_appendix} shows the MAE and MAPE by varying the node embedding size $C$ in the remaining datasets.

\section{Prediction Error at Each Horizon}
\begin{figure*}[ht]
    \centering
    \subfigure[MAE on PeMSD3]{\includegraphics[width=0.48\columnwidth]{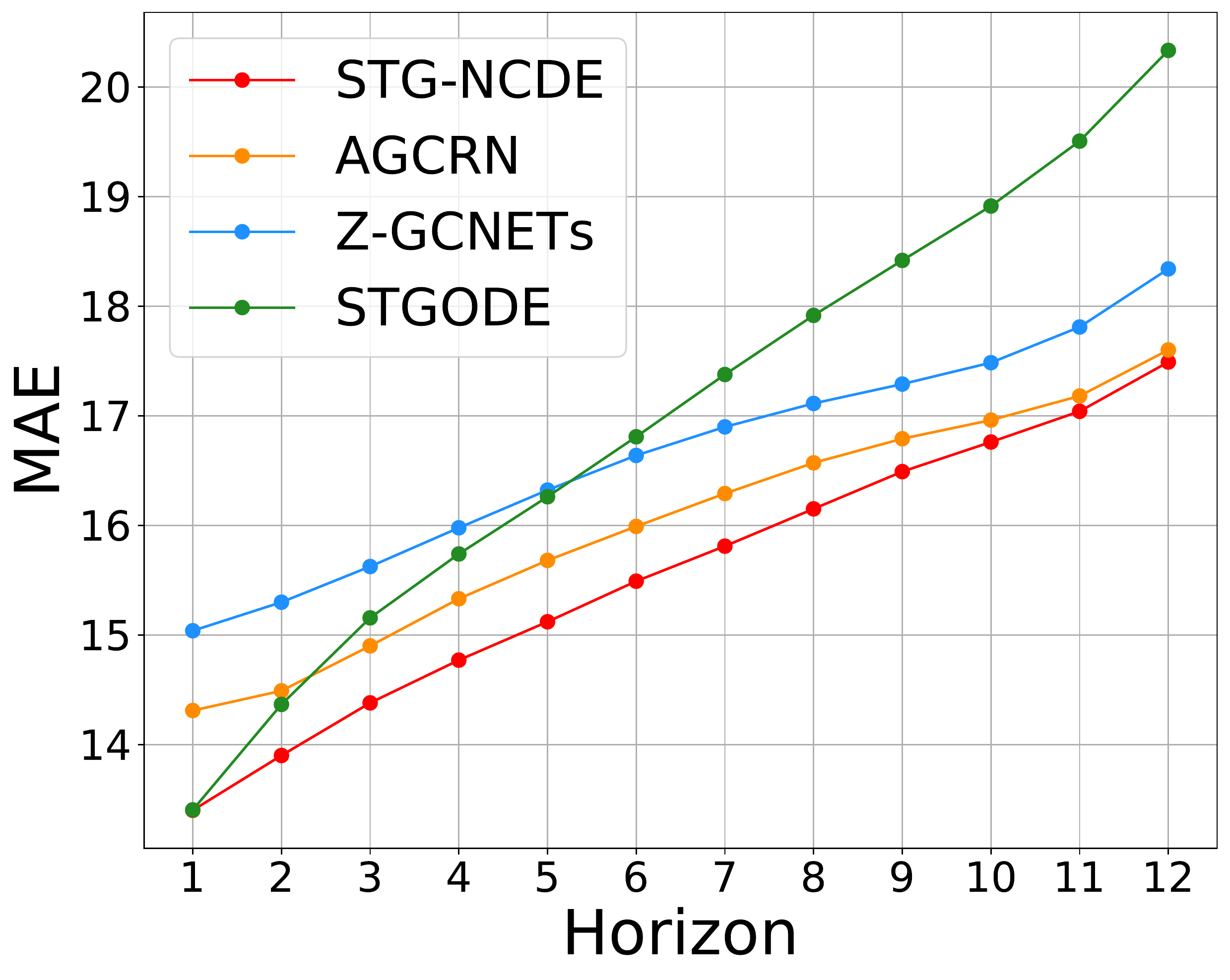}}
    \subfigure[RMSE on PeMSD3]{\includegraphics[width=0.48\columnwidth]{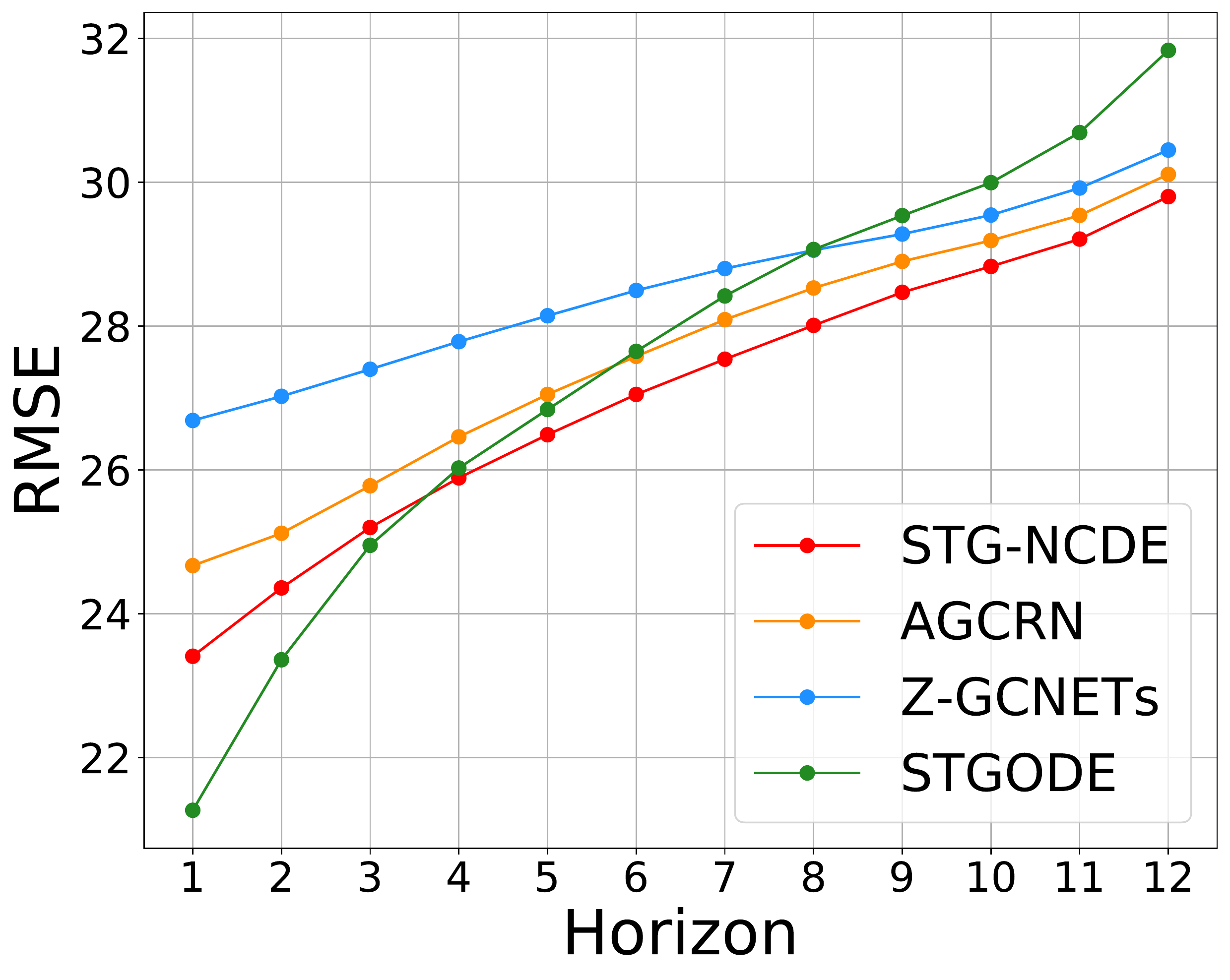}}
    \subfigure[MAPE on PeMSD3]{\includegraphics[width=0.48\columnwidth]{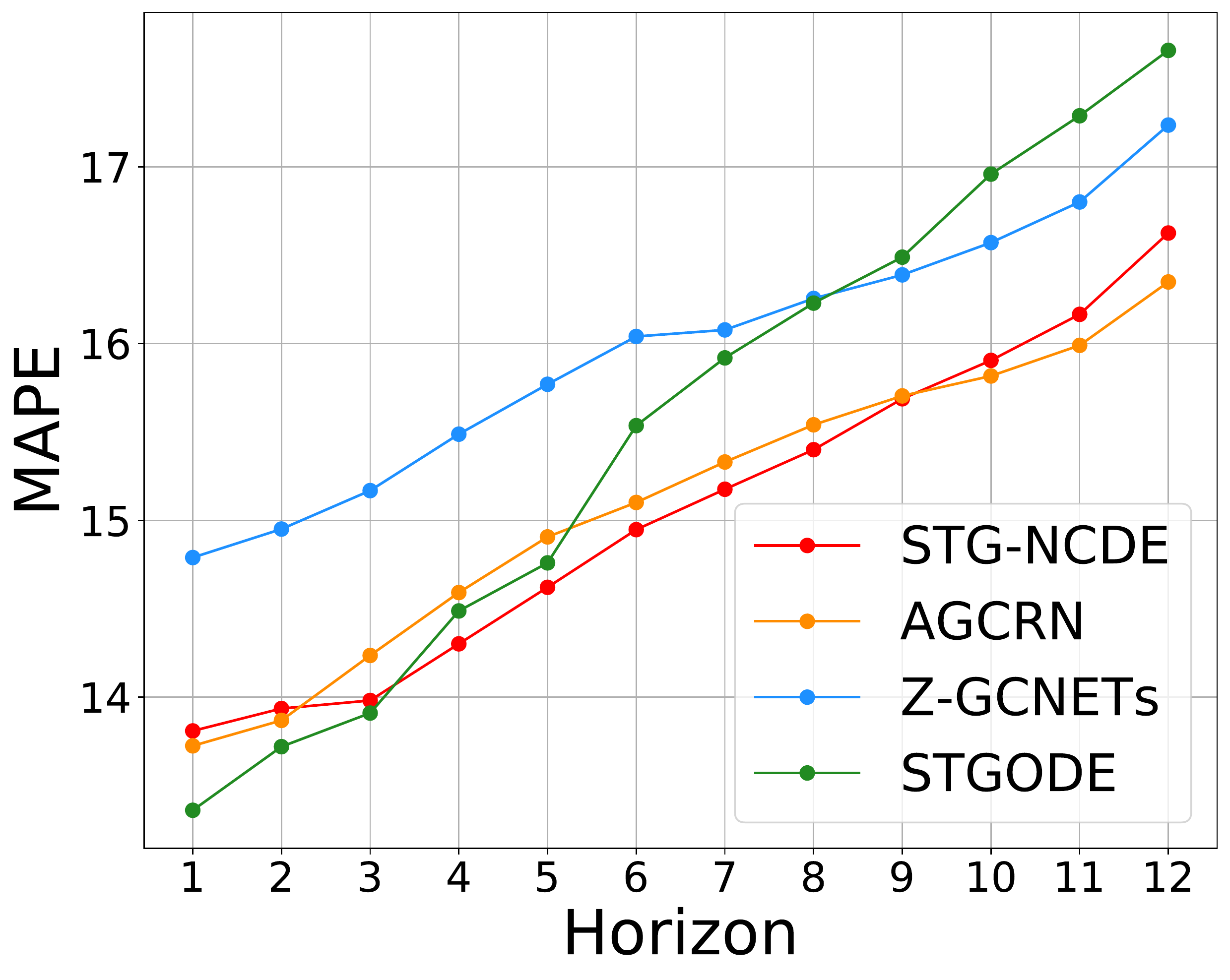}}
    \subfigure[MAE on PeMSD4]{\includegraphics[width=0.48\columnwidth]{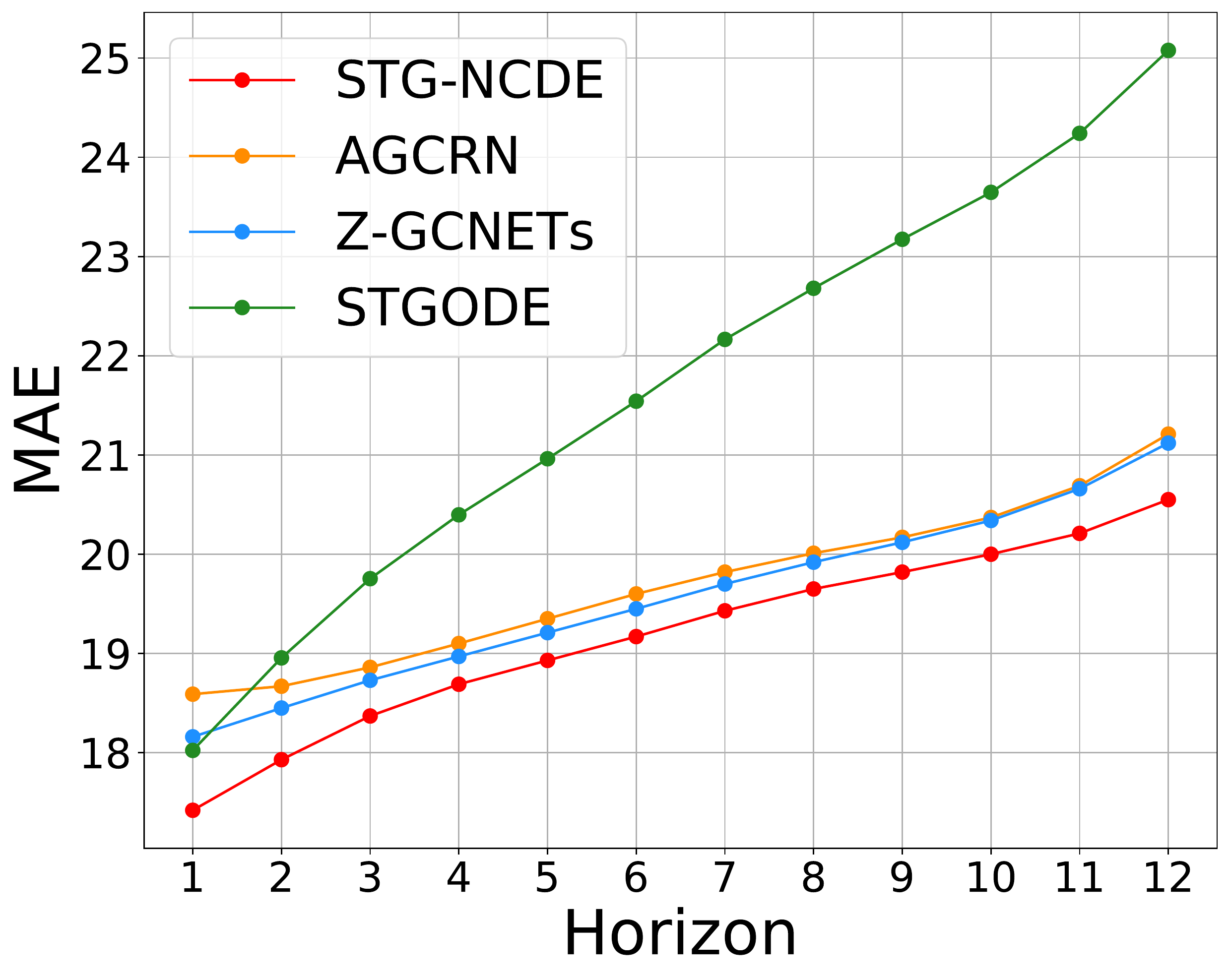}}
    \subfigure[MAPE on PeMSD4]{\includegraphics[width=0.48\columnwidth]{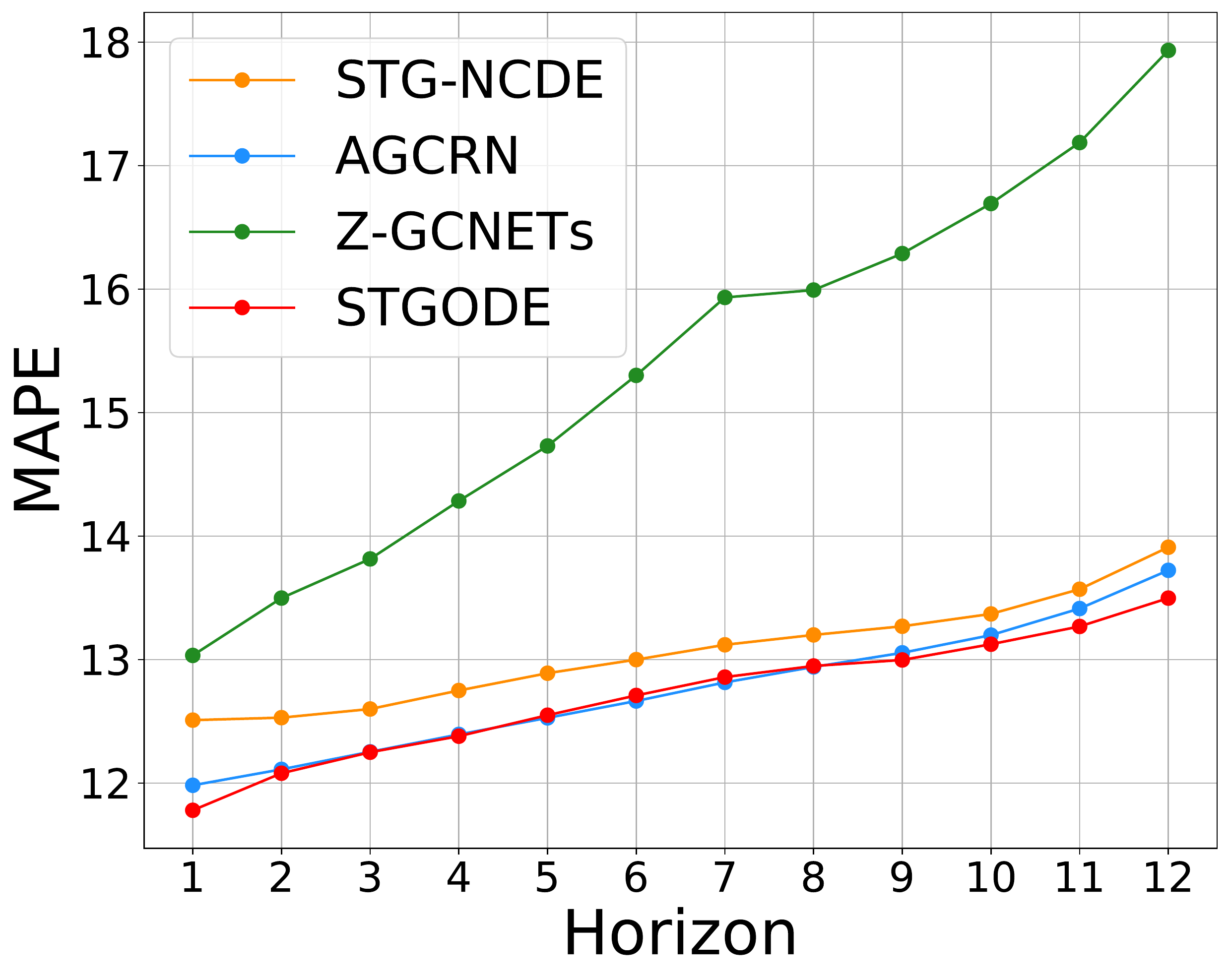}}
    \subfigure[RMSE on PeMSD4]{\includegraphics[width=0.48\columnwidth]{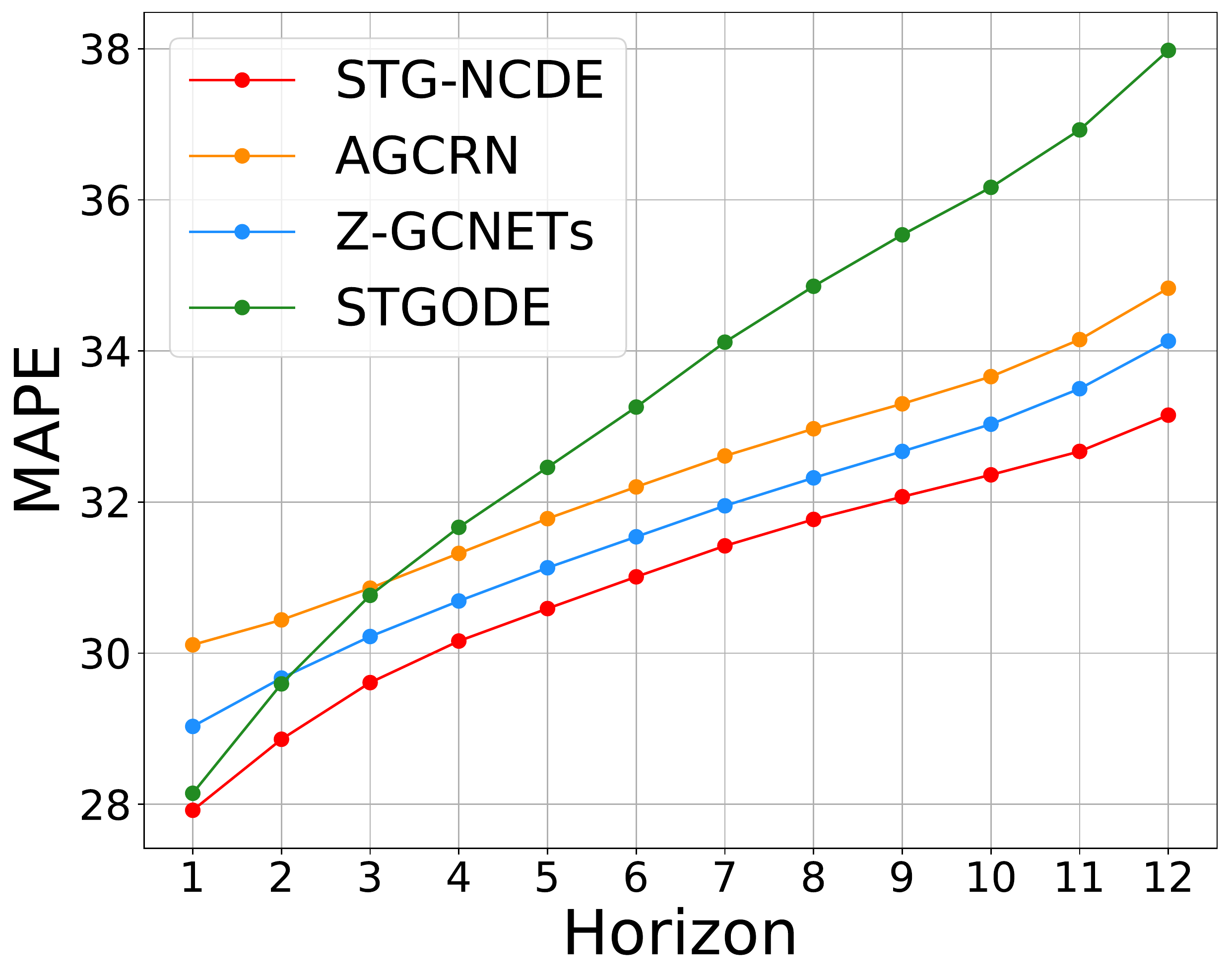}}
    \subfigure[RMSE on PeMSD7]{\includegraphics[width=0.48\columnwidth]{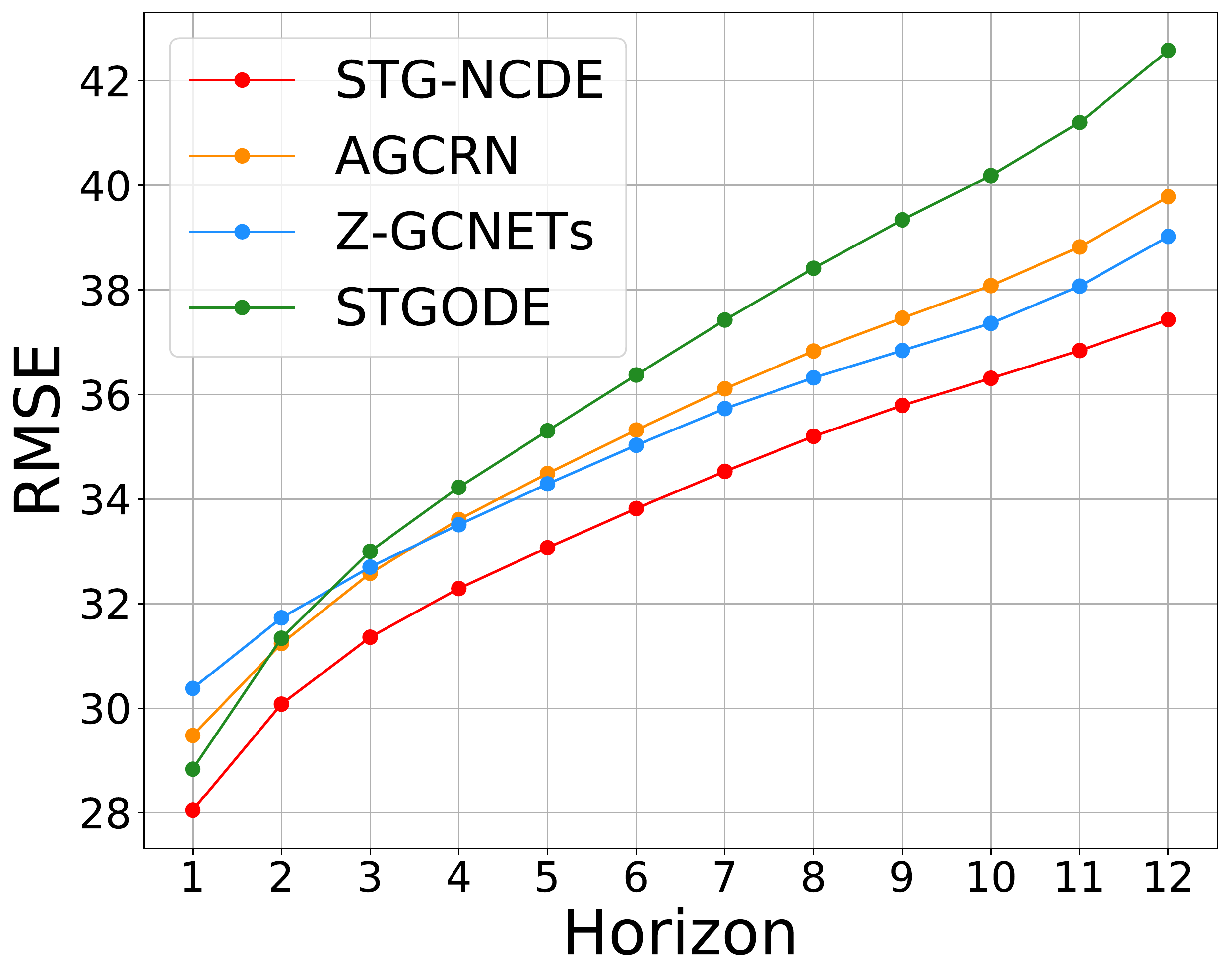}}
    \subfigure[RMSE on PeMSD8]{\includegraphics[width=0.48\columnwidth]{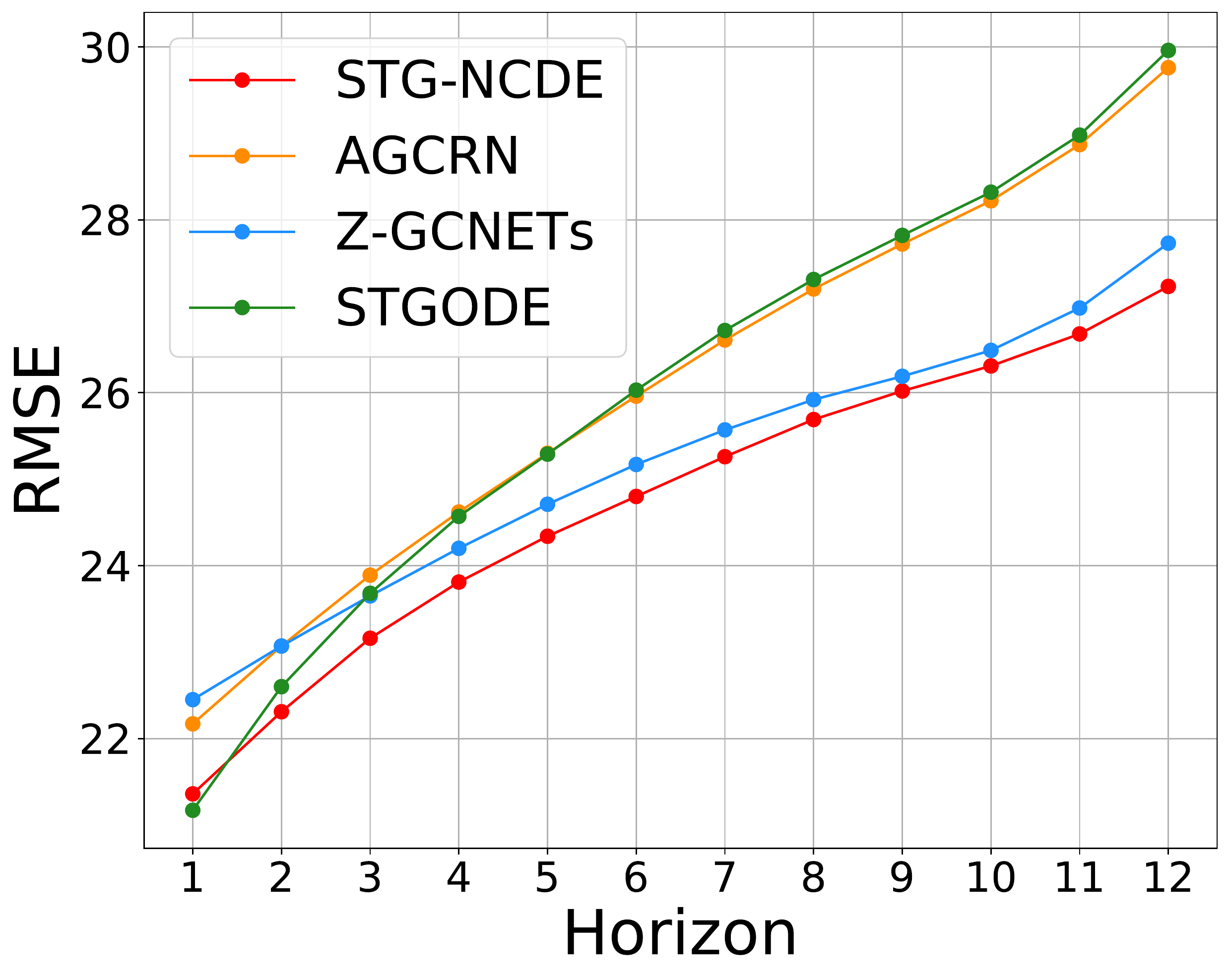}}
    \subfigure[MAE on PeMSD7(M)]{\includegraphics[width=0.48\columnwidth]{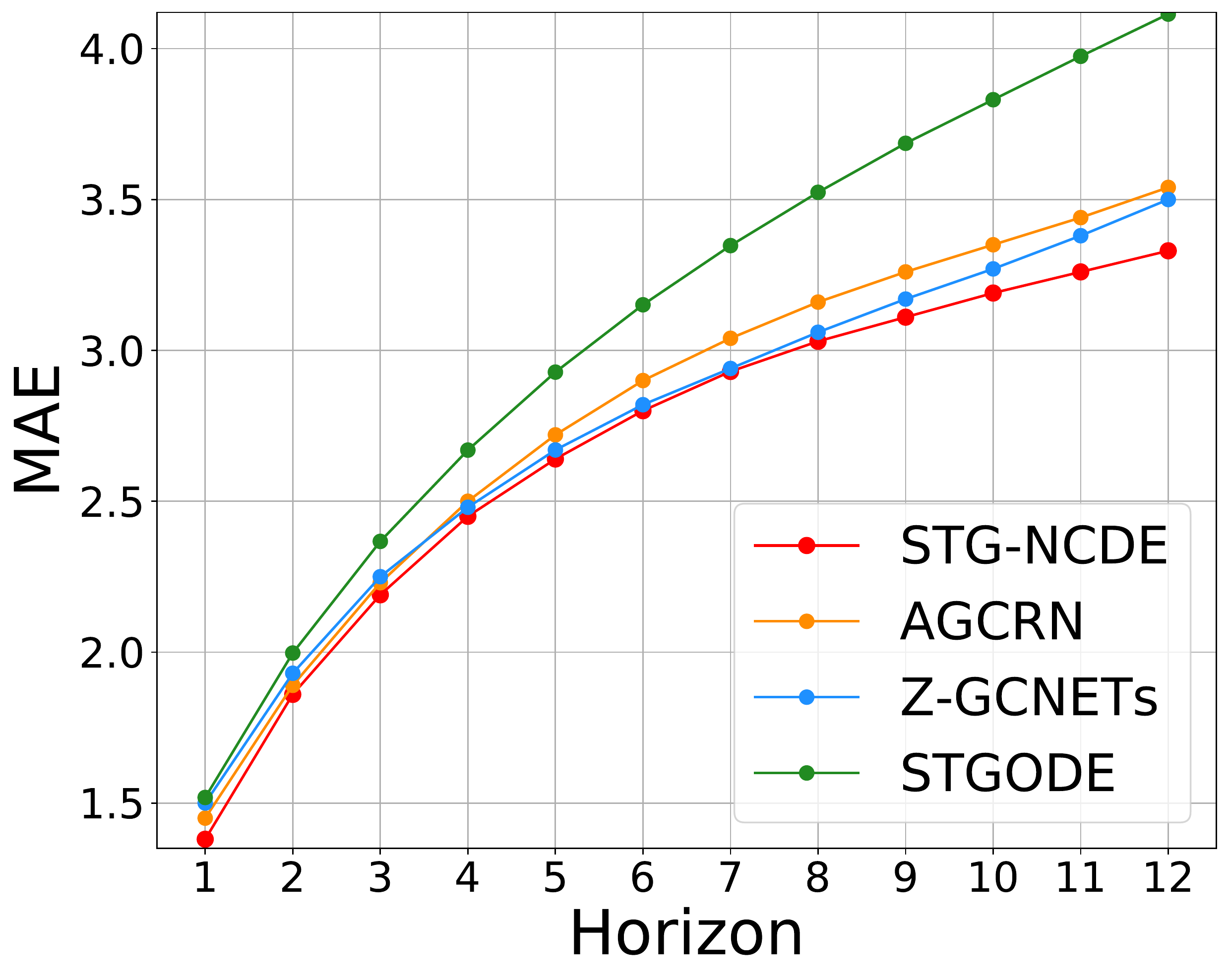}}
    \subfigure[MAPE on PeMSD7(M)]{\includegraphics[width=0.48\columnwidth]{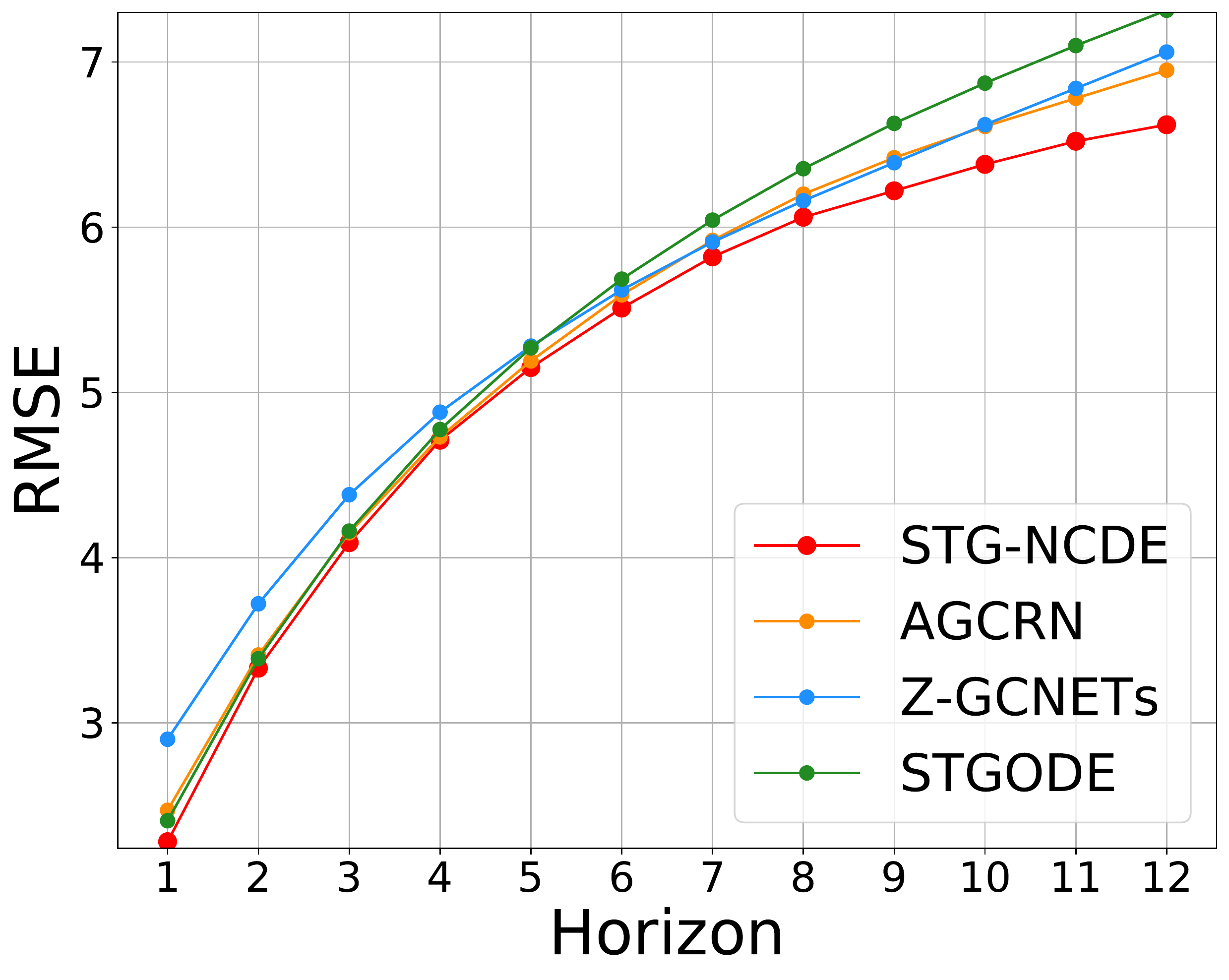}}
    \subfigure[RMSE on PeMSD7(M)]{\includegraphics[width=0.48\columnwidth]{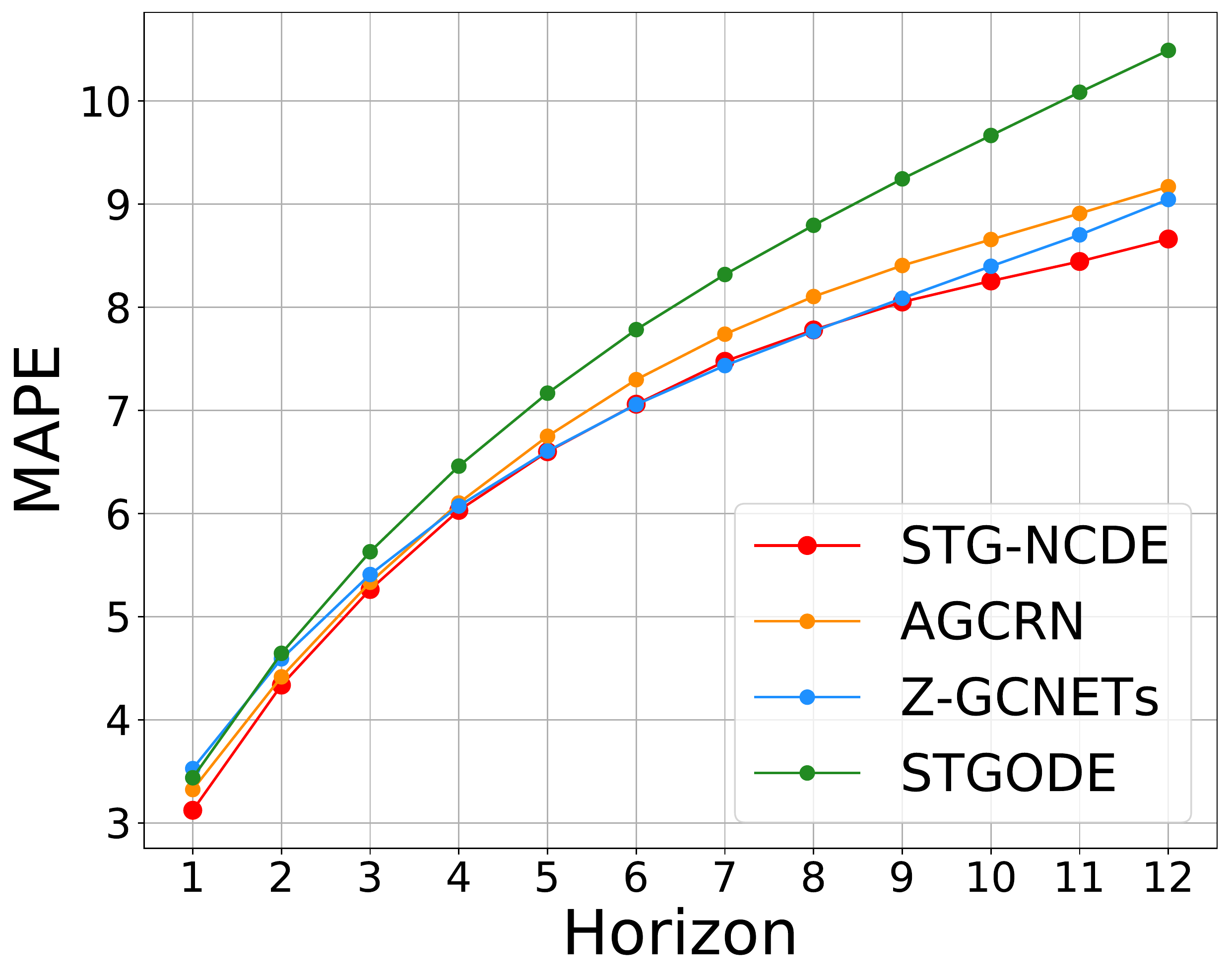}}
    \subfigure[MAE on PeMSD7(L)]{\includegraphics[width=0.48\columnwidth]{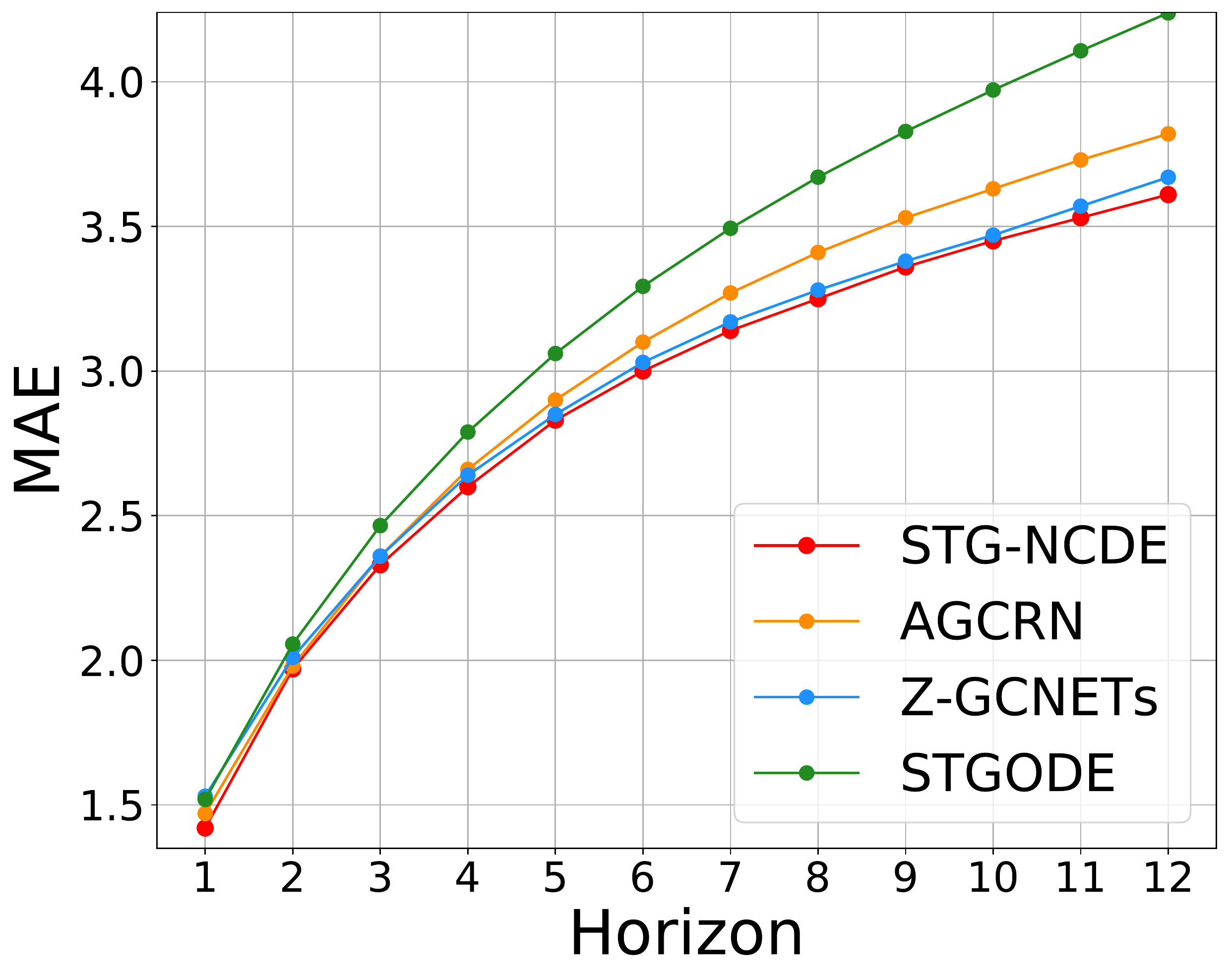}}
    \subfigure[MAPE on PeMSD7(L)]{\includegraphics[width=0.48\columnwidth]{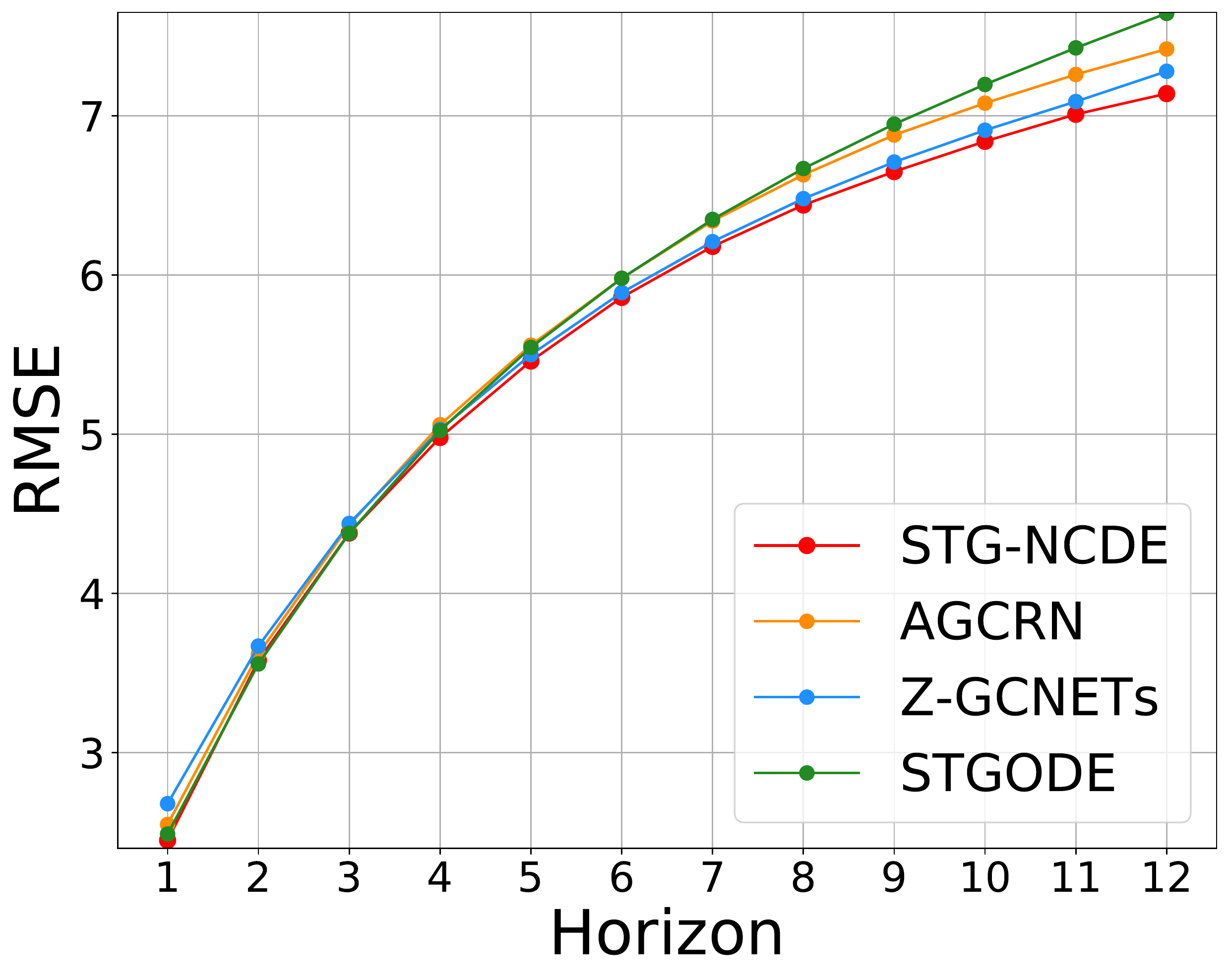}}
    \subfigure[RMSE on PeMSD7(L)]{\includegraphics[width=0.48\columnwidth]{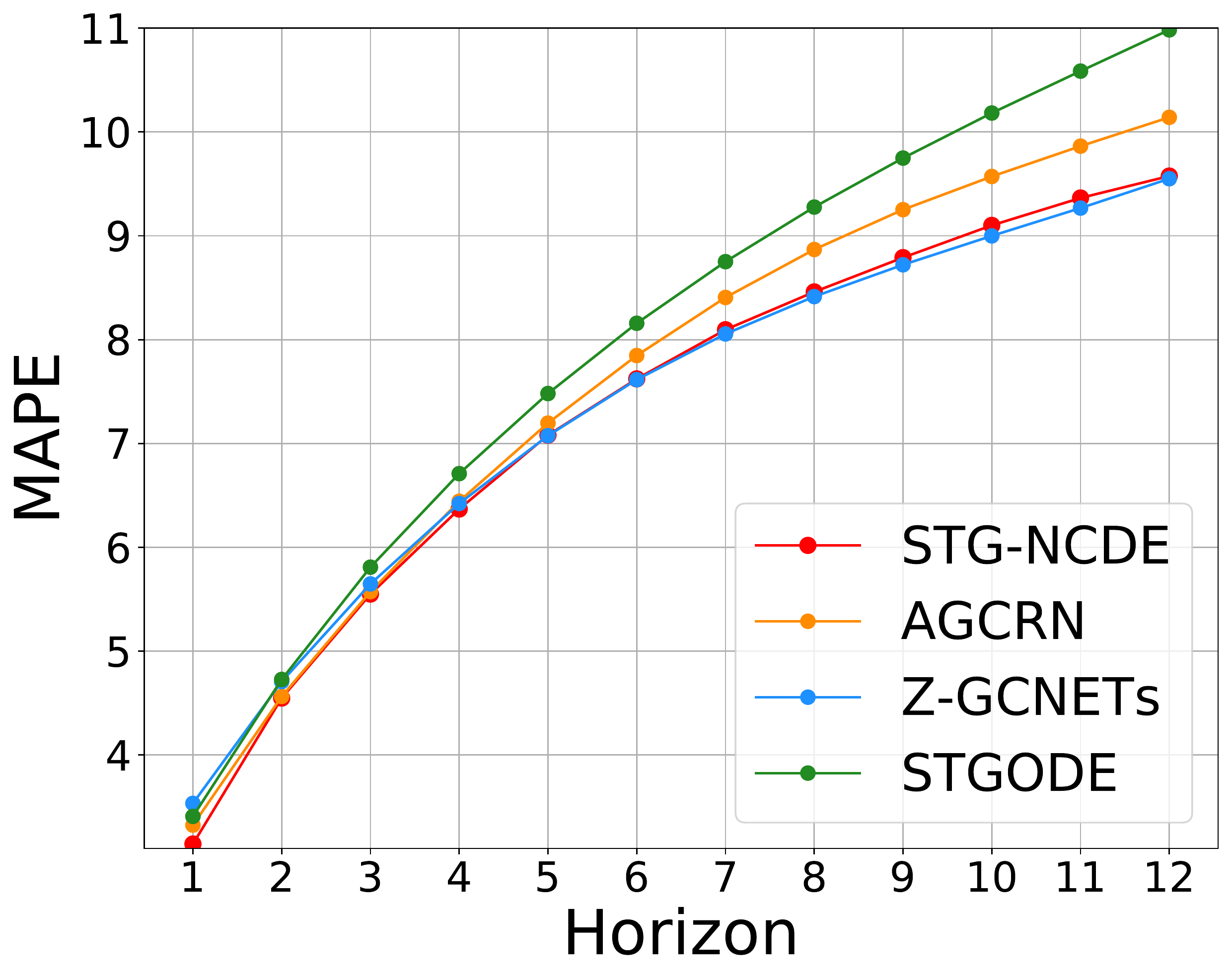}}
    \caption{Prediction error at each horizon}
    \label{fig:horizon_appendix}
\end{figure*}

Fig.~\ref{fig:horizon_appendix} shows the prediction error at each horizon in the remaining datasets that are not reported in our main paper.

\section{Traffic Forecasting Visualization}
\begin{figure*}[ht]
    \centering
    \subfigure[Node 12 in PeMSD3]{\includegraphics[width=0.48\columnwidth]{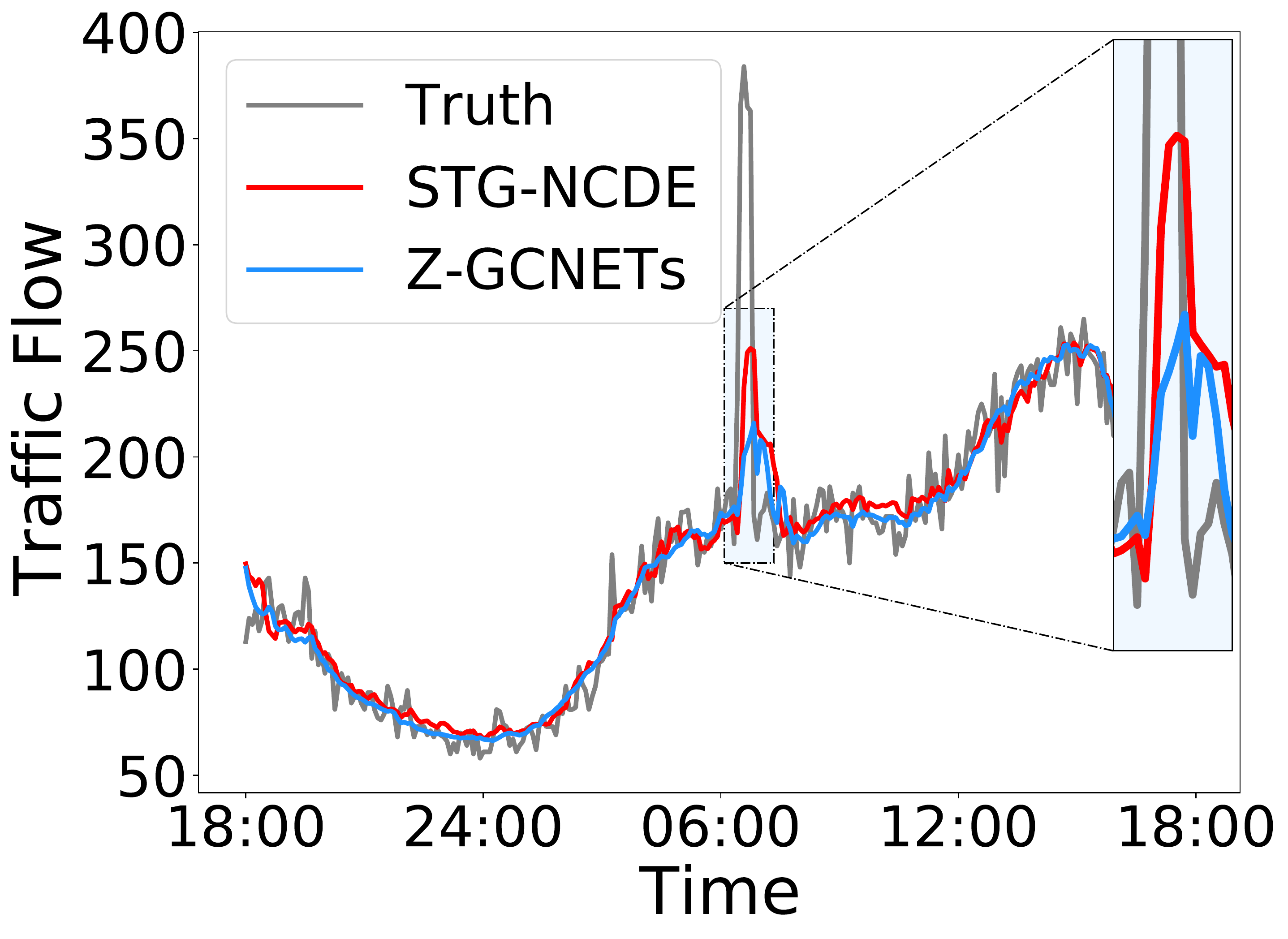}}
    \subfigure[Node 99 in PeMSD3]{\includegraphics[width=0.48\columnwidth]{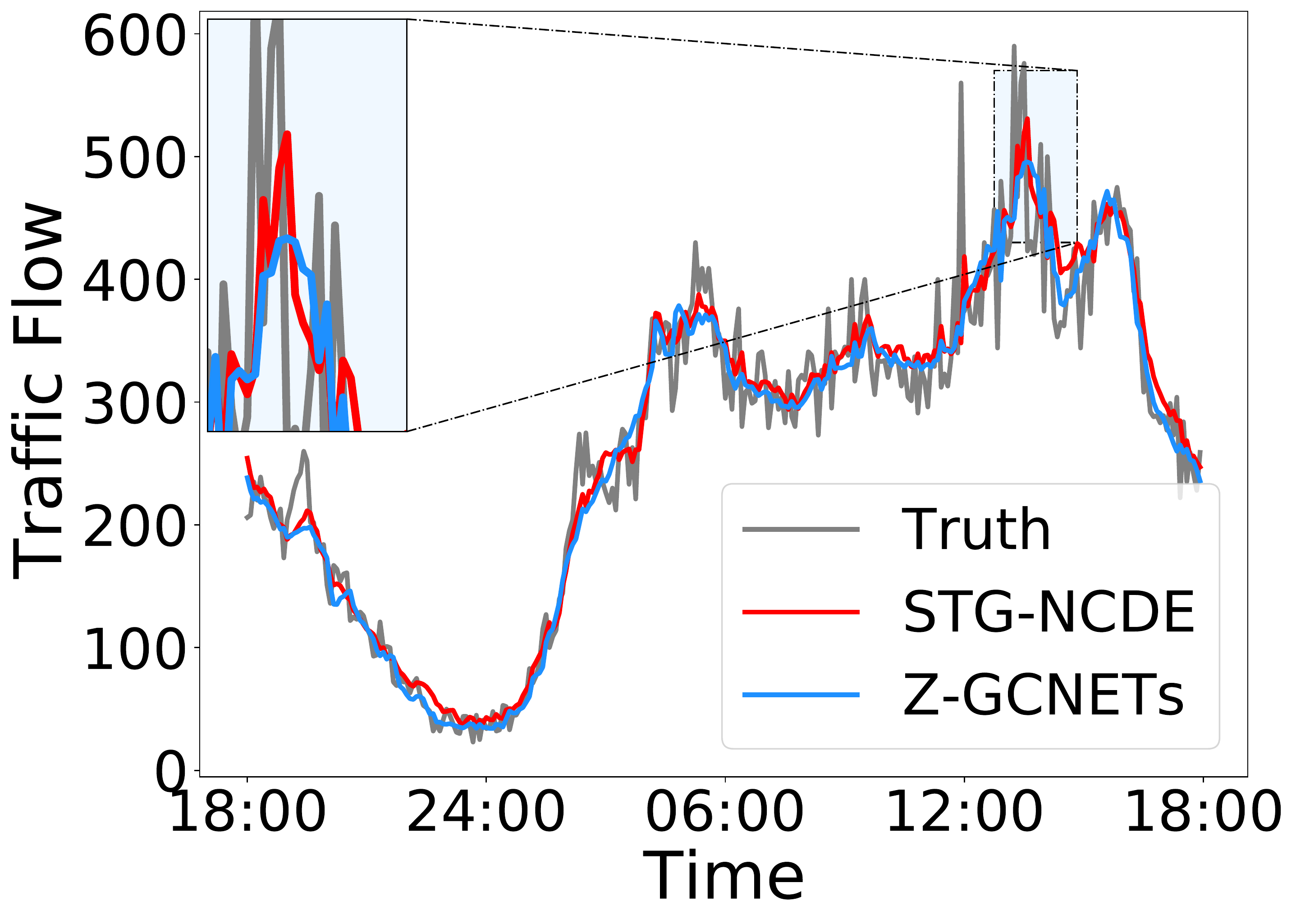}}
    \subfigure[Node 108 in PeMSD3]{\includegraphics[width=0.48\columnwidth]{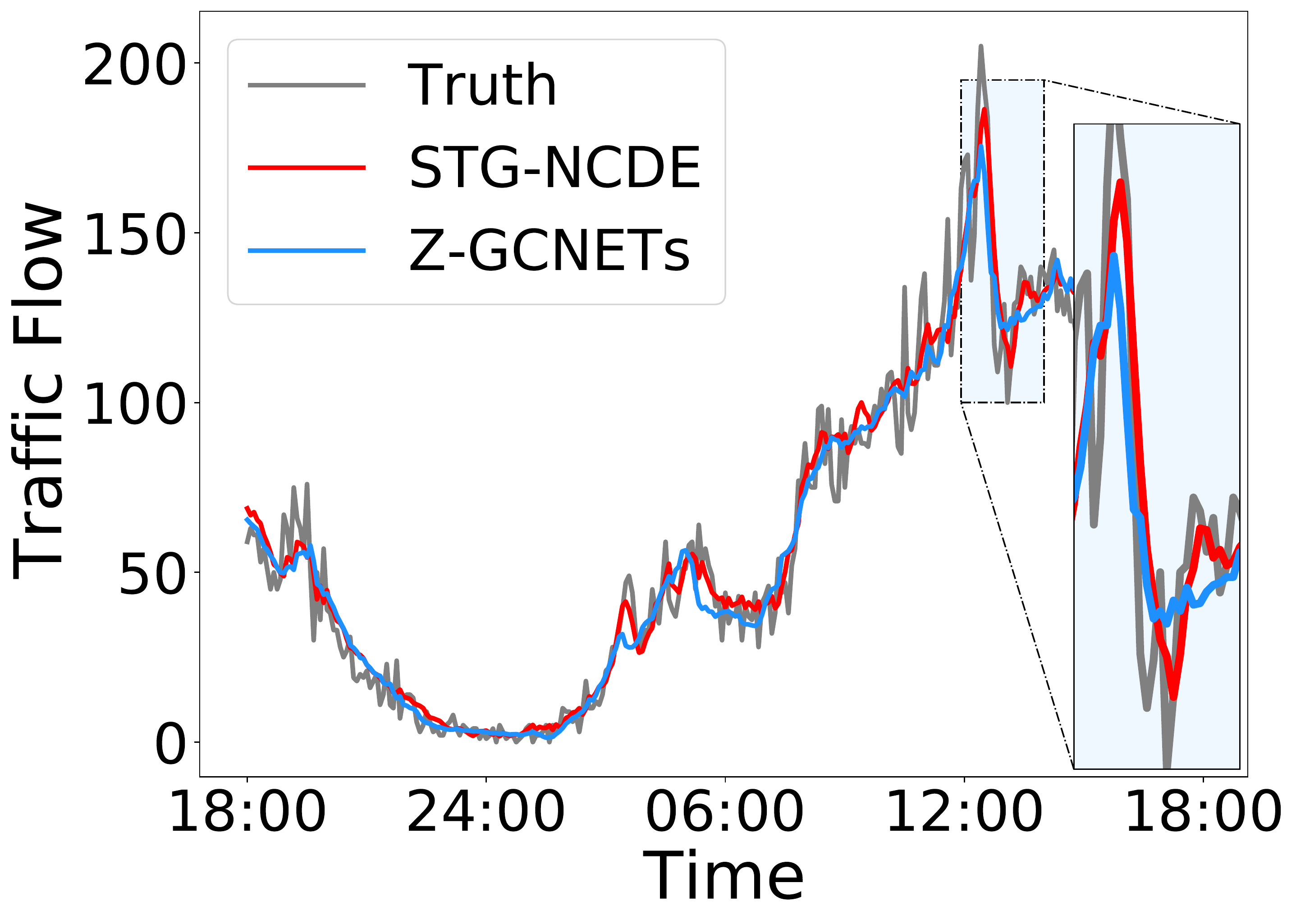}}
    \subfigure[Node 141 in PeMSD3]{\includegraphics[width=0.48\columnwidth]{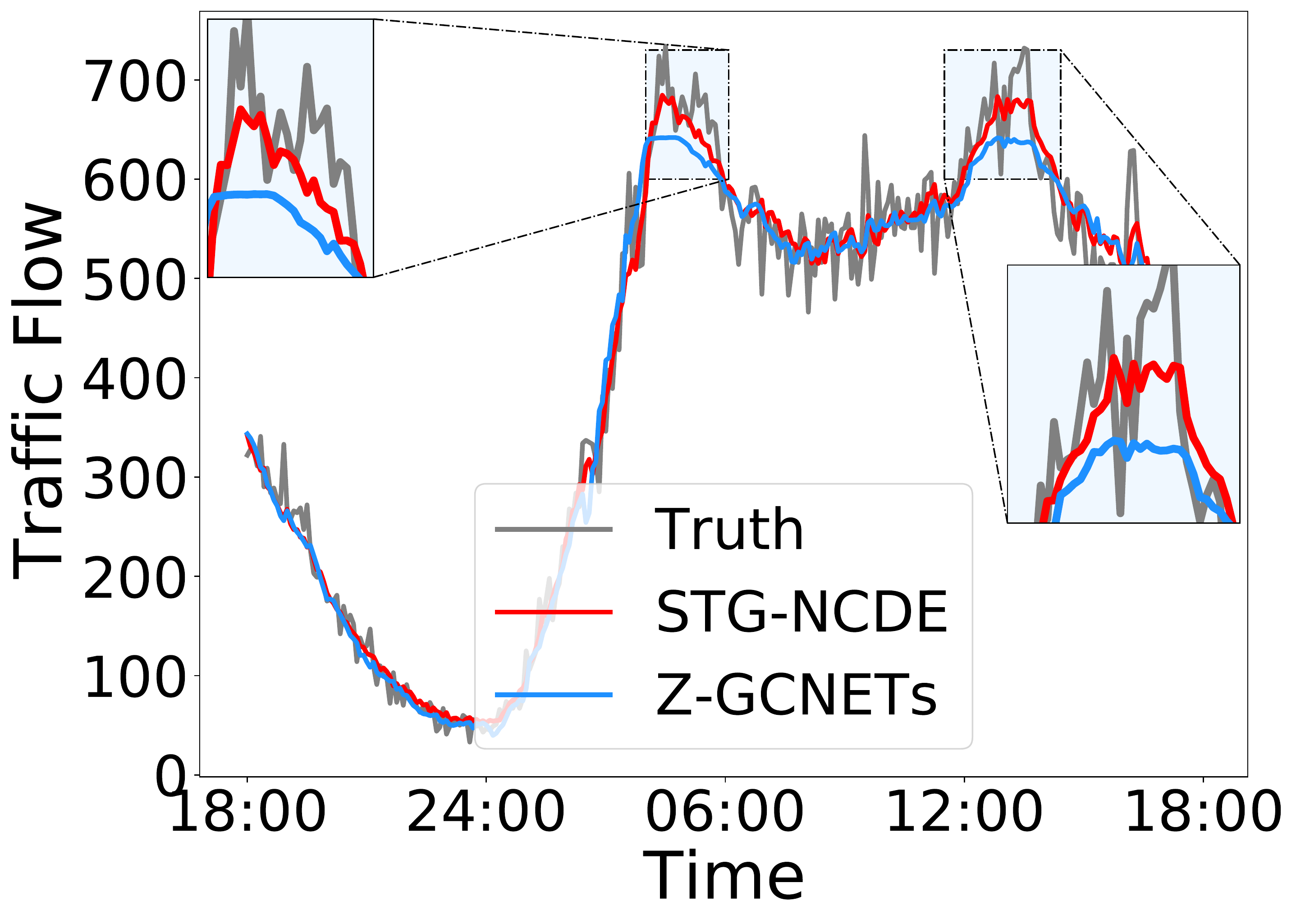}}
    \subfigure[Node 149 in PeMSD4]{\includegraphics[width=0.48\columnwidth]{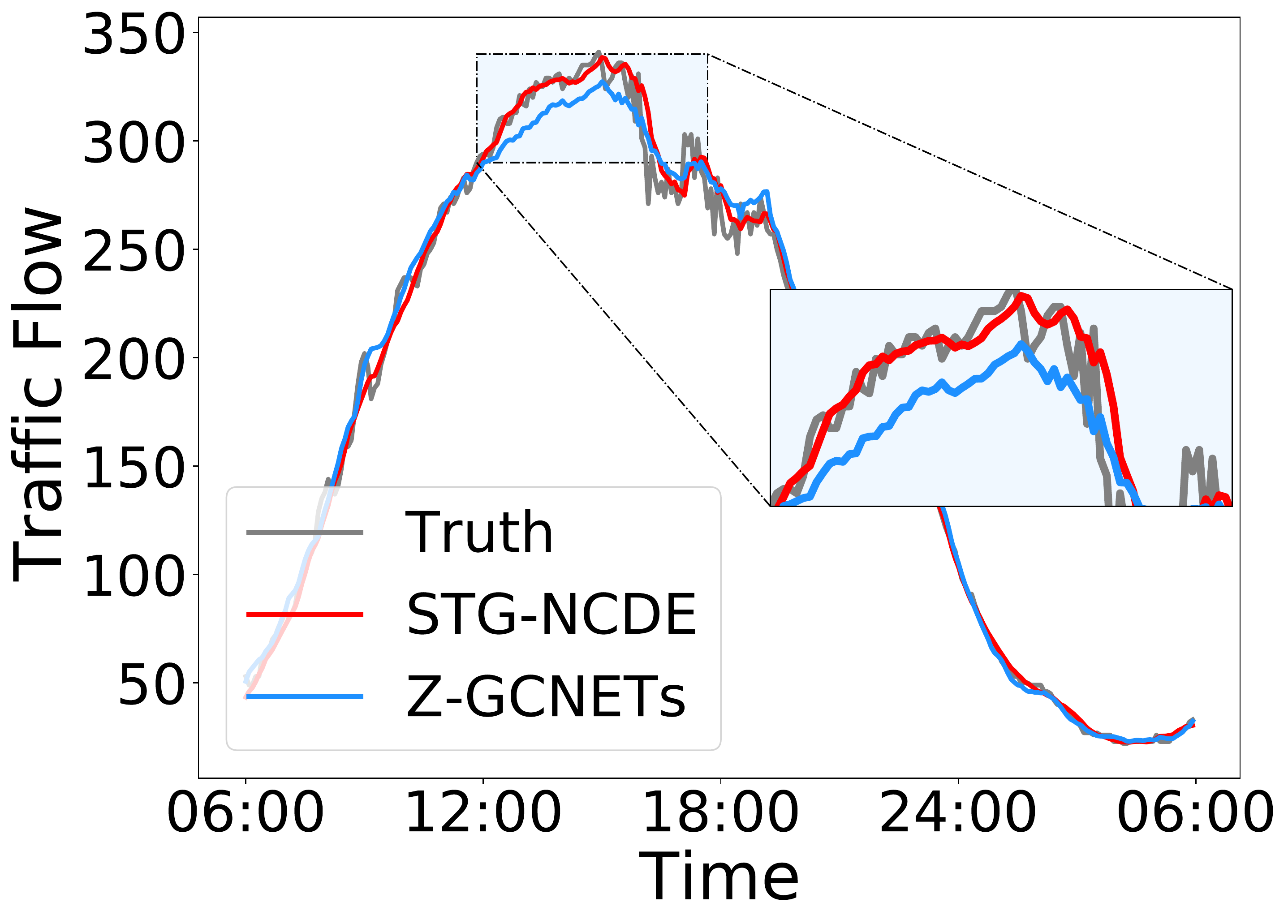}}
    \subfigure[Node 170 in PeMSD4]{\includegraphics[width=0.48\columnwidth]{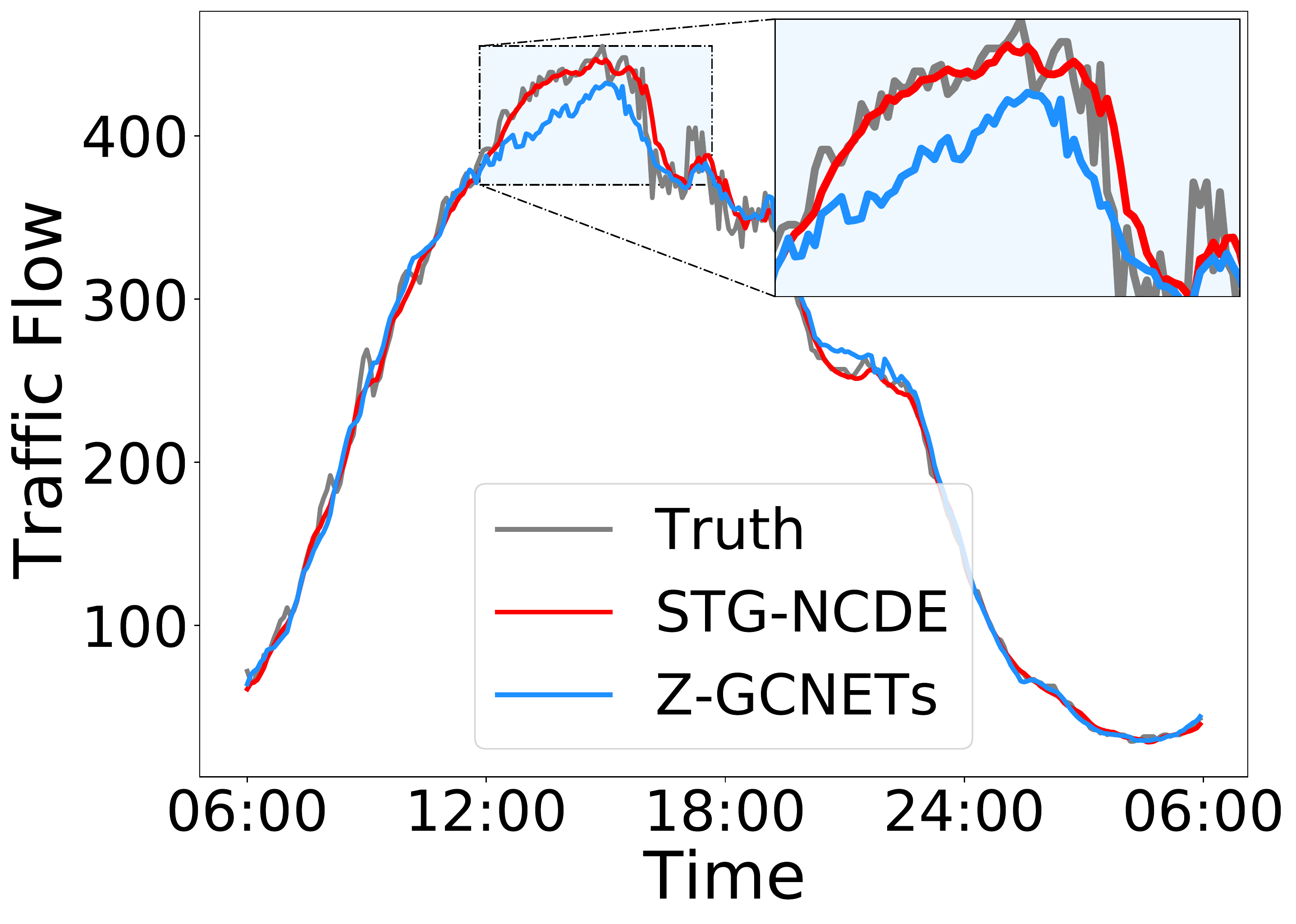}}
    \subfigure[Node 211 in PeMSD4]{\includegraphics[width=0.48\columnwidth]{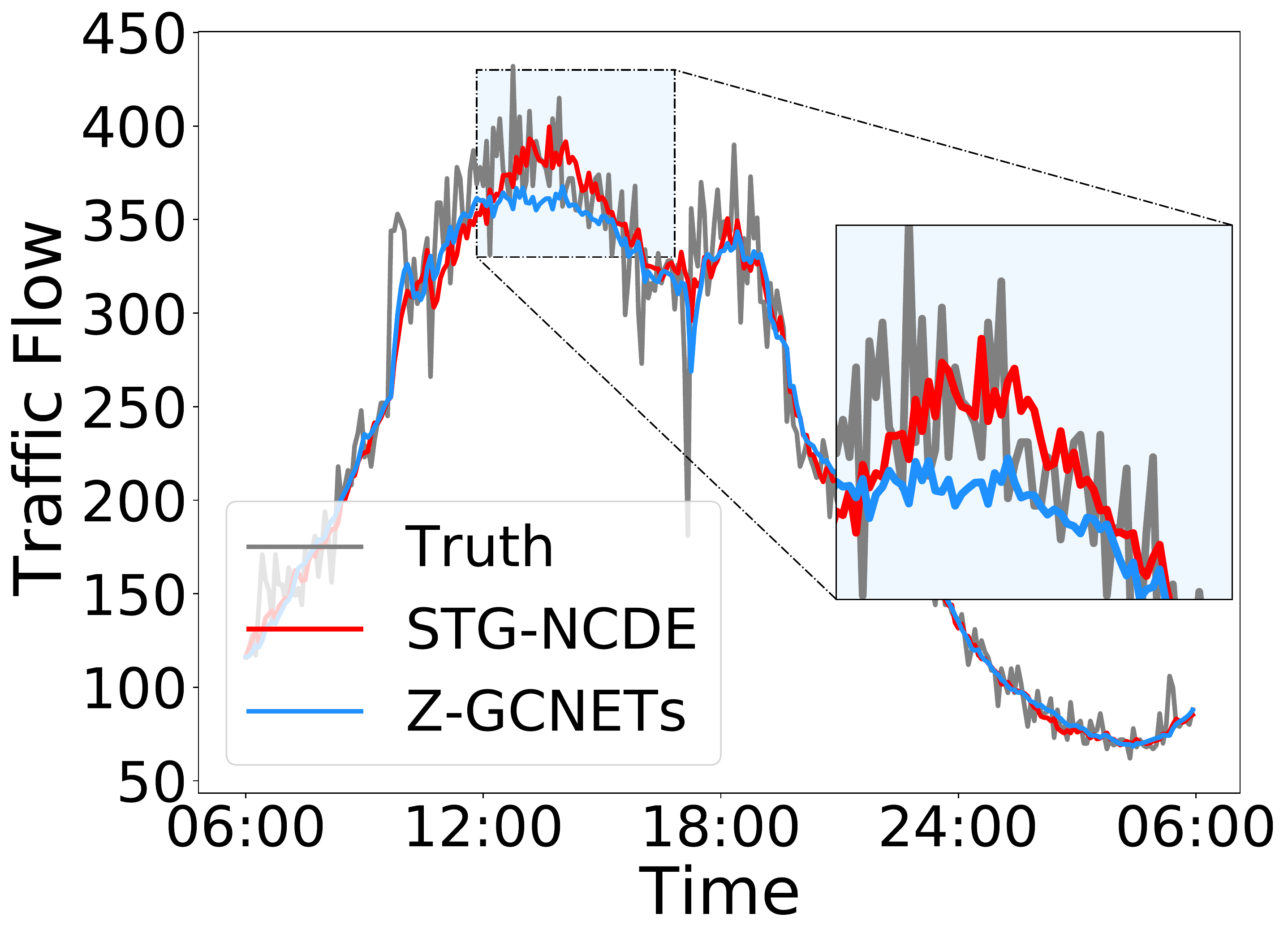}}
    \subfigure[Node 287 in PeMSD4]{\includegraphics[width=0.48\columnwidth]{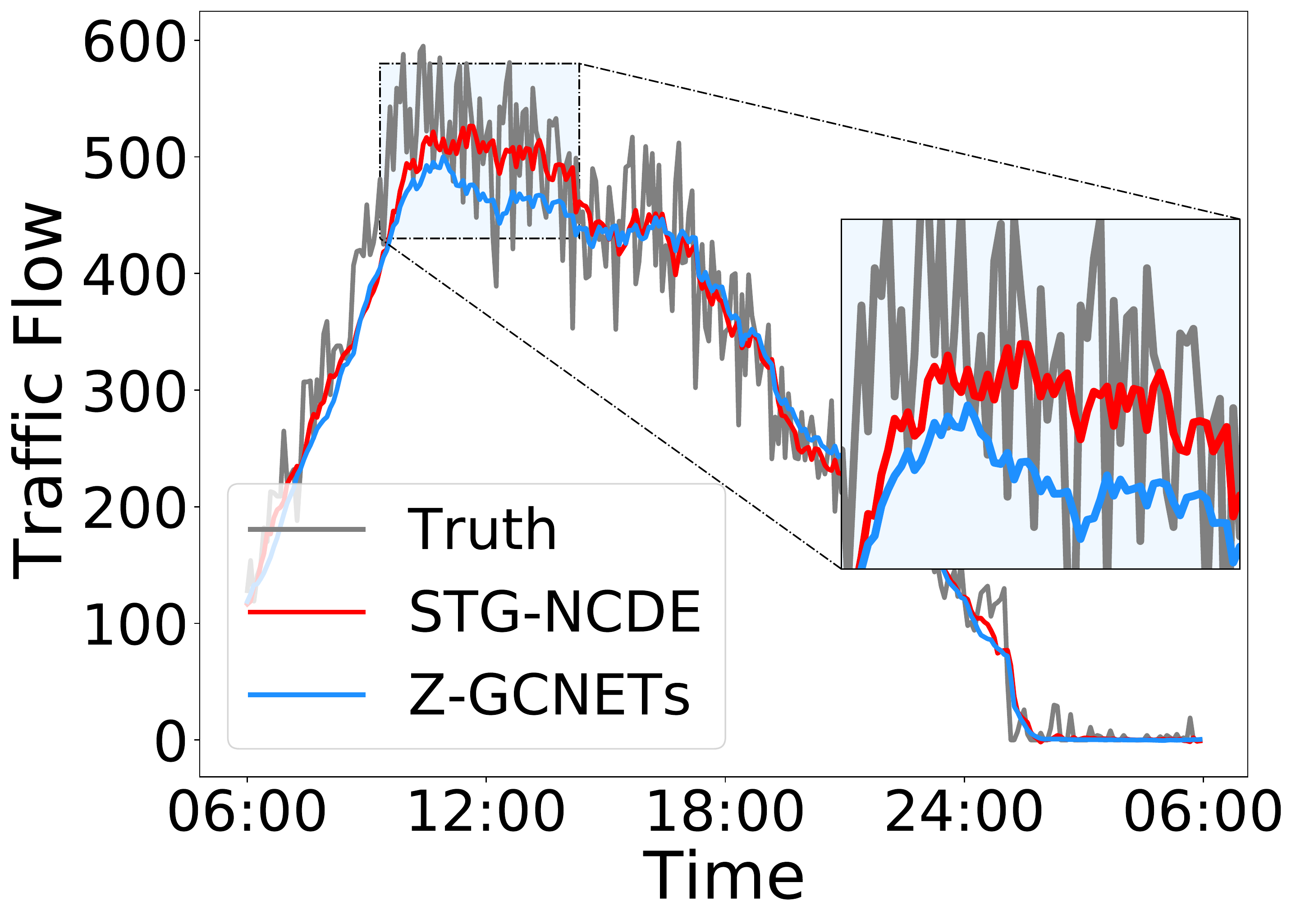}}
    \subfigure[Node 26 in PeMSD7]{\includegraphics[width=0.48\columnwidth]{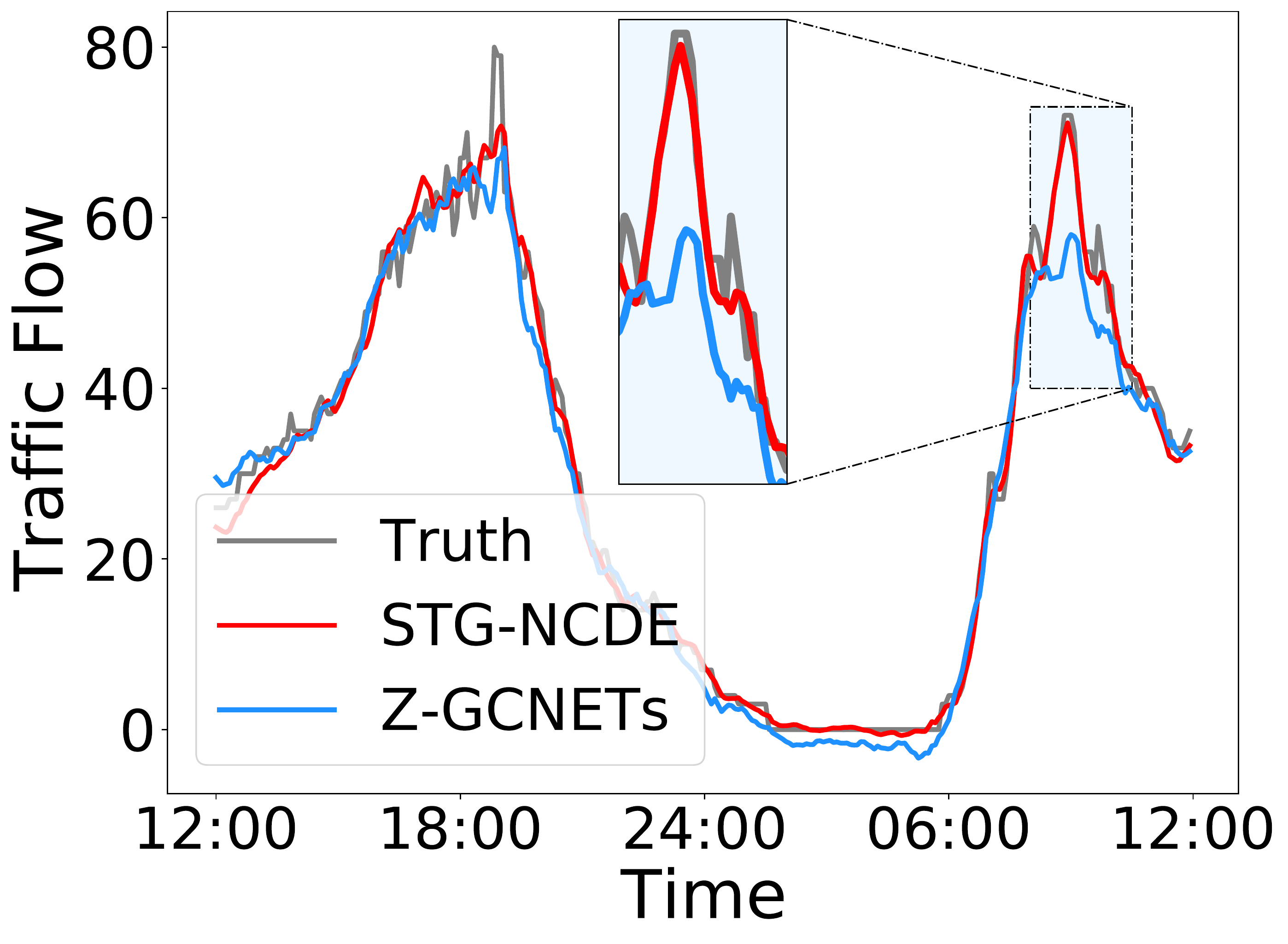}}
    \subfigure[Node 277 in PeMSD7]{\includegraphics[width=0.48\columnwidth]{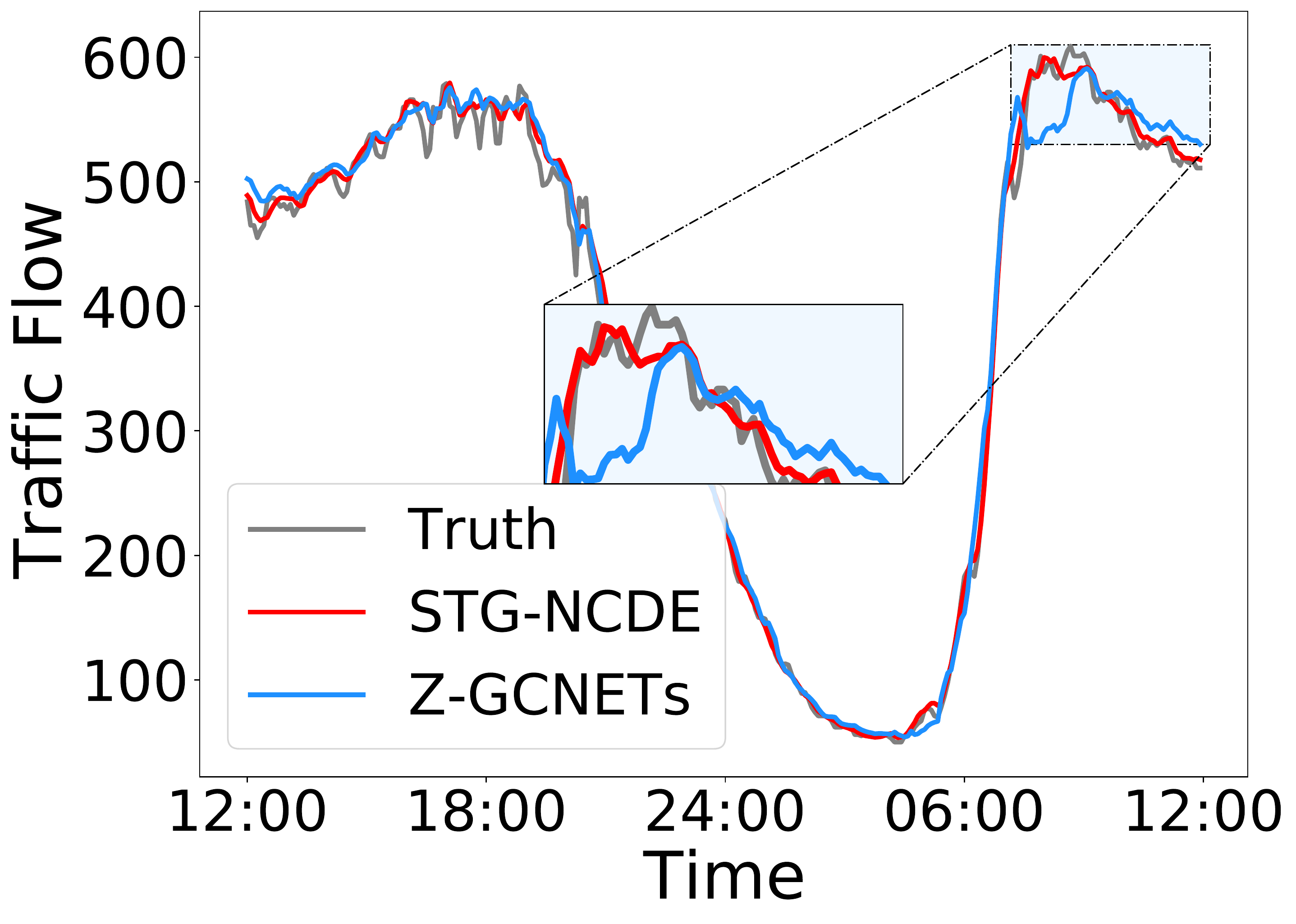}}
    \subfigure[Node 66 in PeMSD7]{\includegraphics[width=0.48\columnwidth]{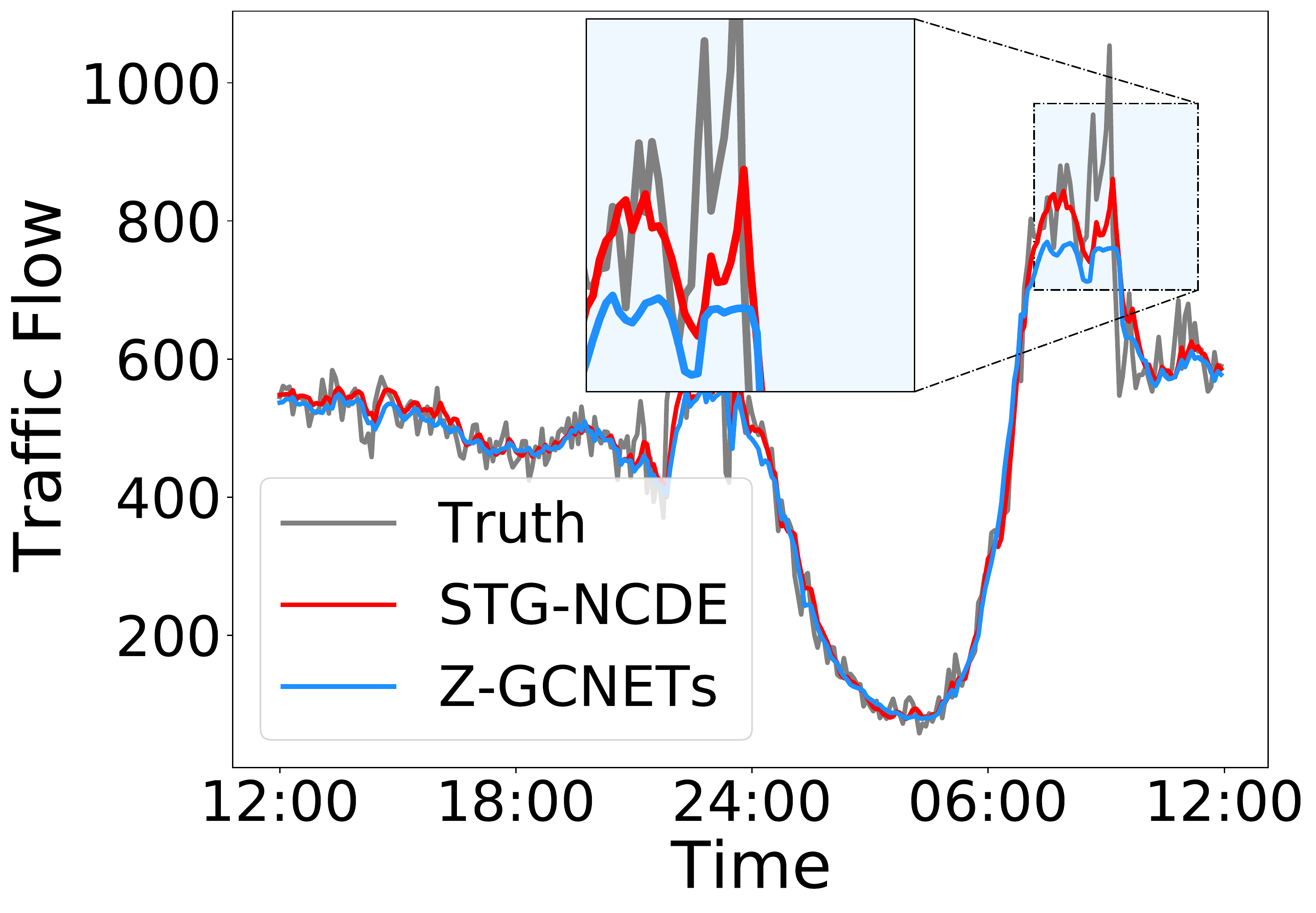}}
    \subfigure[Node 139 in PeMSD7]{\includegraphics[width=0.48\columnwidth]{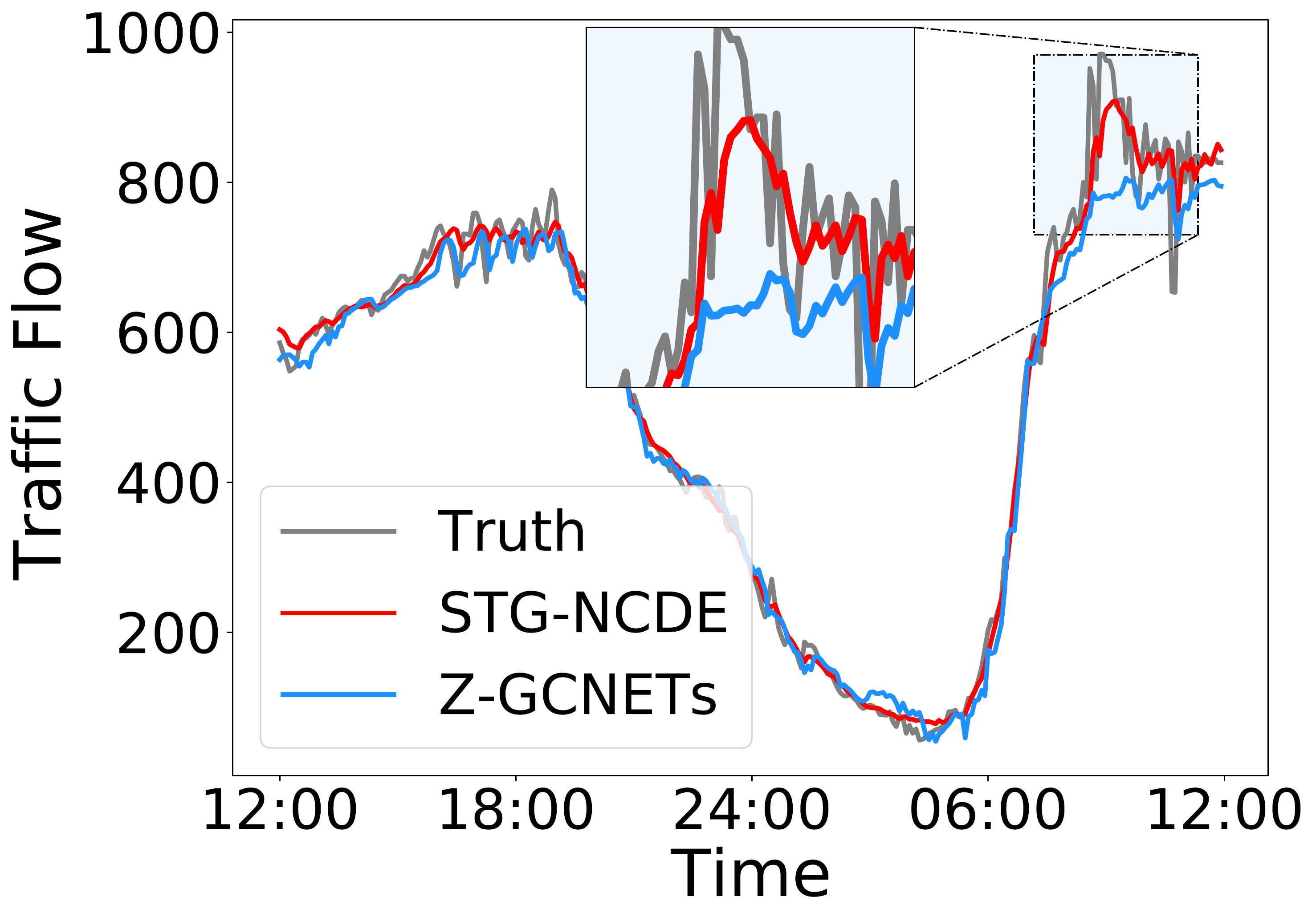}}
    \subfigure[Node 85 in PeMSD8]{\includegraphics[width=0.48\columnwidth]{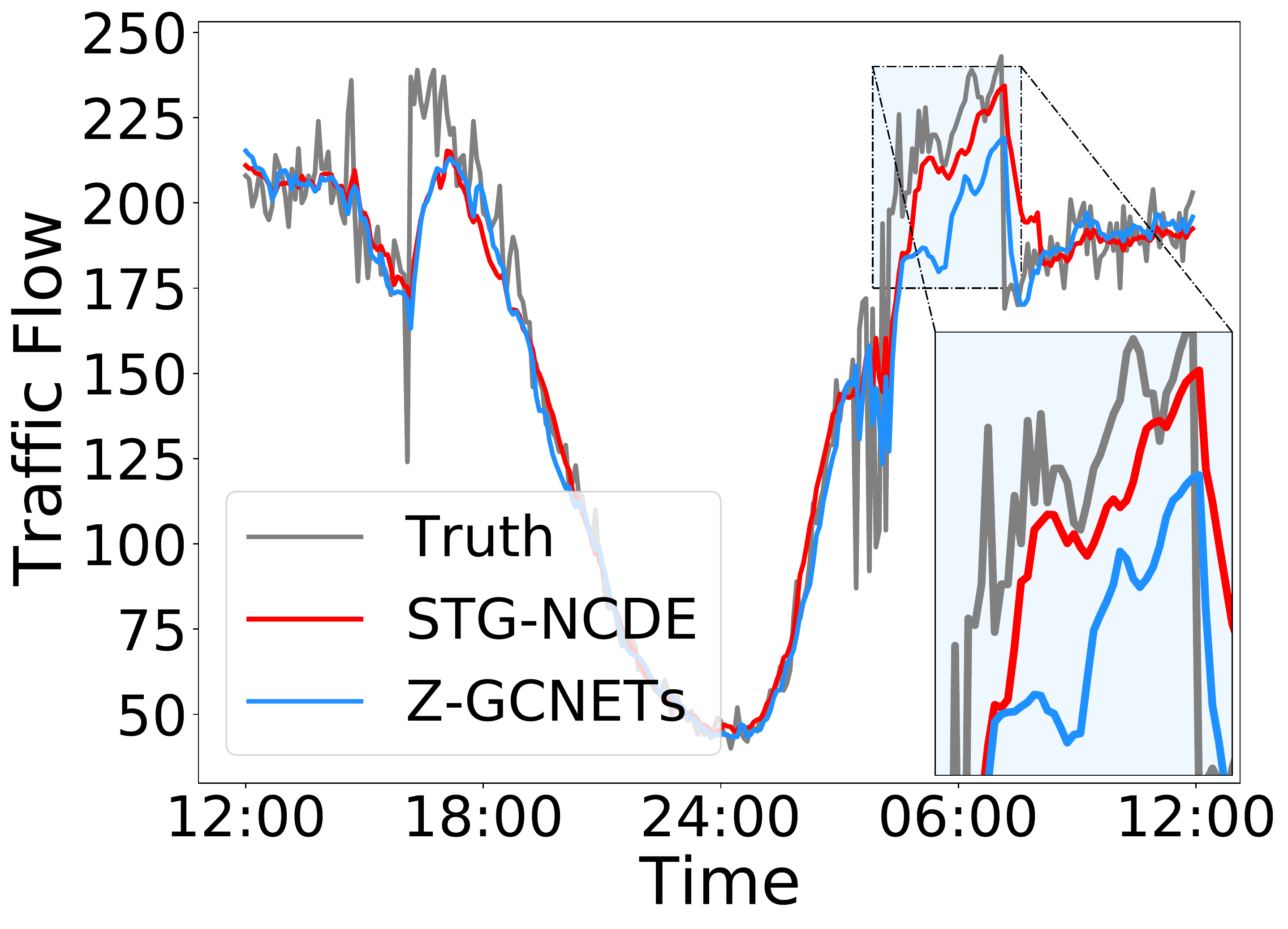}}
    \subfigure[Node 104 in PeMSD8]{\includegraphics[width=0.48\columnwidth]{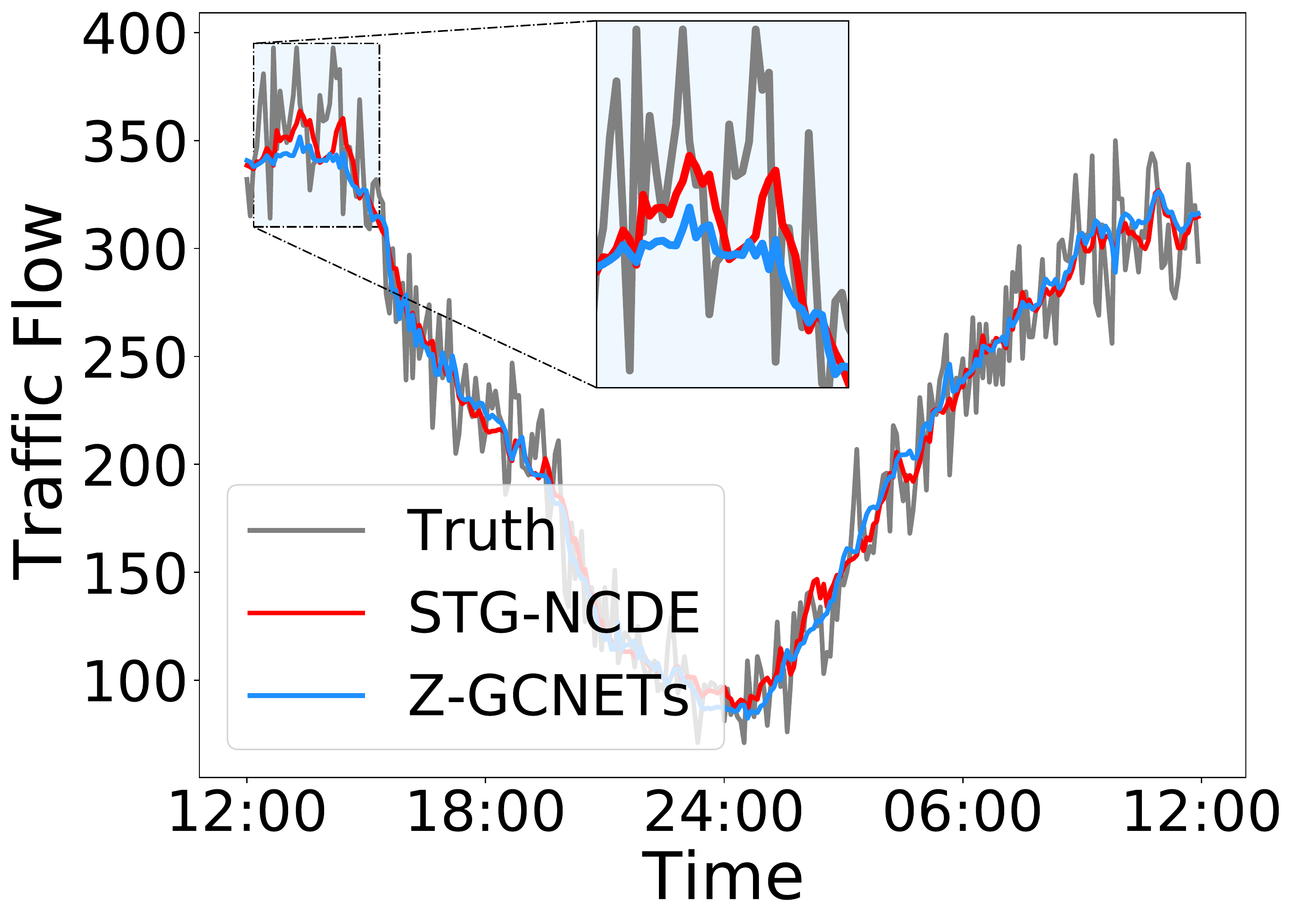}}
    \subfigure[Node 155 in PeMSD8]{\includegraphics[width=0.48\columnwidth]{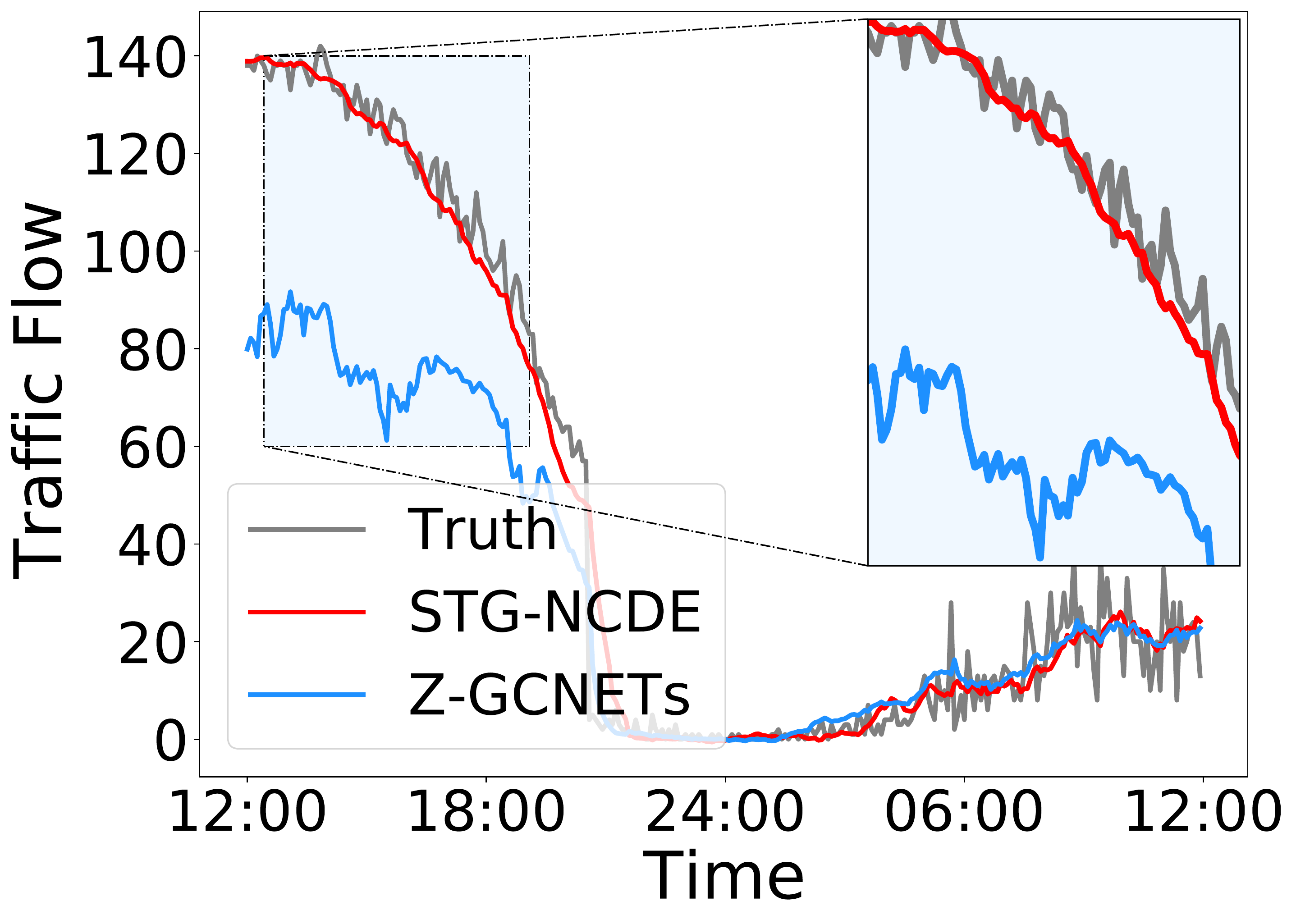}}
    \subfigure[Node 162 in PeMSD8]{\includegraphics[width=0.48\columnwidth]{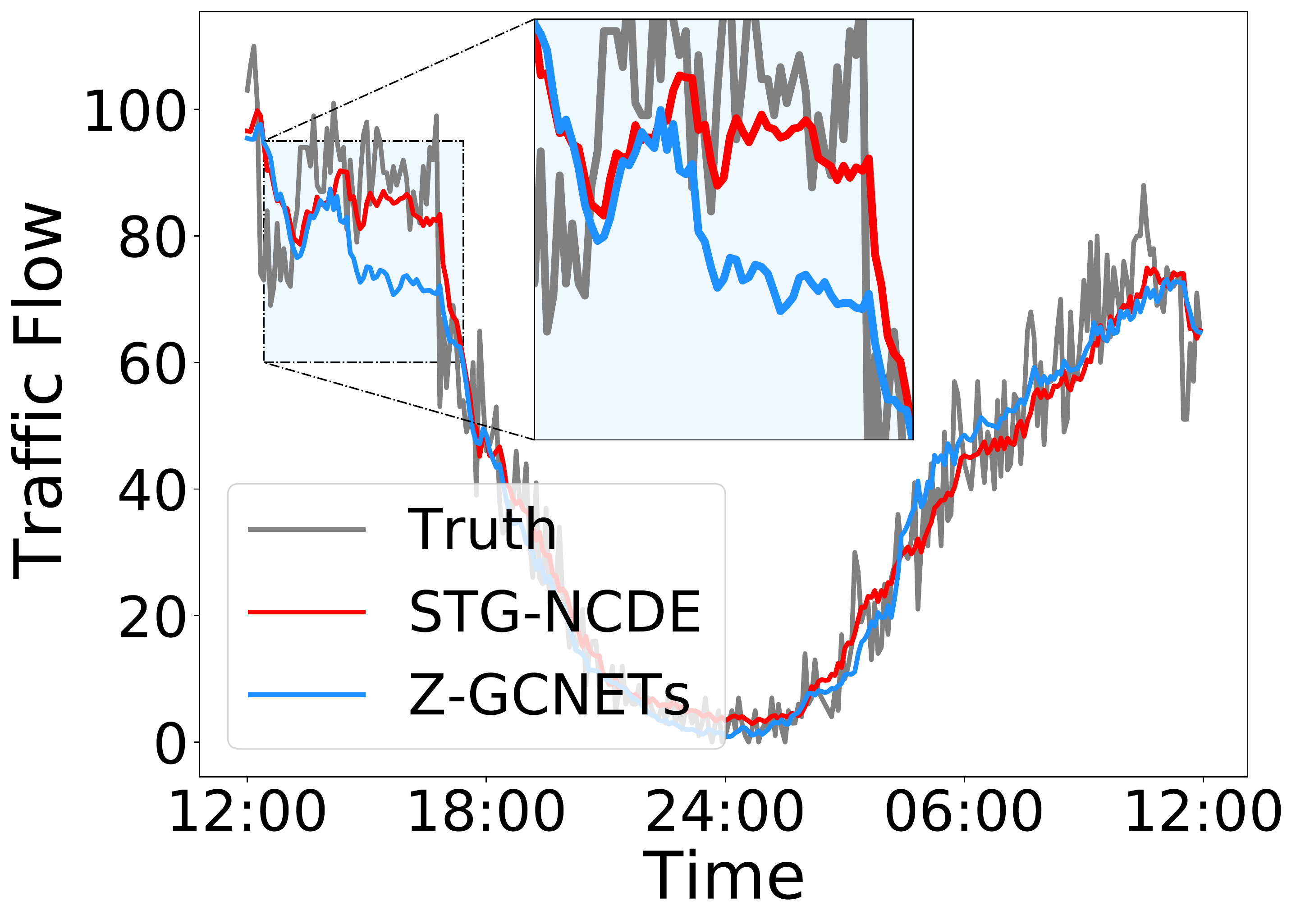}}
    \subfigure[Node 14 in PeMSD7(M)]{\includegraphics[width=0.48\columnwidth]{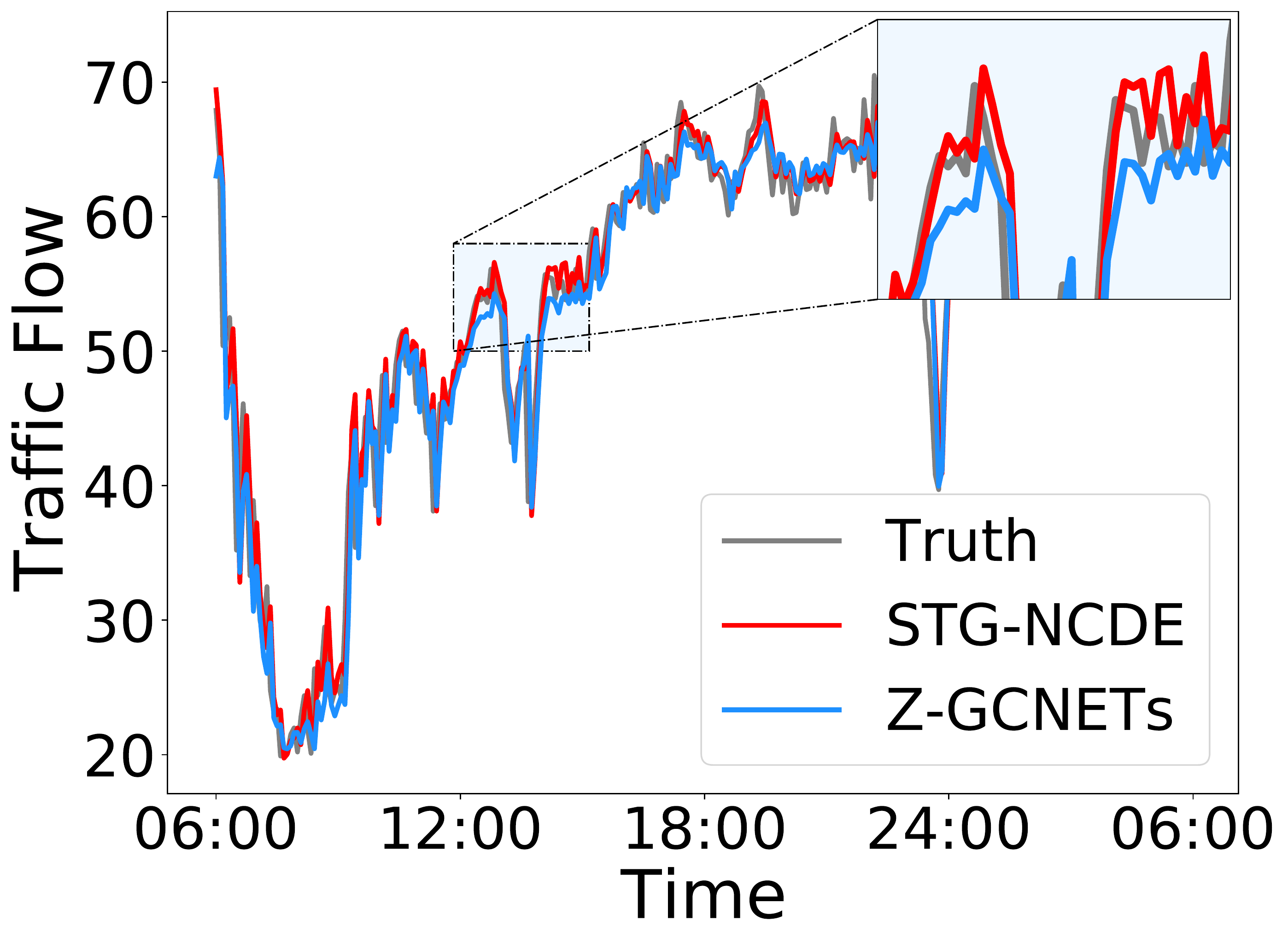}}
    \subfigure[Node 15 in PeMSD7(M)]{\includegraphics[width=0.48\columnwidth]{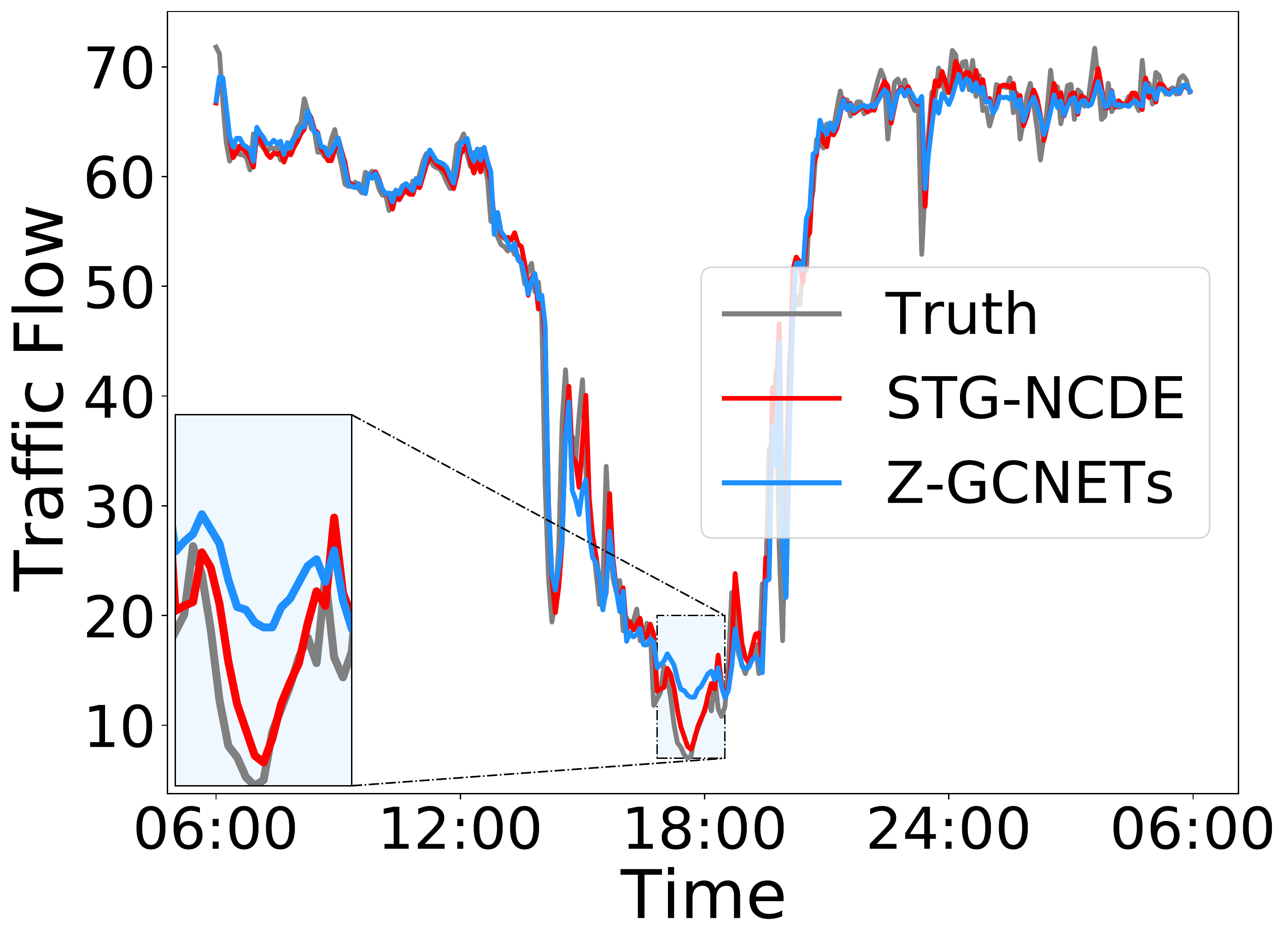}}
    \subfigure[Node 18 in PeMSD7(M)]{\includegraphics[width=0.48\columnwidth]{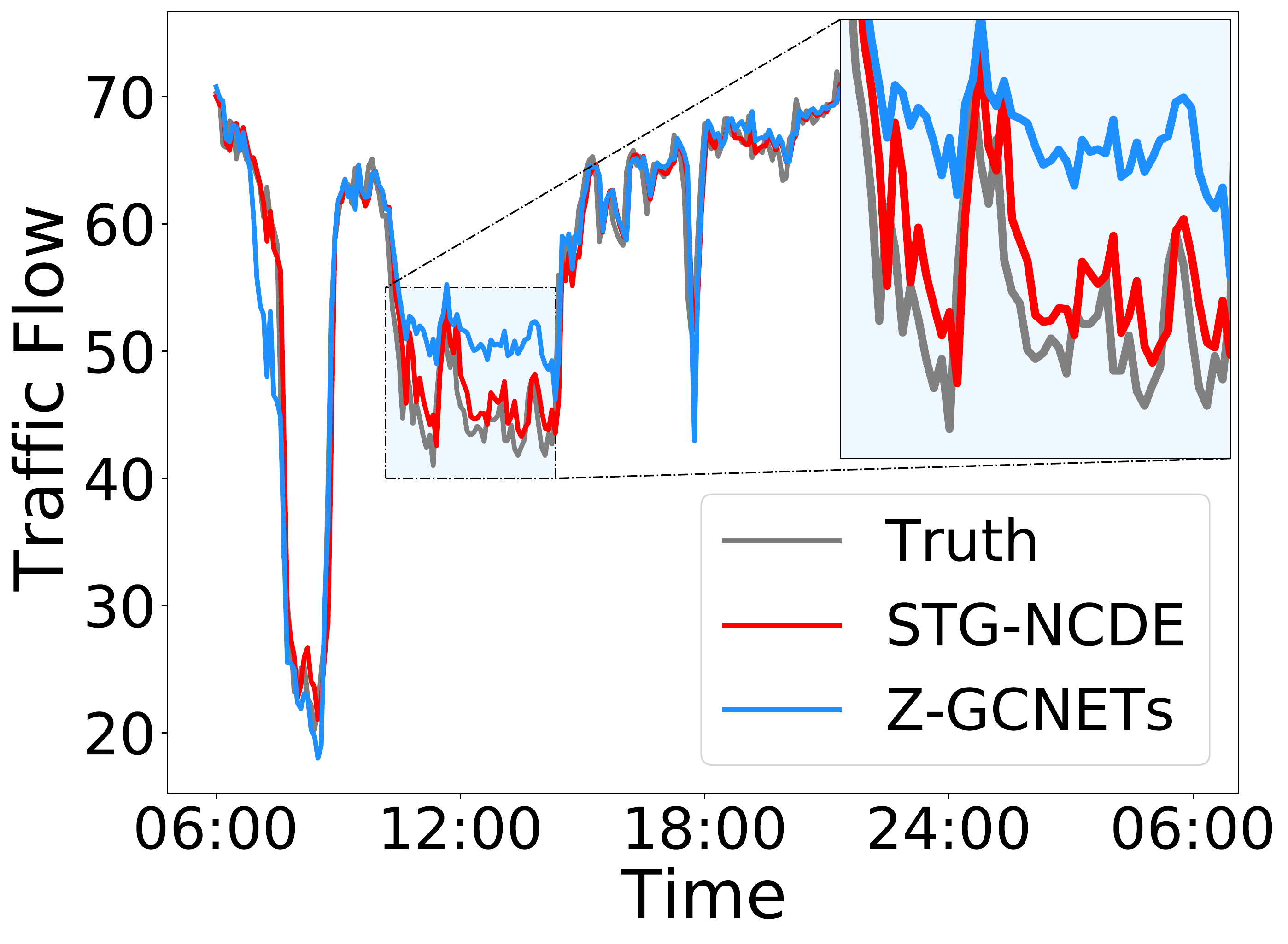}}
    \subfigure[Node 37 in PeMSD7(M)]{\includegraphics[width=0.48\columnwidth]{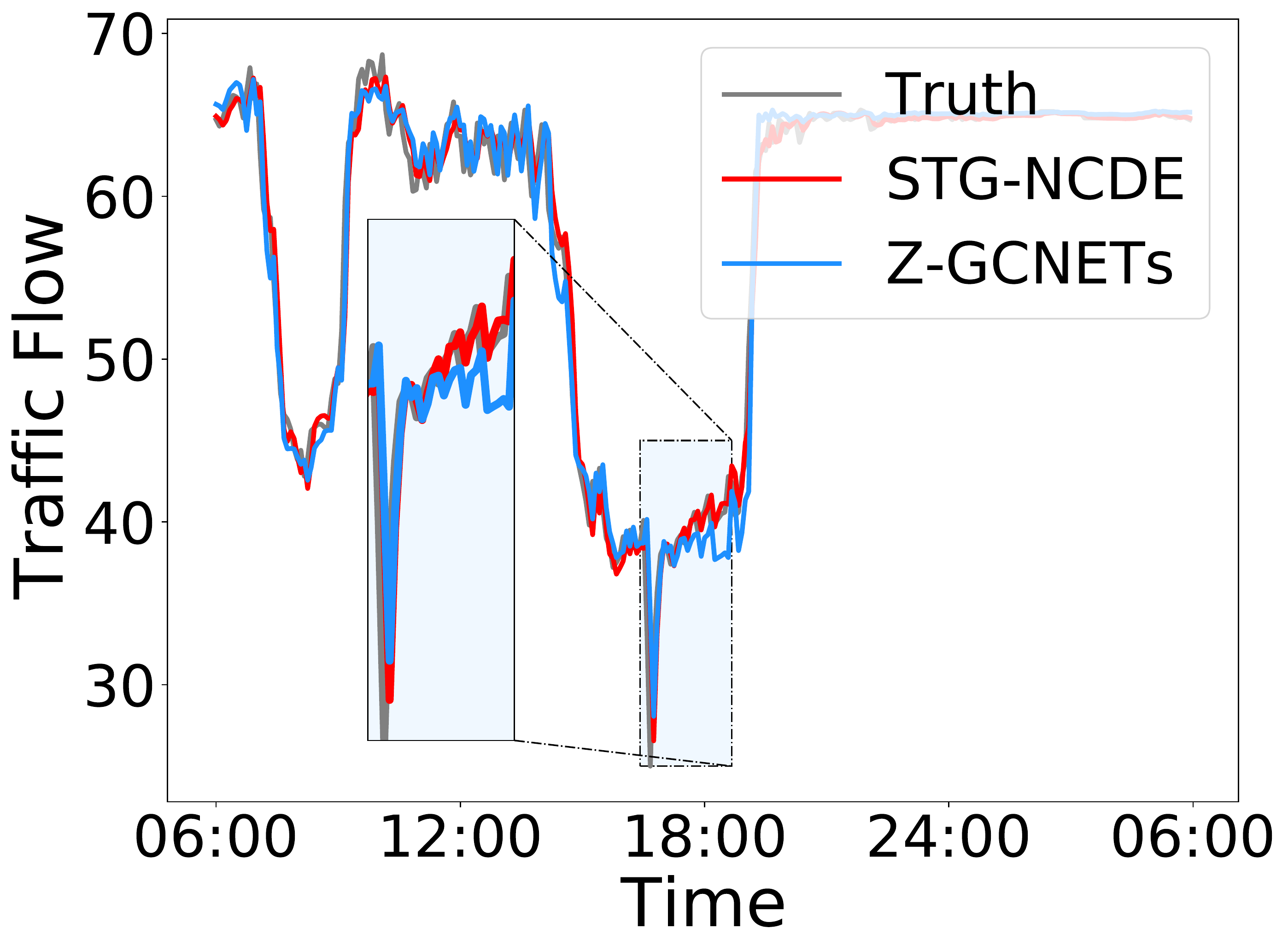}}
    \subfigure[Node 211 in PeMSD7(L)]{\includegraphics[width=0.48\columnwidth]{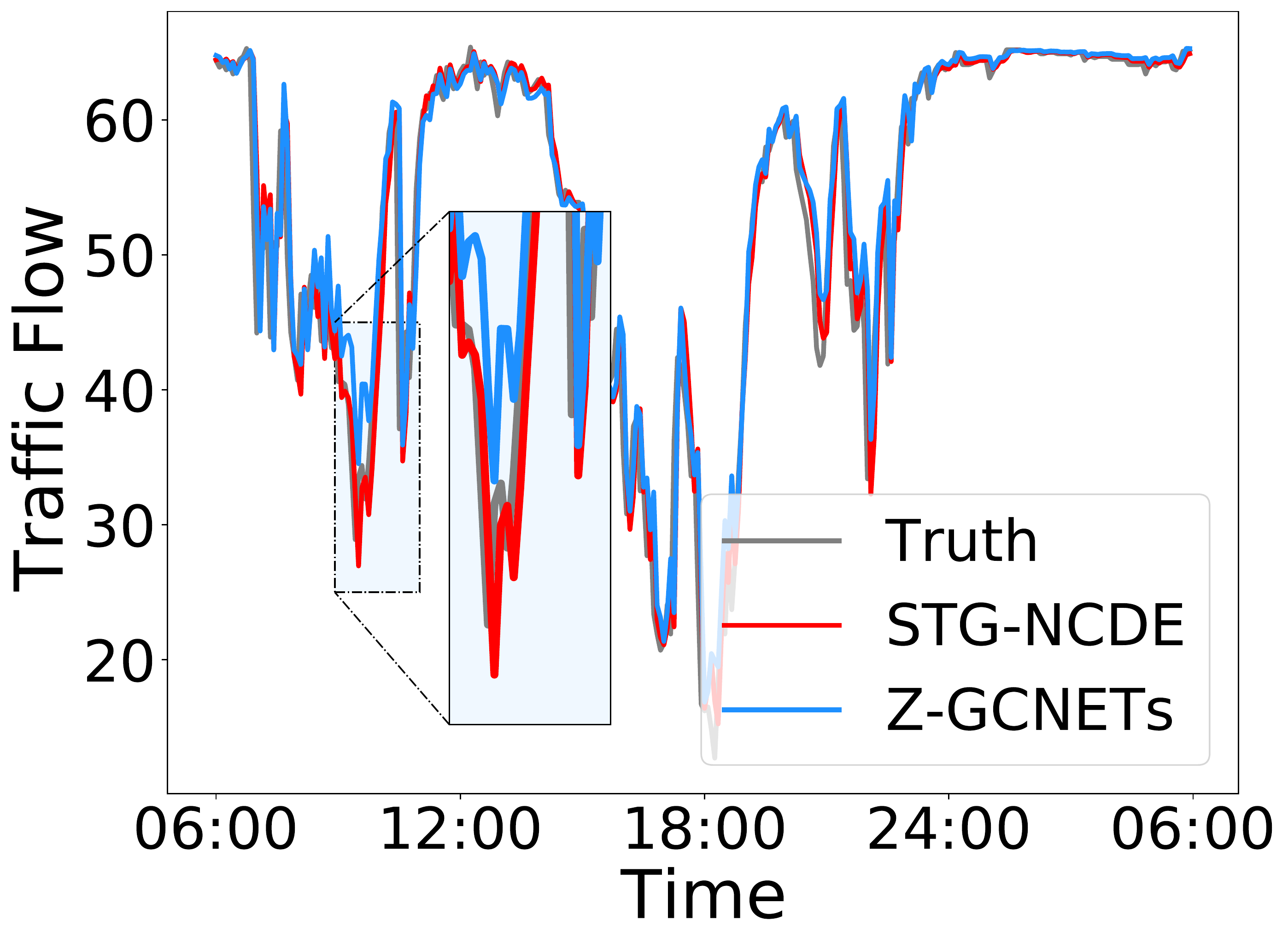}}
    \subfigure[Node 509 in PeMSD7(L)]{\includegraphics[width=0.48\columnwidth]{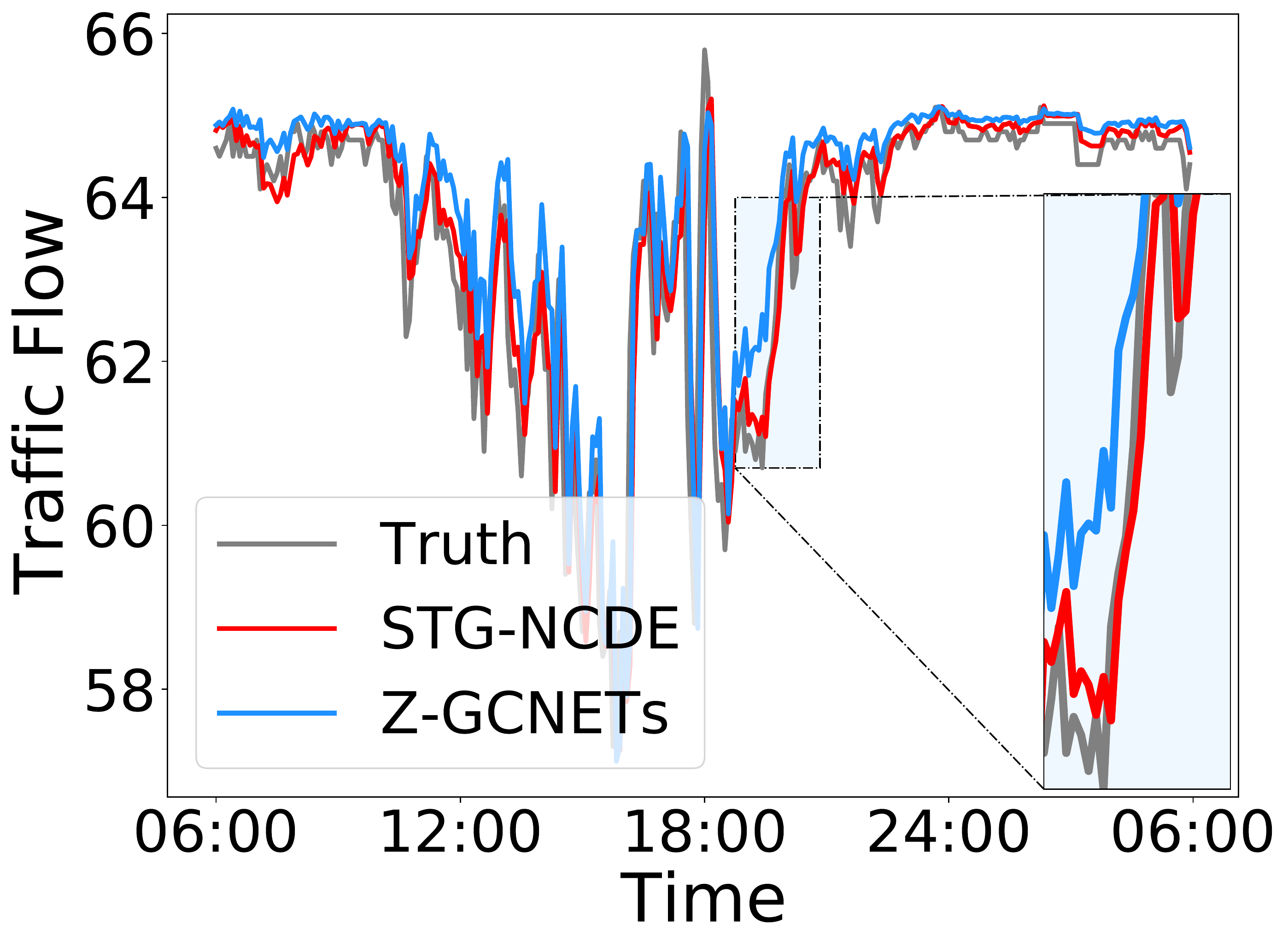}}
    \subfigure[Node 7009 in PeMSD7(L)]{\includegraphics[width=0.48\columnwidth]{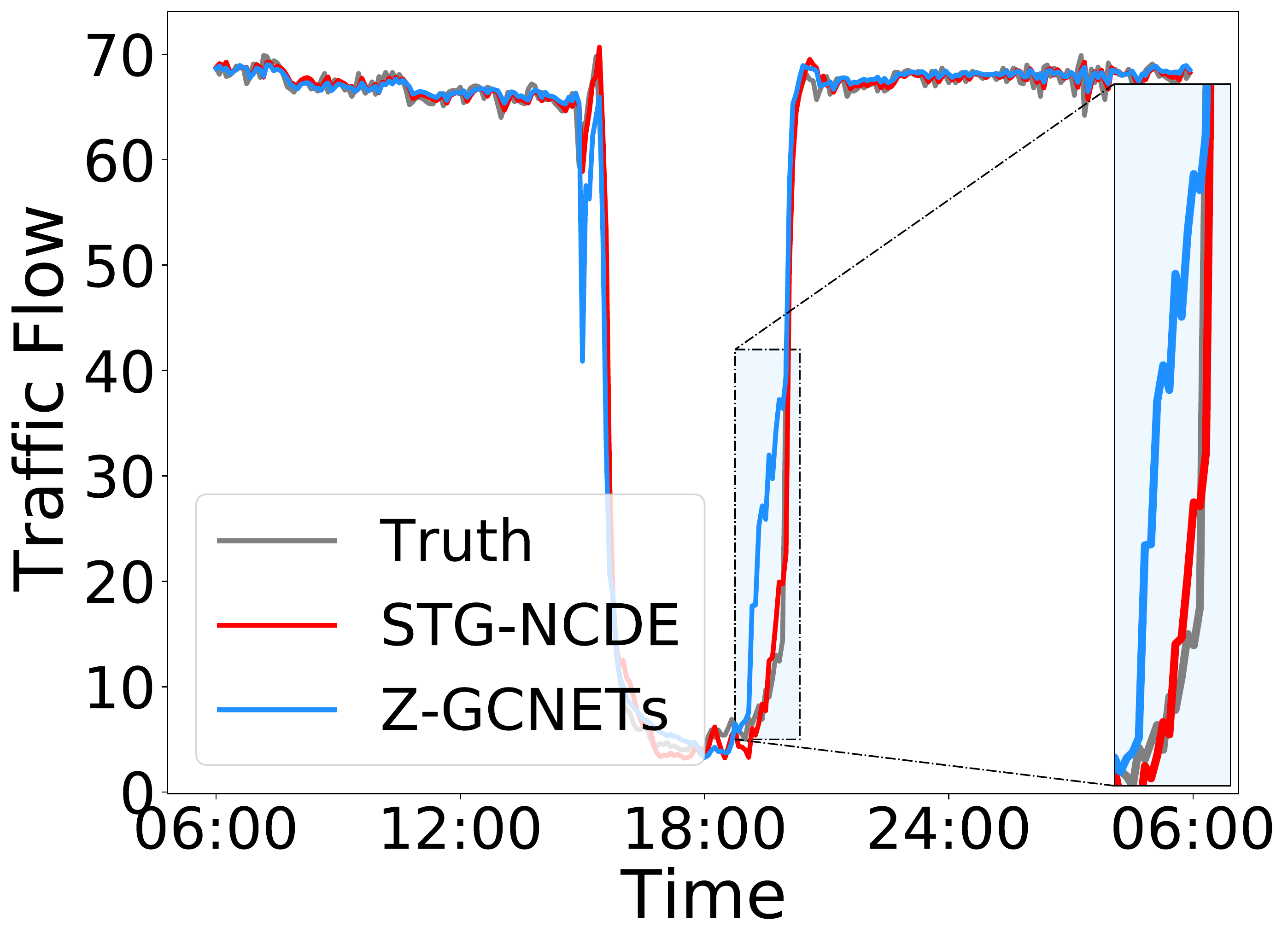}}
    \subfigure[Node 958 in PeMSD7(L)]{\includegraphics[width=0.48\columnwidth]{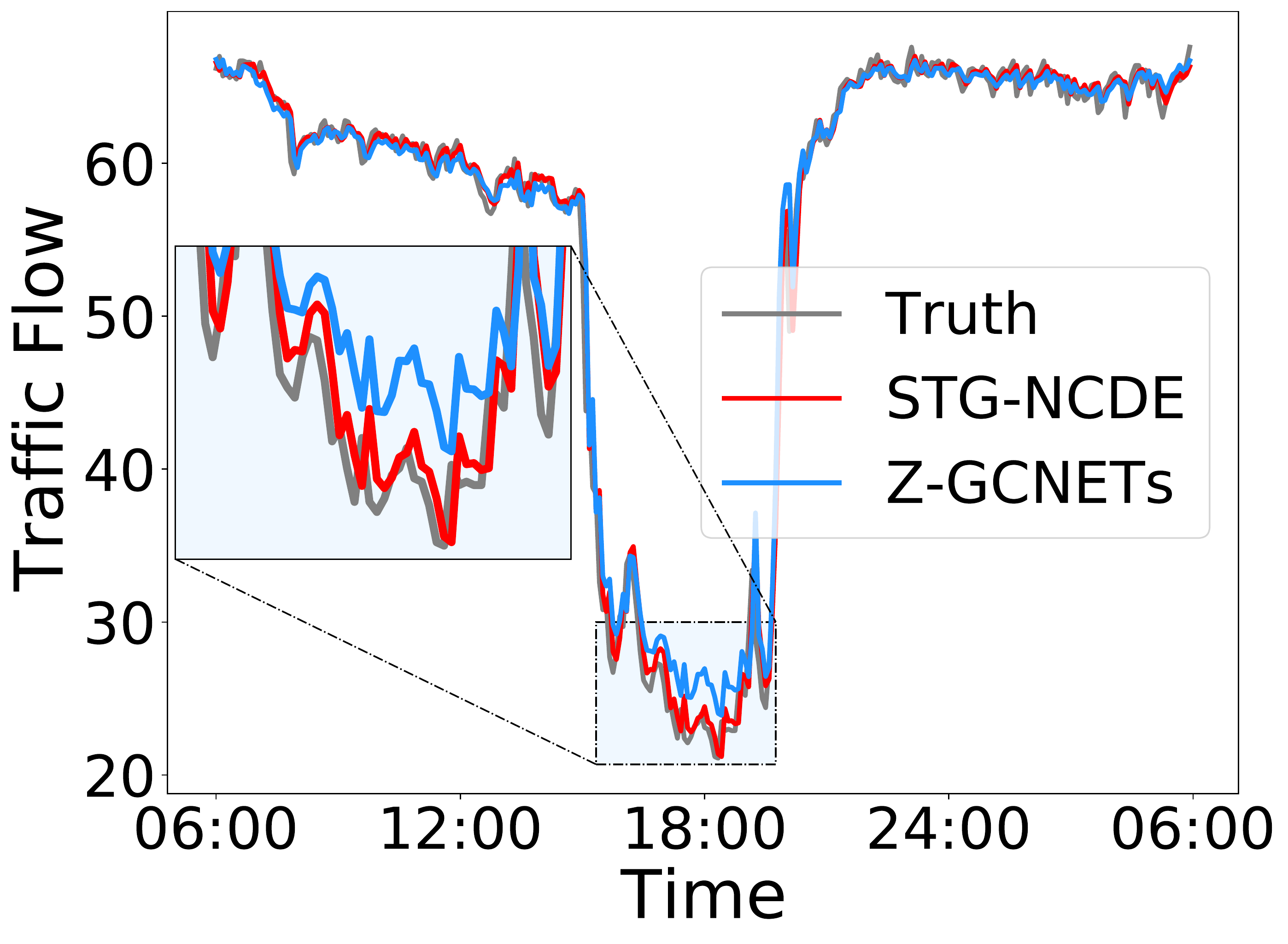}}

    \caption{Traffic forecasting visualization}
    \label{fig:visualize_appendix}
\end{figure*}

We also visualize the ground-truth and some forecasting outcomes by our method and Z-GCNETs in Fig.~\ref{fig:visualize_appendix}.

\end{document}